\documentclass[10pt,twocolumn,letterpaper]{article}

\usepackage{cvpr}              

\usepackage{url}
\usepackage{enumitem}
\usepackage{array}
\usepackage{multirow}
\usepackage{booktabs}
\usepackage{graphicx}
\usepackage{wrapfig}
\usepackage{caption}
\usepackage[font=footnotesize]{caption}
\usepackage{pifont}
\usepackage{titletoc}
\usepackage{extpfeil}
\usepackage{colortbl}   
\usepackage[table]{xcolor} 
\usepackage{subcaption}
\usepackage{tcolorbox}

\usepackage{graphicx}
\usepackage{wrapfig}
\usepackage{booktabs}
\usepackage{makecell}
\usepackage[table]{xcolor} 

\usepackage{booktabs}
\usepackage{array}
\usepackage{colortbl}
\usepackage{subcaption}   
\usepackage{adjustbox}    
\usepackage{tabularx}

\usepackage{booktabs, tabularx, array, xcolor, siunitx, subcaption}
\definecolor{LightGreen}{RGB}{219, 249, 237}

\definecolor{tkcolor}{RGB}{224,223,255}
\definecolor{tkcolor2_back}{RGB}{242,242,242}
\definecolor{tkcolor2_frame}{RGB}{50,50,50}
\definecolor{darkgreen}{RGB}{0,200,0}
\newtcolorbox{takeaways}[2][]{
	width=\columnwidth,
        toprule=0.0pt,
        leftrule=0.9pt,
        bottomrule=0.9pt,
        rightrule=0.9pt,
        arc=0pt,
	colback = tkcolor2_back, 
	colframe = tkcolor2_frame, 
	boxsep=0pt,left=7pt,right=7pt,top=4pt,bottom=4pt,
	fontupper=\linespread{1.1}\selectfont,
	title=#2,#1}

\definecolor{LightGreen}{RGB}{219, 249, 237}

\definecolor{cvprblue}{rgb}{0.21,0.49,0.74}
\usepackage[pagebackref,breaklinks,colorlinks,allcolors=cvprblue]{hyperref}

\def\paperID{6639} 
\def\confName{CVPR}
\def\confYear{2026}

\title{OmniDocLayout: Towards Diverse Document Layout Generation via Coarse-to-Fine LLM Learning}

\vspace{-6mm}
\author{
    Hengrui Kang\textsuperscript{\rm 1,2}\thanks{Equal contribution.},\quad
    \newcounter{example1} 
    \setcounter{example1}{1}
    Zhuangcheng Gu\textsuperscript{\rm 2\fnsymbol{example1}},\quad
    Zhiyuan Zhao\textsuperscript{\rm 2},\quad
    Zichen Wen\textsuperscript{\rm 1,2},
     \\
    Bin Wang\textsuperscript{\rm 2},\quad
    ~Weijia Li\textsuperscript{\rm 2,3}\thanks{Corresponding authors.},\quad
    \newcounter{example} 
    \setcounter{example}{2}
    Conghui He\textsuperscript{\rm 2\fnsymbol{example}}
    \vspace{0.45em}
     \\
    \normalsize \textsuperscript{\rm 1} Shanghai Jiao Tong University, 
    \normalsize \textsuperscript{\rm 2} Shanghai AI Laboratory,
    \normalsize \textsuperscript{\rm 3} Sun Yat-sen University \\
    \normalsize E-mail(s):  \texttt{liweij29@mail.sysu.edu.cn, heconghui@pjlab.org.cn}
}

\begin{document}
\maketitle
\begin{abstract}
Document AI has advanced rapidly and is attracting increasing attention. Yet, while most efforts have focused on document layout analysis (DLA), its generative counterpart, layout generation, remains underexplored. 
Distinct from traditional graphic layout design and room layout planning, document layout generation typically involves a larger number of elements per page and exhibits greater structural diversity and complexity.
Currently, a major obstacle lies in the scarcity of diverse document layouts: academic papers with Manhattan-style structures dominate existing studies, while open-world genres such as newspapers and magazines remain severely underrepresented.
To address this gap, we curate OmniDocLayout-1M, the first million-scale dataset of diverse document layouts, covering six common document types and comprising contemporary layouts collected from multiple sources.
Moreover, since existing methods struggle in complex domains and often fail to arrange long sequences coherently, we introduce OmniDocLayout-LLM, a 0.5B model with designed two-stage Coarse-to-Fine learning paradigm:
1) learning universal layout principles from our dataset with coarse category definitions, and 2) transferring the knowledge to a specific domain with few fine-grained annotated samples.
Extensive experiments demonstrate that our approach achieves strong performance on multiple domains in M\textsuperscript{6}Doc dataset, substantially surpassing both existing layout generation experts and several latest general-purpose LLMs. Our code, dataset, and models will be publicly released.
\end{abstract}    

\vspace{-2mm}
\section{Introduction}
\label{sec:intro}

Document AI has attracted growing attention across both academia and industry recently, as it plays a critical role in enabling machines to understand, process, and generate documents.
On the one hand, increasing efforts have been devoted to document parsing~\citep{wang2024mineruopensourcesolutionprecise,li2025monkeyocrdocumentparsingstructurerecognitionrelation,cui2025paddleocr30technicalreport,dotsocr}, which aims to extract structural and semantic information from massive amounts of pages through layout analysis and optical character recognition (OCR)~\citep{zhang2025ocr}, 
On the other hand, its counterpart—document layout generation~\citep{gupta2021layouttransformer,kong2022blt}, has not yet been fully explored. This task focuses on producing well-organized layouts by arranging visual elements like text blocks, tables, and figures in a coherent manner, with great potential for applications including content-driven layout design and document image generation. 
Compared with graphic layout design~\citep{zheng2019content,zhou2022composition,Hsu_2023_CVPR} or room layout planning~\citep{nauata2020house,hu2020graph2plan}, document layout generation is more challenging: each page often contains far more elements, and different document types exhibit highly heterogeneous structural patterns, making unified representation learning difficult.

\begin{figure}[!t] 
    \centering
    \includegraphics[width=1.0\linewidth, height=0.72\linewidth]{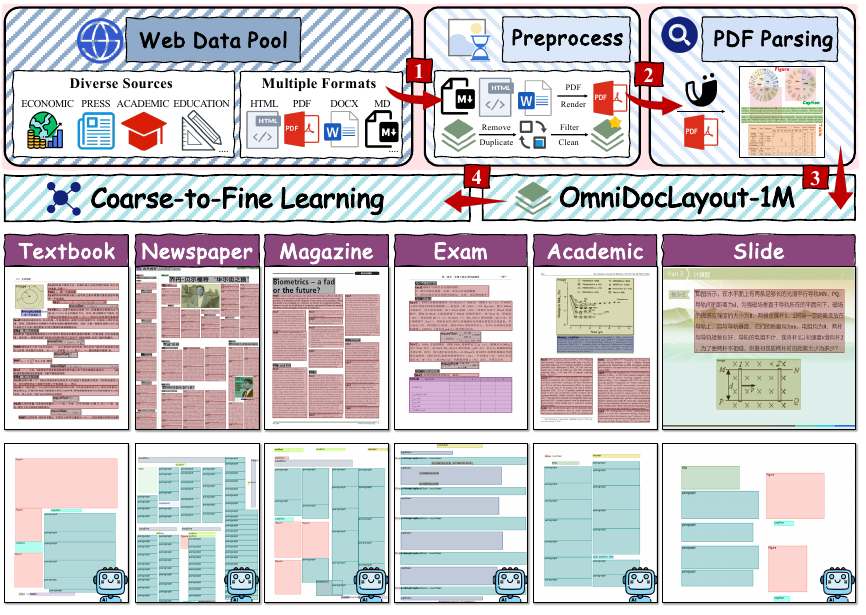}
    \vspace{-6mm}
    \caption{\textbf{Overview of OmniDocLayout.} (Top \& Middle) show the curation process and examples of OmniDocLayout-1M. (Bottom) illustrates diverse layouts unconditionally generated by our OmniDocLayout-LLM.}
    \label{fig:dataset_display}
    \vspace{-7mm}
\end{figure}

A few studies have started investigating document layout generation with different generative models, ranging from early GAN-based approaches~\citep{kikuchi2021constrained} to more recent diffusion or flow-based methods~\citep{inoue2023layoutdm,chen2024towards,guerreiro2024layoutflow}. Subsequently, the rise of large language models (LLMs) has opened new possibilities~\citep{lin2023layoutprompter,tang2023layoutnuwa,shi2025layoutcot,zhang2025smaller} for conditional layout generation, drawing on their extensive prior knowledge and long-context understanding abilities.
However, a thorough review of previous studies and datasets reveals notable limitations: 

\textbf{(\emph{i}) \textit{Data Scarcity Across Diverse Document Domains.}} 
The domain bias in existing public datasets poses a critical obstacle to the development of general Document AI. Widely-used datasets such as PubLayNet~\citep{zhong2019publaynet} and DocBank~\citep{li2020docbank} offer massive annotated samples but primarily focus on a single domain—always academic articles with simple Manhattan layouts~(mostly single- or double-column). Although datasets like DocLayNet~\citep{pfitzmann2022doclaynet} and D\textsuperscript{4}LA~\citep{da2023vision} include several document types, many of these (\emph{e.g.}, letter) are no longer commonly seen in modern real-world scenarios, and their data sources are often outdated. 
Among existing resources, M\textsuperscript{6}Doc~\citep{cheng2023m6doc} and OmniDocBench~\citep{ouyang2025omnidocbench} stand out as the most valuable datasets to date, as they cover a broader spectrum of contemporary document types, even including highly complex layouts such as newspapers. Unfortunately, they contain only a limited number of samples, making them insufficient to support large-scale training~\citep{wang2025data,wen2025spot}.
Overall, the landscape of accessible document layouts exhibits a severe long-tail distribution: academic papers and outdated types are overrepresented, whereas recently published, complicated layouts such as textbooks still remain underrepresented.

\textbf{(\emph{ii}) \textit{Challenges in Complex and Long-Sequence Scenarios.}}
Due to the lack of diverse layout data, most existing approaches are restricted to simple, homogeneous academic layouts, where progress has already plateaued. 
In contrast, most real-world document layouts are more complex, with finer-grained element categories and a larger number of elements per page. Existing methods perform poorly in such scenarios, particularly in long-sequence modeling. Diffusion-based layout generation models, such as LayoutDM~\citep{inoue2023layoutdm} and LACE~\citep{chen2024towards} are particularly data-hungry and require extensive training to converge in complex domains. 
While recent LLM-based layout generation approaches, such as LayoutRAG~\citep{wu2025layoutrag}, LayoutCoT~\citep{shi2025layoutcot} and LayoutPrompter~\citep{lin2023layoutprompter}, offer promise, direct fine-tuning or in-context learning on complex domains increases learning difficulty and leads to frequent failures. 
Domain-agnostic models like LayoutNUWA~\citep{tang2023layoutnuwa} and LGGPT~\citep{zhang2025smaller} represent latest progress, but are only tested on limited document types and require substantial computational resources.

To this end, we introduce OmniDocLayout-1M, the first million-scale dataset for diverse document layout generation. \ding{182} It contains twice as many samples as DocBank, \ding{183} covers six common document types from real-world scenarios, and \ding{184} adopts a fully automated annotation pipeline, providing a powerful foundation for training layout generation models. 
Moreover, to enable diverse document layout generation under limited fine-grained annotated data, we propose a unified framework that formulates the task as a two-stage \emph{Coarse-to-Fine learning paradigm}. Specifically, we first let an LLM learn basic layout principles such as alignment and spatial organization on OmniDocLayout-1M across sufficiently diverse document types with coarse-grained labels. Then, with only a small amount of fine-grained annotated data, we perform fine-grained adaptation on a specific domain, enabling controllable and adaptable layout generation with minimal supervision and parameter footprint.
Our contribution is summarized as follows:

\begin{itemize}[leftmargin=10pt, topsep=0pt, itemsep=1pt, partopsep=1pt, parsep=1pt]
\item We introduce \textbf{OmniDocLayout-1M}, the first million-scale document layout dataset curated through an automated pipeline, comprising six commonly used document types and annotations for ten element categories.


\item We propose \textbf{OmniDocLayout-LLM}, a lightweight 0.5B model trained under our designed \textit{coarse-to-fine learning paradigm}, where aesthetic rules are first learned from diverse layouts and then adapted to a specific document domain using only a few fine-grained annotated samples.

\item  Extensive experiments across multiple document domains demonstrate that our method achieves state-of-the-art (SOTA) performance consistently. In addition, our visualization examples demonstrate alignment with both aesthetic principles and user expectations.

\end{itemize}

\section{Related Work}
\label{sec:related_work}

\noindent \textbf{Document Layout Dataset.}
RICO~\citep{deka2017rico} is widely used, but is tailored for mobile app UIs rather than document layouts.
Existing layout datasets for document domains largely originate from parsing tasks and primarily focus on academic articles. PubLayNet~\citep{zhong2019publaynet} auto-annotates papers from PubMed Central via XML matching, while DocBank~\citep{li2020docbank} uses weak supervision on arXiv. Recent efforts broaden document types but still remain limited in both scale and diversity, and often require manual annotation. Although DocLayNet~\citep{pfitzmann2022doclaynet} and D\textsuperscript{4}LA~\citep{da2023vision} cover six and twelve types, respectively, they contain only on the order of $10^4$ pages, most of which are rarely seen in contemporary real-world documents. M\textsuperscript{6}Doc~\citep{cheng2023m6doc} and OmniDocBench~\citep{ouyang2025omnidocbench} reflect real-world formats (e.g., textbooks, newspapers) but are even smaller in scale~($<$10K samples), and the latter serves purely as a benchmark without a training split.

\vspace{0.8mm}
\noindent \textbf{Layout Generation.}
Document layout generation has gained growing traction recently. Early methods usually used GANs or transformers: 
LayoutGAN++~\citep{kikuchi2021constrained} enhances the GAN framework with transformer blocks and optimizes latent codes to achieve constrained layout generation. LayoutTransformer~\citep{gupta2021layouttransformer} leverages self-attention to learn contextual relationships among layout elements, BLT~\citep{kong2022blt} adopts a non-autoregressive bidirectional transformer that iteratively refines layouts by masking and predicting low-confidence attributes through a hierarchical sampling strategy. LayoutFormer++~\citep{jiang2023layoutformer++} employs constraint tokenization and restricted decoding space to strike a balance between user constraint satisfaction and overall layout quality.  
More recently, diffusion-based methods have sparked a new wave of interest.
LayoutDM~\citep{inoue2023layoutdm} and LACE~\citep{chen2024towards}) denoise element coordinates and labels in discrete/continuous spaces, respectively, and attempt to inject hard/soft constraints as conditions. Flow-based LayoutFlow~\citep{guerreiro2024layoutflow} frames the task as flow matching, speeding both training and inference.

\vspace{0.8mm}
\noindent \textbf{Large Language Model.}
%
With the remarkable success of LLMs~\citep{wen2025ai,openai2024hello,anthropic2025claude4,gemini-2.5-pro,yang2024memorization,yang2025dtpa,kang2025legion,wen2024aidbench}, autoregressive generation has become the mainstream paradigm for document layout generation in recent years. LayoutPrompter~\citep{lin2023layoutprompter} casts layouts into unified HTML representations and employs adaptive exemplar retrieval to enable in-context learning. LayoutCoT~\citep{shi2025layoutcot} leverages the deep reasoning capabilities of LLMs through chain-of-thought prompting, substantially improving the performance and practicality. LayoutRAG~\citep{wu2025layoutrag} retrieves optimal reference layouts from a layout database and introduces a condition-modulated attention module to selectively incorporate prior knowledge. Although shown to be effective to some extent, these approaches remain largely domain-specific and rely heavily on prompt engineering or retrieval heuristics. In contrast, LayoutNUWA~\cite{tang2023layoutnuwa} and LGGPT~\citep{zhang2025smaller} pioneer domain-agnostic paradigms that fully exploit the generalization strength of LLMs: the former formulates the task as HTML code completion, whereas the latter demonstrates that pure string-based input/output reduces redundant tokens~\citep{liu2025global,wen2025stop,wen2025token,ma2025mmg} and yields better efficiency~\citep{liu2025shifting,wen2025efficient,xiong2025prune2drive}.



\section{OmniDocLayout-1M Dataset}
\label{sec:dataset}



\subsection{Motivation}
\label{sec:dataset:motivation}

Despite the rapid advances in document parsing that have given rise to a variety of document layout datasets in recent years, we observe that existing resources still suffer from several notable limitations.
\textbf{(\emph{i}) Limited Diversity.} Early document layout datasets, such as PubLayNet~\citep{zhong2019publaynet} and DocBank~\citep{li2020docbank}, are largely derived from large-scale academic paper repositories (\emph{e.g.}, PubMed, arXiv) and thus consist of single-domain pages with relatively simple Manhattan layouts.
\textbf{(\emph{ii}) Deficient Volume.} Generative tasks typically require more data than detection tasks,
particularly for diffusion-based models. However, existing diverse datasets like M\textsuperscript{6}Doc~\citep{cheng2023m6doc} and OmniDocBench~\citep{ouyang2025omnidocbench}, 
contain samples on the order of $10^{2}\sim10^{3}$ per document type, making them inadequate for training layout generation models.
\textbf{(\emph{iii}) Outdated Source.} As document layouts evolve toward improved aesthetics
, the timeliness of data sources is critical.
Although D\textsuperscript{4}LA~\citep{da2023vision} covers 12 document types, its images are sourced from RVL-CDIP~\citep{harley2015evaluation}, which contains obsolete formats (\emph{e.g.}, handwritten letters) and consists largely of noisy or skewed scans, substantially degrading its quality and value.
\textbf{(\emph{iv}) Inefficient Annotation.} Recent diverse datasets like DocLayNet~\citep{pfitzmann2022doclaynet}, often rely on labor-intensive manual annotation, which hinders scalability. With the rapid advancement of document parsing tools (\emph{e.g.}, MinerU~\citep{wang2024mineruopensourcesolutionprecise}), returning to fully automated pipelines for accurate annotation has become feasible and convenient.

\begin{table}[!t]
    \vspace{1mm}
    \caption{\textbf{Comparison with Existing Layout Datasets.} \textbf{\textcolor{red}{*}} indicates that most of the document types are outdated.}
    \vspace{-2mm}
    \resizebox{\linewidth}{!}{
        \begin{tabular}{lccccc}
            \toprule
            \textbf{Dataset} & \textbf{Volume} &  \makecell{\textbf{Element} \\ \textbf{Number}} & \makecell{\textbf{Layout} \\ \textbf{Type}} & \makecell{\textbf{Annotation} \\ \textbf{Method}} & \makecell{\textbf{Source} \\ \textbf{Count}} \\
            \hline
            DSSE-200~\citep{yang2017learning} & 200 & 2,546 & 2 & Automatic & Unknown \\
            Prima-LAD~\citep{antonacopoulos2009realistic} & 478 & 7,453 & 5 & Automatic & Unknown \\
            PubLayNet~\citep{zhong2019publaynet} & $\sim$0.36M & $\sim$3.30M & 1 & Automatic & 1 \\
            DocBank~\citep{li2020docbank} & $\sim$0.50M & $\sim$6.70M & 1 & Automatic & 1 \\
            DocLayNet~\citep{pfitzmann2022doclaynet} & $\sim$0.08M & $\sim$1.10M & 6 & Manual & Unknown \\
            D\textsuperscript{4}LA~\citep{da2023vision} & $\sim$0.01M & $\sim$0.30M & 12\textbf{\textcolor{red}{*}} & Manual & 1 \\
            M\textsuperscript{6}Doc~\citep{cheng2023m6doc} & $<$0.01M & $\sim$0.24M & 7 & Manual & $\geq 3$ \\
            \rowcolor{LightGreen}
            \textbf{OmniDocLayout-1M~(Ours)} & \textbf{$\sim$1.00M} & \textbf{$\sim$48.0M} & \textbf{6} & \textbf{Automatic} & \textbf{36} \\
            \bottomrule
        \end{tabular}
    }
    \label{tab:dataset_compare}
    \vspace{-6mm}
\end{table}

\subsection{Dataset Construction}
To address the limitations outlined in Section~\ref{sec:dataset:motivation}, we present OmniDocLayout-1M, the first million-scale layout dataset tailored for document domains, featuring diverse and complicated document types, up-to-date data, and a fully automated annotation pipeline. 

\noindent \textbf{Preprocessing.} 
%
To ensure both diversity and timeliness, we collect documents for OmniDocLayout-1M from massive sources on the Internet. During the preprocessing stage, we use format standardization techniques to handle different document formats including PDF, Markdown, etc. Meanwhile, procedures such as deduplication and quality analysis are applied to filter out noisy pages and ensure high quality. Finally, we collect data from 36 public and copyright-clean sources in total, including Academic Databases (13 sources), Publishers (7 sources), and Document-sharing Platforms (16 sources), covering various fields, such as academia, education, news, economics and etc. 

\begin{figure*}[!t]
    \vspace{-3mm}
    \centering
    
    \begin{subfigure}[b]{0.26\textwidth}
        \centering
        \includegraphics[width=\linewidth]{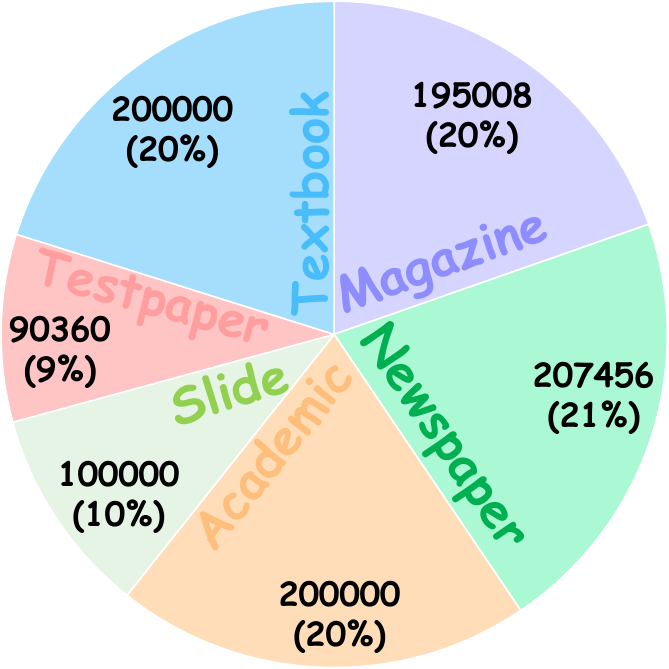}
        \caption{Document types distribution.}
        \label{fig:document_type}
    \end{subfigure}
    \hfill
    \begin{subfigure}[b]{0.335\textwidth}
        \centering
        \includegraphics[width=\linewidth]{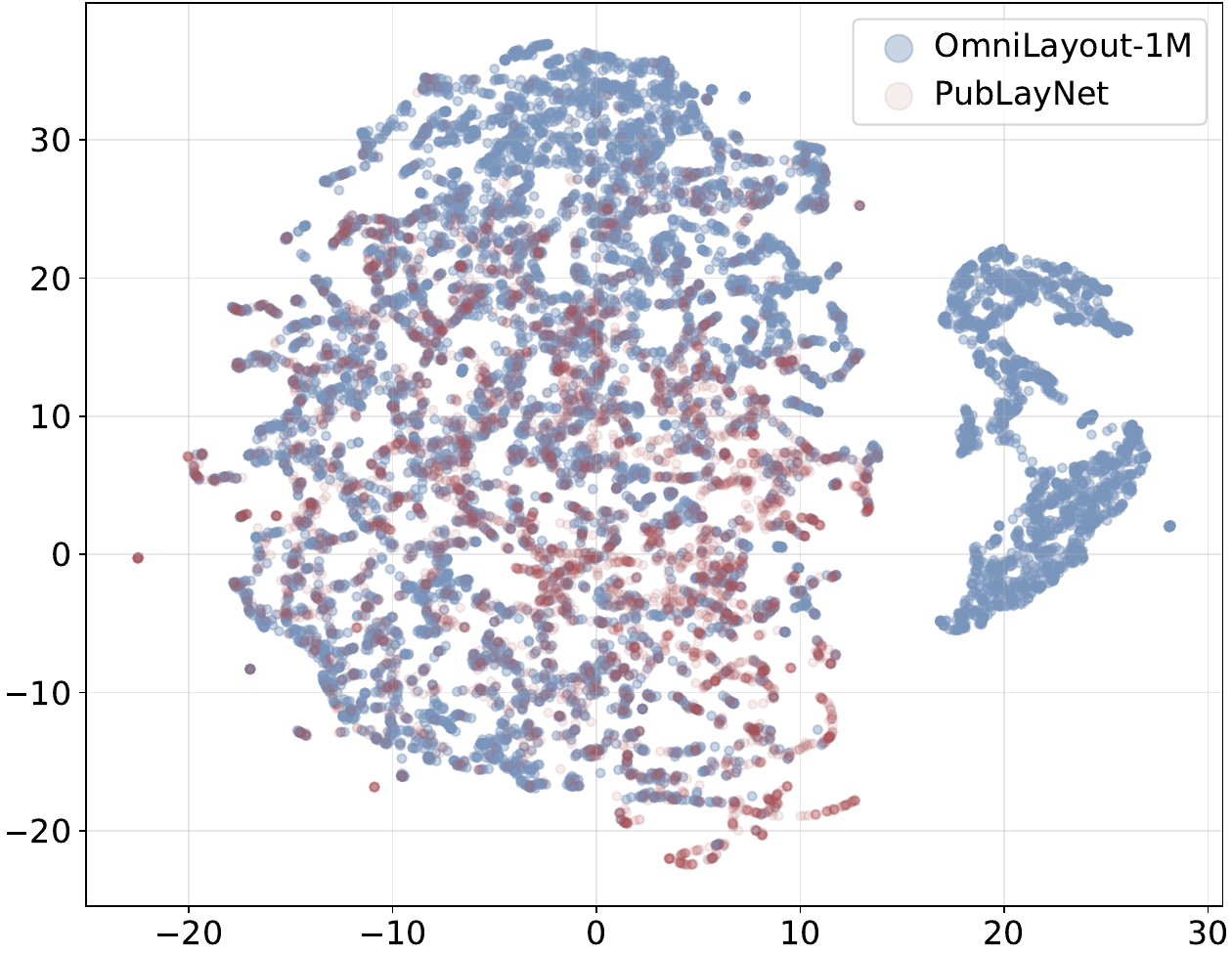}
        \caption{Hand-crafted feature distribution}
        \label{fig:feature_distribution}
    \end{subfigure}
    \hfill
    \begin{subfigure}[b]{0.315\textwidth}
        \centering
        \includegraphics[width=\linewidth]{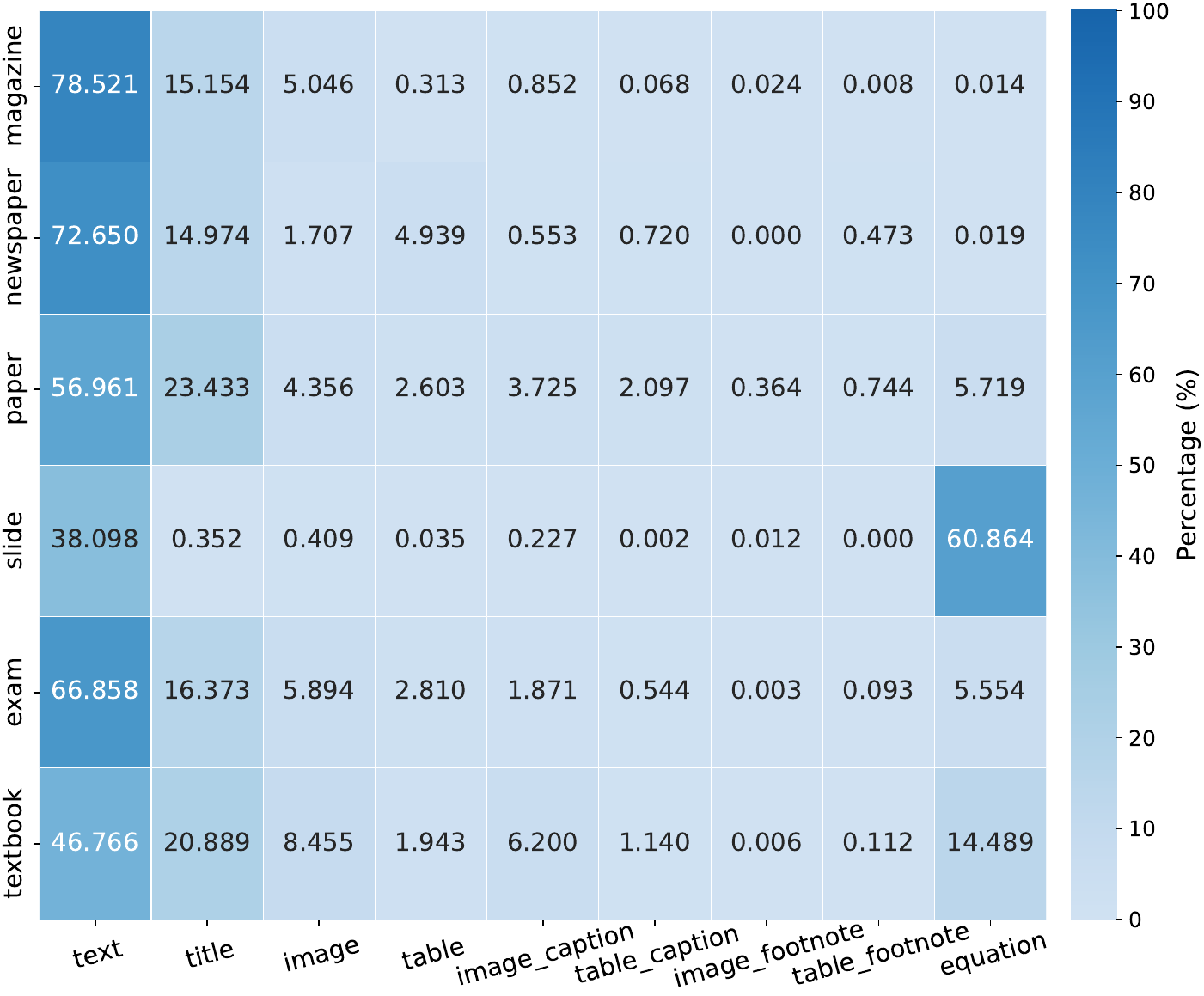}
        \caption{Co-occurrence heatmap}
        \label{fig:cooccurrence}
    \end{subfigure}
    \vspace{-0.4em}
    \caption{\textbf{Statistical Analysis of OmniDocLayout-1M.} (a) \& (b) show the multi-dimensional diversity, (c) proves its consistency with prior knowledge.}
    \vspace{-1em}
\end{figure*}

\noindent \textbf{Annotation.} 
To accurately convert document pages into corresponding element sequences, we employ MinerU~\citep{wang2024mineruopensourcesolutionprecise}, a powerful open-source toolkit, to automatically annotate the samples. Furthermore, MinerU outputs the element sequence more aligned with the natural reading order, a property essential for reliable and coherent layout generation, which is not considered and provided by any other existing datasets.
Notably, for newspapers whose layouts are too complicated to be directly processed by MinerU, we manually annotate 1,000 in-domain newspaper layouts and fine-tune a DocLayout-YOLO~\citep{zhao2024doclayout} to produce better performance, especially in capturing dense and irregular layouts.

\noindent\textbf{Quality Assessment.} To validate the reliability of model-generated bounding-box annotations, we conduct a blind human study on 1,200 OmniDocLayout-1M pages, comparing manually labeled layouts with model outputs. Annotators report over $\geq 92\%$ similar perceived quality between the two. Please refer to Appendix for more details.

\vspace{-2mm}
\subsection{Dataset Statistical Analysis}

\noindent \textbf{Comparisons with existing datasets.} 
Table~\ref{tab:dataset_compare} highlights OmniDocLayout-1M's advantages over existing datasets. In terms of diversity, our dataset significantly surpasses existing datasets in the number of document elements~($\sim$48M instances), latest document types~(6 domains), data volume~($\sim$1M samples) and data source diversity~(36 sources). The comprehensive diversity of OmniDocLayout-1M effectively meets the demand for generating realistic and various document layouts in the modern world.

\noindent \textbf{Document type distribution.}
The distribution of different document types is shown in Fig.~\ref{fig:document_type}. OmniDocLayout-1M encompasses six common layout types: textbook, newspaper, magazine, exam, academic, and slide. Data is specifically balanced across all layout types to prevent domain bias and ensure fair training.

\noindent \textbf{Hand-crafted feature distribution.}
OmniDocLayout-1M exhibits significantly greater layout diversity than the simple and homogeneous distribution of academic papers in PubLayNet, as evidenced by the UMAP~\citep{mcinnes2018umap} visualization in Fig.~\ref{fig:feature_distribution}. We select hand-crafted features that best capture layout diversity and complexity, such as the number of elements per page, page-level filling ratio, and element centroids distribution for visualization.

\noindent \textbf{Element co-occurrence analysis.}
To validate the plausibility of OmniDocLayout-1M, Fig.~\ref{fig:cooccurrence} visualizes element co-occurrence patterns. The distributions align with expectations: text and title are most frequent across all document types, followed by image and table, while formula is prominent in academic content such as textbook, paper, and exucational slide. These observations confirm the OmniDocLayout-1M's adherence to real-world principles.
\section{Methodology}
\label{sec:method}


\subsection{Problem Formulation}
Following previous work~\citep{inoue2023layoutdm,guerreiro2024layoutflow,zhang2025smaller}, a document layout is represented as a set of 5-tuples:
\begin{equation}
\mathcal{L} = \{\, e_i = (c, x, y, w, h) \mid i = 1, \ldots, N \,\},
\end{equation}
\vspace{-4mm}

\noindent where $c$ denotes the element category, $x,y$ are the horizontal and vertical coordinates of the bounding box (either the top-left corner or the center point), and $w,h$ are its width and height, respectively.

As stated in our contributions, unlike all prior approaches, we formulate complex layout generation task as a two-stage \textit{Coarse-to-Fine learning paradigm}, which enables enable the model to learn complex layout logic from easy to hard. 
This is motivated by our observation that, although different document types exhibit highly diverse layout styles, they share fundamental aesthetic principles, such as maintaining alignment and avoiding overlap.
In our implementation, as shown in Eq.~\ref{eq:2}, we let $\displaystyle \mathbb{D}_{\mathrm{coar}}=\{ D^{(m)}_{\mathrm{coar}} \}_{m=1}^{M}\vphantom{\big|}$ be a \textbf{diverse collection} of document types with a \textbf{coarse-grained} label set $\displaystyle \mathbb{C}_{\mathrm{coar}}\vphantom{\big|}$, and let $\displaystyle D_{\mathrm{fine}}$ be the \textbf{specific complex} domain with a \textbf{fine-grained} label set $\displaystyle \mathbb{C}_{\mathrm{fine}}\vphantom{\big|}$.
We first perform Stage 1 on the diverse data of OmniDocLayout-1M to acquire basic layout principles in spatial organization, and then conduct Stage 2 on a fine-grained annotated dataset (\emph{e.g.}, M\textsuperscript{6}Doc) to adapt to the target domain with complex element categories. 
We find that this paradigm significantly reduces learning difficulty, enabling Stage 2 to require only a small number of diverse, fine-grained annotated samples (typically a few hundred) for effective adaptation, thus overcoming both the scarcity of fine-grained annotation and the difficulty of directly learning complex layouts.

\vspace{-5mm}

\begin{equation}\label{eq:2}
\resizebox{0.9\linewidth}{!}{$
\underbrace{\Big(\mathbb{D}_{\mathrm{coar}},\;
\mathbb{C}_{\mathrm{coar}}\Big)}_{\substack{\text{Stage 1: Coarse Learning [EASY]}\\ \text{(Diverse Domains, Coarse-grained Labels)}}}
\;\xRightarrow[\scriptstyle \mathrm{Transfer}]{}\;
\underbrace{\Big(D_{\mathrm{fine}},\;
\mathbb{C}_{\mathrm{fine}}\Big)}_{\substack{\text{Stage 2: Fine Adaptation [HARD]}\\ \text{(Specific Domain, Fine-grained Labels)}}}
$}
\end{equation}

\subsection{Layout Generation Modeling}
We cast layout generation as conditional sequence modeling over a unified token space that encodes both semantic categories and normalized bounding boxes. Given a layout $\mathcal{L}$ serialized into a sequence of discrete tokens $T = (t_1, t_2, \ldots, t_K)$, the model is trained to maximize the conditional log-likelihood of this sequence.

\begin{figure*}[!t]
    \centering
    \includegraphics[width=1\textwidth]{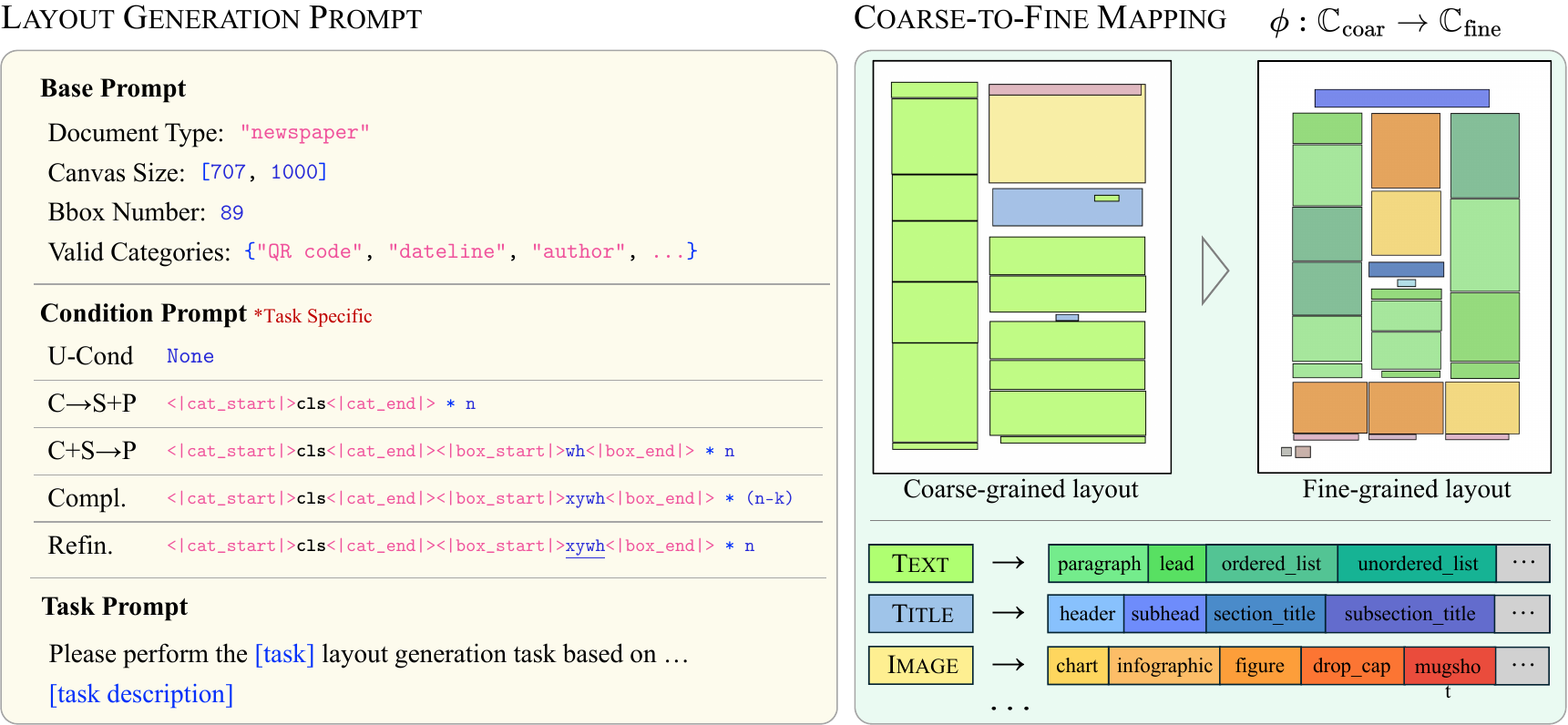}
    \vspace{-1.6em}
    \caption{\textbf{Overview of our layout generation framework~(OmniDocLayout-LLM).}
Left: The unified layout prompt consists of a \emph{Base Prompt} (document metadata), a \emph{Condition Prompt} for U-Cond, C$\rightarrow$S+P, C+S$\rightarrow$P, Completion, and Refinement, and a \emph{Task Prompt} defining the layout objective.
Right: A \emph{Coarse-to-Fine Mapping} $\phi : \mathcal{C}_{\text{coar}} \rightarrow \mathcal{C}_{\text{fine}}$ transfers coarse layout categories into fine-grained domain-specific labels.}
    \label{fig:main}
    \vspace{-5mm}
\end{figure*}

\vspace{0.75mm}
\noindent \textbf{Layout Generation Tasks.}
We follow the task setting introduced in~\citep{zhang2025smaller}, and use five conditioning regimes that factor layout generation into category ($C$), size ($S$), and position ($P$), enabling controllable synthesis, constrained placement, completion, and editing:
(\underline{1}) \textbf{U-Cond}: Unconditional generation without external constraints.
(\underline{2}) \textbf{C\textrightarrow S+P}: Given the category of each element, the model predicts both its size and position.
(\underline{3}) \textbf{C+S\textrightarrow P}: The position of each element is masked; the model infers it from the provided category and size.
(\underline{4}) \textbf{Completion}: A subset ($0\text{--}20\%$) of elements is retained on the page; the model completes the remaining layout to form a coherent structure.
(\underline{5}) \textbf{Refinement}: Geometric attributes are perturbed by Gaussian noise $\mathcal{N}(0,10^{-2})$; the model recovers the original layout.

\noindent \textbf{Layout Representation \& Generation Prompt.}
Some recent studies~\citep{lin2023layoutprompter,tang2023layoutnuwa} argue that LLMs are extensively exposed to web data during pre-training, and therefore adopting HTML-style prompts may better activate their layout generation capability. However, empirical evidence from LGGPT~\citep{zhang2025smaller} demonstrates that HTML-based formats contain a large amount of meaningless structural descriptors, such as \texttt{<body>} and \texttt{</body>}, which convey little informative content while severely slowing down both training and inference. Moreover, its performance is shown to be inferior compared to pure string-based modeling. Therefore, following LGGPT, each element $e_i=(c,x,y,w,h)$ is serialized with a string-based prefix-aware encoding
{\ttfamily\small
\texttt{<|cat\_start|>}\allowbreak $c$\allowbreak
\texttt{<|cat\_end|><|box\_start|>}\allowbreak
\texttt{0}$x$~\allowbreak
\texttt{1}$y$~\allowbreak
\texttt{2}$w$~\allowbreak
\texttt{3}$h$\allowbreak
\texttt{<|box\_end|>}
},
where coordinates $x, y, w,\\ h$ are normalized and uniformly quantized to $[0,999]$. 

As shown in Fig.~\ref{fig:main}, the page-level \emph{layout generation prompt} concatenates three components:
(\emph{i}) \textbf{Base Prompt}: a base header that specifies the document type, canvas size, total number of bounding boxes, and the set of valid semantic categories;
(\emph{ii}) \textbf{Condition Prompt}: a task-specific conditioning sequence that provide either no input (U-Cond), category-only tokens (C→S+P), category plus size tokens (C+S→P), partial tuples for Completion, or fully perturbed tuples for Refinement;
and (\emph{iii}) \textbf{Task Prompt}: an instruction describing the desired layout generation objective.

\renewcommand{\multirowsetup}{\centering}
\definecolor{mygray}{gray}{.92}
\definecolor{ForestGreen}{RGB}{34,139,34}
\definecolor{Forestred}{RGB}{220,50,50}
\newcommand{\fr}[1]{\mathbf{\mathcolor{Forestred}{#1}}}
\begin{table*}[!t]
    \centering
    \vspace{-3mm}
    \belowrulesep=-0.25pt
    \aboverulesep=-0.25pt
    \setlength{\tabcolsep}{3pt}
    \renewcommand{\arraystretch}{1.35}
    \footnotesize
	\centering
    \caption{\textbf{Comparison with Layout Generation Experts across Five Document Types in M\textsuperscript{6}Doc.} For metrics, Ali. and Ove. denote \textit{Alignment} and \textit{Overlap}, $\rightarrow$ means closer to ground truth is better. For tasks, Compl. and Refin. denote Completion and Refinement. ``-'' indicates not applicable.}
     \vspace{-0.6em}
    \resizebox{\textwidth}{!}{
        \begin{tabular}{
          >{\centering\arraybackslash}p{1cm} 
          >{\centering\arraybackslash}p{2.5cm} 
          *{20}{>{\centering\arraybackslash}p{0.9cm}}
        }
            \toprule[1.2pt]
            \multirow{2}{*}{\textbf{Task}} & \multirow{2}{*}{\textbf{Method}} & \multicolumn{4}{c}{\textbf{Textbook}} & \multicolumn{4}{c}{\textbf{Newspaper}} & \multicolumn{4}{c}{\textbf{Magazine}} & \multicolumn{4}{c}{\textbf{Exam}} & \multicolumn{4}{c}{\textbf{Academic}} \\
            \cmidrule(lr){3-6} \cmidrule(lr){7-10} \cmidrule(lr){11-14} \cmidrule(lr){15-18} \cmidrule(lr){19-22}
            & & 
            FID$\downarrow$ & Ali.$\rightarrow$ & Ove.$\rightarrow$ & mIoU$\uparrow$ &
            FID$\downarrow$ & Ali.$\rightarrow$ & Ove.$\rightarrow$ & mIoU$\uparrow$ &
            FID$\downarrow$ & Ali.$\rightarrow$ & Ove.$\rightarrow$ & mIoU$\uparrow$ &
            FID$\downarrow$ & Ali.$\rightarrow$ & Ove.$\rightarrow$ & mIoU$\uparrow$ &
            FID$\downarrow$ & Ali.$\rightarrow$ & Ove.$\rightarrow$ & mIoU$\uparrow$ \\

            \hline
            \multirow{5}{*}{\textbf{U-Cond}} & LayoutDM &  \underline{180.25} & \underline{0.024} & \underline{0.310} & - & 281.56 & \underline{0.008} & 0.628 & - & 281.91 & 0.233 & \underline{0.462} & - & 287.58 & 0.131 & \underline{0.382} & - & \underline{153.66} & 0.440 & 0.362 & - \\

            & LACE &  251.41 & 0.001 & 3.206 & - & 423.21 & 0.001 & 4.982 & - & 325.67 & \underline{0.001} & 6.789 & - & 325.45 & \underline{0.002} & 3.602 & - & 276.05 & \underline{0.001} & 9.980 & - \\

            & LayoutPrompter & - & - & - & - & - & - & - & - & - & - & - & - & - & - & - & - & - & - & - & - \\

            & LGGPT & 197.81 & 0.980 & 1.049 & 0.000 & \underline{154.20} & 2.591 & \underline{0.350} & 0.000 & \underline{162.94} & 3.190 & 0.813 & 0.038 & \underline{157.11} & 1.324 & 0.135 & 0.047 & 236.72 & 0.533 & \textbf{0.100} & 0.000 \\

            \rowcolor{LightGreen}
            & Ours & \textbf{40.28} & \textbf{0.219} & \textbf{0.102} & \textbf{0.288} & \textbf{39.73} & \textbf{0.015} & \textbf{0.084} & 0.000 & \textbf{41.82} & \textbf{0.089} & \textbf{0.151} & \textbf{0.266} & \textbf{40.32} & \textbf{0.072} & \textbf{0.182} & \textbf{0.236} & \textbf{36.48} & \textbf{0.089} & \underline{0.062} & \textbf{0.415} \\
            \hline

            \multirow{5}{*}{\textbf{C$\rightarrow$S+P}} & LayoutDM & 178.24 & 0.520 & 0.246 & 0.041 & 288.98 & \underline{0.010} & 0.582 & 0.091 & 271.52 & 0.401 & 0.453 & 0.081 & 296.86 & \underline{0.171} & 0.365 & 0.043 & 141.28 & 1.458 & 0.422 & 0.068 \\

            & LACE & 187.31 & 0.005 & 2.345 & 0.025 & 308.24 & 0.009 & 2.873 & 0.000 & 220.00 & \underline{0.001} & 0.862 & 0.069 & 276.12 & 0.010 & 0.442 & 0.009 & 212.41 & \underline{0.006} & 2.645 & 0.088 \\

            & LayoutPrompter & 44.67 & \underline{0.512} & \underline{0.191} & \textbf{0.166} & \underline{124.45} & 0.312 & 0.899 & \underline{0.160} & \underline{65.24} & 0.899 & \underline{0.362} & \textbf{0.224} & \underline{46.56} & 0.262 & 0.439 & \underline{0.098} & \underline{20.89} & 0.221 & \underline{0.264} & \underline{0.216} \\

            & LGGPT & \underline{177.91} & 0.990 & 0.463 & 0.064 & 167.39 & 2.731 & \underline{0.444} & 0.000 & 172.45 & 3.161 & 1.133 & 0.026 & 186.38 & 1.392 & \underline{0.229} & 0.032 & 244.44 & 0.710 & 3.182 & 0.000 \\

            \rowcolor{LightGreen}
            & Ours & \textbf{18.38} & \textbf{0.228} & \textbf{0.121} & \underline{0.154} & \textbf{10.71} & \textbf{0.014} & \textbf{0.086} & \textbf{0.185} & \textbf{21.08} & \textbf{0.092} & \textbf{0.138} & \underline{0.221} & \textbf{8.68} & \textbf{0.074} & \textbf{0.241} & \textbf{0.121} & \textbf{16.84} & \textbf{0.084} & \textbf{0.070} & \textbf{0.246} \\
            \hline

            \multirow{5}{*}{\textbf{C+S$\rightarrow$P}} & LayoutDM & 174.82 & 0.471 & \underline{0.452} & 0.093 & 285.43 & \textbf{0.010} & 0.679 & 0.135 & 172.01 & 0.441 & 0.537 & 0.136 & 144.29 & 0.162 & 0.468 & 0.066 & 76.72 & 1.135 & 0.479 & 0.118 \\

            & LACE & 28.79 & 0.001 & 6.345 & 0.015 & 256.08 & \underline{0.005} & 4.158 & 0.006 & 196.78 & \underline{0.002} & 6.015 & 0.048 & 160.28 & \underline{0.008} & 6.327 & 0.050 & 99.86 & 0.008 & 1.402 & 0.097 \\

            & LayoutPrompter & \underline{42.38} & \textbf{0.224} & 0.469 & \underline{0.199} & \underline{126.78} & 0.219 & 1.387 & \underline{0.156} & \underline{41.52} & 0.245 & \underline{0.442} & \underline{0.235} & \underline{14.58} & \textbf{0.042} & \underline{0.341} & \underline{0.138} & \underline{14.58} & 0.203 & \underline{0.332} & \underline{0.286} \\

            & LGGPT & 181.61 & 0.940 & 0.587 & 0.000 & 185.72 & 2.930 & \underline{0.402} & 0.000 & 169.95 & 3.297 & 1.102 & 0.038 & 180.76 & 1.441 & 0.189 & 0.026 & 244.67 & 0.561 & 3.151 & 0.000 \\

            \rowcolor{LightGreen}
            & Ours & \textbf{16.92} & \underline{0.366} & \textbf{0.122} & \textbf{0.219} & \textbf{6.13} & 0.021 & \textbf{0.188} & \textbf{0.240} & \textbf{20.74} & \textbf{0.130} & \textbf{0.174} & \textbf{0.256} & \textbf{5.42} & \underline{0.083} & \textbf{0.235} & \textbf{0.200} & \textbf{9.02} & \textbf{0.162} & \textbf{0.085} & \textbf{0.360} \\
            \hline

            \multirow{5}{*}{\textbf{Compl.}} & LayoutDM & 172.35 & 0.012 & 0.429 & 0.000 & 270.15 & \underline{0.007} & 0.704 & 0.000 & 260.15 & 0.113 & 0.557 & 0.000 & 255.17 & 0.073 & 0.459 & 0.000 & 134.51 & 0.370 & 0.418 & 0.000 \\

            & LACE & 268.36 & \underline{0.185} & \underline{0.158} & 0.000 & 432.76 & 0.034 & 2.865 & 0.000 & 316.32 & \underline{0.043} & 0.342 & 0.000 & 332.15 & 0.071 & \textbf{0.218} & 0.000 & 284.16 & \underline{0.107} & 0.768 & 0.000 \\

            & LayoutPrompter & \underline{46.76} & 0.491 & 0.244 & \underline{0.169} & \underline{86.99} & 0.357 & 0.481 & 0.000 & \underline{39.02} & 0.676 & \underline{0.234} & \textbf{0.342} & \underline{32.83} & \textbf{0.066} & \underline{0.321} & \underline{0.109} & \underline{32.24} & 0.168 & \underline{0.287} & \textbf{0.642} \\

            & LGGPT & 192.32 & 1.180 & 0.892 & 0.158 & 160.25 & 2.696 & \underline{0.335} & 0.000 & 153.43 & 3.511 & 0.743 & 0.052 & 153.79 & 1.461 & 0.167 & 0.000 & 242.17 & 0.975 & 2.812 & 0.000 \\

            \rowcolor{LightGreen}
            & Ours & \textbf{31.58} & \textbf{0.235} & \textbf{0.123} & \textbf{0.478} & \textbf{22.48} & \textbf{0.013} & \textbf{0.098} & 0.000 & \textbf{38.56} & \textbf{0.098} & \textbf{0.153} & \underline{0.288} & \textbf{25.92} & \underline{0.068} & 0.203 & \textbf{0.310} & \textbf{30.56} & \textbf{0.106} & \textbf{0.070} & \underline{0.620} \\
            \hline

            \multirow{5}{*}{\textbf{Refin.}} & LayoutDM & 124.85 & 0.521 & \underline{0.269} & 0.123 & 264.69 & 0.010 & \underline{0.624} & 0.142 & 203.91 & 0.361 & 0.431 & 0.163 & 228.59 & 0.138 & \underline{0.410} & 0.090 & 63.15 & 1.212 & \underline{0.411} & 0.132 \\

            & LACE & 143.95 & \underline{0.174} & 0.736 & 0.165 & 291.35 & \textbf{0.013} & 2.047 & 0.234 & 216.23 & \underline{0.141} & 0.693 & 0.256 & 214.19 & \textbf{0.048} & 0.921 & 0.187 & 42.89 & 0.219 & 0.621 & 0.254 \\

            & LayoutPrompter & \underline{11.82} & 0.149 & 0.511 & \underline{0.672} & \underline{113.84} & 0.154 & 1.216 & \underline{0.647} & \underline{10.26} & 0.919 & \underline{0.430} & \underline{0.689} & \underline{10.87} & \underline{0.072} & 0.422 & \underline{0.502} & \underline{9.23} & \underline{0.177} & 0.437 & \underline{0.667} \\

            & LGGPT & 217.05 & 1.031 & 1.282 & 0.010 & 220.33 & 3.675 & 0.695 & 0.000 & 145.76 & 3.665 & 0.662 & 0.024 & 215.24 & 2.420 & 0.578 & 0.000 & 246.62 & 0.408 & 3.342 & 0.010 \\

            \rowcolor{LightGreen}
            & Ours & \textbf{4.51} & \textbf{0.317} & \textbf{0.145} & \textbf{0.681} & \textbf{10.60} & \underline{0.017} & \textbf{0.064} & \textbf{0.732} & \textbf{4.73} & \textbf{0.113} & \textbf{0.072} & \textbf{0.752} & \textbf{6.66} & 0.079 & \textbf{0.295} & \textbf{0.641} & \textbf{8.25} & \textbf{0.132} & \textbf{0.138} & \textbf{0.708} \\
            \hline

            \multicolumn{2}{c}{\textbf{Test Data}} & - & 0.289 & 0.010 & - & - & 0.012 & 0.051 & - & - & 0.083 & 0.054 & - & - & 0.062 & 0.266 & - & - & 0.097 & 0.126 \\

            \bottomrule[1.2pt]
    	\end{tabular}
    }
    \label{tab:expert_comparision}
     \vspace{-1.8em}
\end{table*}

\subsection{Coarse-to-Fine Learning}
The above formulation defines layout representation and conditioning regimes. A key challenge is how to train models that generalize across diverse document types and complex element categories. Directly learning fine-grained structures 
from limited data risks overfitting and poor transfer. We therefore adopt a \emph{Coarse-to-Fine learning paradigm}, where our OmniDocLayout-LLM first acquires robust spatial priors and structural regularities from diverse domains with coarse-grained labels, and then adapts to specific domains with fine-grained supervision. This staged strategy allows the model to progress from easy to hard, aligning training objectives with increasing complexity.

\vspace{-0.3mm}
\noindent \textbf{Coarse-grained Learning.}
The coarse-grained pre-training stage aims to establish a strong foundation for layout generation by harnessing the diversity of pre-training domains $\mathbb D_{\mathrm{coar}}$. At this stage, the model is exposed to a wide range of document types, enabling it to acquire a broad understanding of document structures and the spatial relationships among various layout elements. Central to our approach is the unified representation of layout elements and the harmonization of label spaces. To promote generalization across domains, we employ a coarse-grained label set $\mathbb C_{\mathrm{coar}}$ that covers essential document components, such as text, table, image, and title, as well as associated classes like caption and footnote. This unified labeling strategy ensures the model learns transferable structural priors applicable to diverse document layouts.


\noindent \textbf{Fine-grained Adaptation.}
Given a target domain $D_{\mathrm{fine}}$ with fine-grained labels $\mathbb{C}_{\mathrm{fine}}$, 
we adapt the foundation model under a supervised sequence-modeling objective. 
The adaptation relies on a label mapping $\phi:\mathbb{C}_{\mathrm{coar}} \rightarrow \mathbb{C}_{\mathrm{fine}}$, 
where each coarse class is expanded into its fine-grained descendants (\emph{e.g.}, ``text'' $\mapsto$ \{``paragraph'', ``lead'', ``ordered\_list''\}). 
We fine-tune models for heterogeneous targets containing multiple document types (\emph{e.g.}, \textsc{newspaper}, \textsc{exam}, \textsc{academic}), 
yielding sharper, type-aware categories while preserving the structural priors and cross-type generalization acquired during coarse-grained pretraining.

\section{Experiments}
\label{sec:experiment}

\subsection{Experimental Setup}

\renewcommand{\multirowsetup}{\centering}
\definecolor{mygray}{gray}{.92}
\definecolor{ForestGreen}{RGB}{34,139,34}
\definecolor{Forestred}{RGB}{220,50,50}
\begin{table*}[!t]
    \centering
    \vspace{-2mm}
    \belowrulesep=-0.25pt
    \aboverulesep=-0.25pt
    \setlength{\tabcolsep}{3pt}
    \renewcommand{\arraystretch}{1.35}
    \footnotesize
	\centering
    \caption{\textbf{Comparison with Powerful General-purpose LLMs in 0-shot Setting across Five Document Types in M\textsuperscript{6}Doc.} For models, Gemini-2.5* and Claude-3.7* denote Gemini-2.5-Flash and Claude-3.7-Sonnet.}
     \vspace{-2mm}
    \resizebox{\textwidth}{!}{
        \begin{tabular}{
          >{\centering\arraybackslash}p{1cm} 
          >{\centering\arraybackslash}p{2.5cm} 
          *{20}{>{\centering\arraybackslash}p{0.9cm}}
        }
            \toprule[1.2pt]
            \multirow{2}{*}{\textbf{Task}} & \multirow{2}{*}{\textbf{Method}} & \multicolumn{4}{c}{\textbf{Textbook}} & \multicolumn{4}{c}{\textbf{Newspaper}} & \multicolumn{4}{c}{\textbf{Magazine}} & \multicolumn{4}{c}{\textbf{Exam}} & \multicolumn{4}{c}{\textbf{Academic}} \\
            \cmidrule(lr){3-6} \cmidrule(lr){7-10} \cmidrule(lr){11-14} \cmidrule(lr){15-18} \cmidrule(lr){19-22}
            & & 
            FID$\downarrow$ & Ali.$\rightarrow$ & Ove.$\rightarrow$ & mIoU$\uparrow$ &
            FID$\downarrow$ & Ali.$\rightarrow$ & Ove.$\rightarrow$ & mIoU$\uparrow$ &
            FID$\downarrow$ & Ali.$\rightarrow$ & Ove.$\rightarrow$ & mIoU$\uparrow$ &
            FID$\downarrow$ & Ali.$\rightarrow$ & Ove.$\rightarrow$ & mIoU$\uparrow$ &
            FID$\downarrow$ & Ali.$\rightarrow$ & Ove.$\rightarrow$ & mIoU$\uparrow$ \\

            \hline
            \multirow{4}{*}{\textbf{U-Cond}} & GPT-4o & 135.32 & 0.017 & \textbf{0.007} & 0.060 & 193.13 & \underline{0.020} & 0.007 & 0.000 & 236.11 & 0.015 & \textbf{0.040} & 0.089 & 163.94 & 0.020 & 0.010 & 0.000 & 135.60 & 0.006 & 0.006 & 0.000 \\

            & Gemini-2.5* & 147.88 & \textbf{0.264} & 6.490 & 0.154 & 194.77 & 0.078 & 0.098 & 0.000 & \underline{118.78} & \underline{0.041} & 0.213 & \underline{0.226} & 140.48 & \textbf{0.071} & \textbf{0.313} & 0.000 & \underline{57.36} & 0.347 & \underline{0.089} & 0.000 \\

            & Claude-3.7* & \underline{96.23} & 0.102 & 0.145 & \underline{0.236} & \underline{171.01} & 0.079 & \textbf{0.031} & 0.000 & 165.76 & 0.030 & 0.182 & 0.000 & \underline{114.90} & 0.014 & 0.038 & 0.000 & 106.98 & \underline{0.030} & \textbf{0.101} & 0.000 \\

            \rowcolor{LightGreen}
            & Ours & \textbf{40.28} & \underline{0.219} & \underline{0.102} & \textbf{0.288} & \textbf{39.73} & \textbf{0.015} & \underline{0.084} & 0.000 & \textbf{41.82} & \textbf{0.089} & \underline{0.151} & \textbf{0.266} & \textbf{40.32} & \underline{0.072} & \underline{0.182} & \textbf{0.236} & \textbf{36.48} & \textbf{0.089} & 0.062 & \textbf{0.415} \\
            \hline

            \multirow{4}{*}{\textbf{C$\rightarrow$S+P}} & GPT-4o & 103.15 & 0.084 & 0.072 & 0.119 & 202.84 & \underline{0.002} & \underline{0.112} & 0.028 & 165.36 & 0.055 & 0.164 & 0.097 & 107.17 & 0.040 & 0.049 & 0.082 & 90.67 & 0.224 & 0.027 & 0.123 \\

            & Gemini-2.5* & 54.74 & \underline{0.175} & \textbf{0.060} & 0.078 & 69.40 & 0.569 & 0.666 & 0.070 & \underline{53.32} & 0.065 & \textbf{0.104} & 0.101 & 42.25 & \textbf{0.053} & 0.063 & 0.036 & \underline{31.79} & 0.351 & 0.051 & 0.084 \\

            & Claude-3.7* & \underline{42.99} & 0.117 & \underline{0.068} & \underline{0.127} & \underline{53.62} & 0.001 & 0.520 & \underline{0.079} & 87.00 & \underline{0.071} & \underline{0.109} & \underline{0.126} & \underline{27.96} & 0.041 & \underline{0.080} & \underline{0.087} & 66.22 & \underline{0.138} & \textbf{0.075} & \underline{0.139} \\

            \rowcolor{LightGreen}
            & Ours & \textbf{18.38} & \textbf{0.228} & 0.121 & \textbf{0.154} & \textbf{10.71} & \textbf{0.014} & \textbf{0.086} & \textbf{0.185} & \textbf{21.08} & \textbf{0.092} & 0.138 & \textbf{0.221} & \textbf{8.68} & \underline{0.074} & \textbf{0.241} & \textbf{0.121} & \textbf{16.84} & \textbf{0.084} & \underline{0.070} & \textbf{0.246} \\
            \hline

            \multirow{4}{*}{\textbf{C+S$\rightarrow$P}} & GPT-4o & 64.67 & 0.448 & 0.363 & 0.091 & 106.97 & 0.043 & 4.759 & 0.052 & 112.38 & 0.332 & 0.765 & 0.076 & 61.67 & 0.187 & 0.905 & 0.049 & 58.49 & 0.743 & 0.852 & 0.075 \\

            & Gemini-2.5* & 139.01 & 1.103 & 0.751 & 0.057 & 117.93 & 0.034 & 6.159 & 0.039 & 110.78 & 0.259 & 0.969 & 0.085 & 43.38 & 0.138 & 0.937 & 0.050 & 62.75 & 0.994 & 0.788 & 0.063 \\

            & Claude-3.7* & \underline{26.86} & \underline{0.147} & \textbf{0.103} & \underline{0.136} & \underline{30.80} & \underline{0.002} & \underline{0.300} & \underline{0.127} & \underline{39.05} & \textbf{0.086} & \underline{0.247} & \underline{0.160} & \underline{12.69} & \textbf{0.054} & \underline{0.170} & \underline{0.096} & \underline{26.47} & \underline{0.236} & \textbf{0.116} & \underline{0.161} \\

            \rowcolor{LightGreen}
            & Ours & \textbf{16.92} & \textbf{0.366} & \underline{0.122} & \textbf{0.219} & \textbf{6.13} & \textbf{0.021} & \textbf{0.188} & \textbf{0.240} & \textbf{20.74} & \underline{0.130} & \textbf{0.174} & \textbf{0.256} & \textbf{5.42} & \underline{0.083} & \textbf{0.235} & \textbf{0.200} & \textbf{9.02} & \textbf{0.162} & \underline{0.085} & \textbf{0.360} \\
            \hline

            \multirow{5}{*}{\textbf{Compl.}} & GPT-4o & 61.20 & \textbf{0.240} & \textbf{0.051} & \textbf{0.522} & 97.60 & 0.227 & \textbf{0.057} & 0.000 & 155.36 & 0.115 & \textbf{0.072} & 0.075 & 116.18 & 0.124 & \underline{0.068} & 0.000 & 93.49 & \underline{0.068} & \underline{0.063} & 0.000 \\

            & Gemini-2.5* & 108.60 & 0.511 & 7.337 & 0.219 & 95.02 & 0.165 & 0.252 & 0.000 & \underline{111.59} & 0.209 & 0.463 & \textbf{0.355} & 91.24 & 0.225 & 0.929 & \underline{0.210} & \underline{52.29} & 0.254 & 0.778 & \underline{0.284} \\

            & Claude-3.7* & \underline{61.14} & 0.135 & \underline{0.054} & 0.275 & \underline{90.96} & \underline{0.025} & \underline{0.072} & 0.000 & 118.13 & \underline{0.062} & \underline{0.103} & 0.195 & \underline{63.31} & \textbf{0.063} & 0.042 & 0.000 & 77.85 & 0.067 & 0.053 & 0.000 \\

            \rowcolor{LightGreen}
            & Ours & \textbf{31.58} & \underline{0.235} & 0.123 & \underline{0.478} & \textbf{22.48} & \textbf{0.013} & 0.098 & 0.000 & \textbf{38.56} & \textbf{0.098} & 0.153 & \underline{0.288} & \textbf{25.92} & \underline{0.068} & \textbf{0.203} & \textbf{0.310} & \textbf{30.56} & \textbf{0.106} & \textbf{0.070} & \textbf{0.620} \\
            \hline

            \multirow{5}{*}{\textbf{Refin.}} & GPT-4o & \underline{12.71} & 0.371 & \underline{0.162} & 0.616 & 67.25 & 0.040 & 0.172 & 0.628 & \underline{7.76} & 0.198 & \underline{0.108} & 0.654 & \textbf{5.88} & 0.121 & \textbf{0.278} & 0.577 & \underline{3.27} & 0.178 & \underline{0.127} & 0.618 \\

            & Gemini-2.5* & 23.88 & 0.394 & 0.171 & \underline{0.631} & \underline{8.92} & 0.034 & 0.186 & 0.627 & 20.76 & 0.206 & 0.111 & \underline{0.661} & 10.59 & 0.125 & 0.350 & \underline{0.585} & 5.78 & 0.203 & 0.167 & 0.624 \\

            & Claude-3.7* & 15.02 & \textbf{0.272} & 0.176 & 0.603 & \textbf{3.86} & \underline{0.028} & \underline{0.118} & \underline{0.635} & 17.93 & \underline{0.116} & 0.304 & 0.635 & \underline{6.08} & \underline{0.095} & 0.375 & 0.584 & \textbf{1.67} & \underline{0.136} & \textbf{0.127} & \underline{0.651} \\

            \rowcolor{LightGreen}
            & Ours & \textbf{4.51} & \underline{0.317} & \textbf{0.145} & \textbf{0.681} & 10.60 & \textbf{0.017} & \textbf{0.064} & \textbf{0.732} & \textbf{4.73} & \textbf{0.113} & \textbf{0.072} & \textbf{0.752} & 6.66 & \textbf{0.079} & \underline{0.295} & \textbf{0.641} & 8.25 & \textbf{0.132} & 0.138 & \textbf{0.708} \\
            \hline

            \multicolumn{2}{c}{\textbf{Test Data}} & - & 0.289 & 0.010 & - & - & 0.012 & 0.051 & - & - & 0.083 & 0.054 & - & - & 0.062 & 0.266 & - & - & 0.097 & 0.126 \\ 
    
            \bottomrule[1.2pt]
    	\end{tabular}
    }
    \label{tab:LLMs_comparision}
    \vspace{-1mm}
\end{table*}

\noindent\textbf{Datasets.}
We perform fine-grained adaptation on five document types from M\textsuperscript{6}Doc~\citep{cheng2023m6doc} dataset, which includes:
(\underline{1}) \textbf{Textbook}: 2,080 samples spanning three grade levels and nine subjects, annotated with 42 element categories.
(\underline{2}) \textbf{Newspaper}: 1,000 samples from People’s Daily
and The Wall Street Journal
, with 37 element categories.
(\underline{3}) \textbf{Magazine}: 2,000 samples from globally recognized publishers such as Time USA, annotated with 26 categories.
(\underline{4}) \textbf{Exam}: 2,000 exam paper samples covering the same nine subjects as textbooks, with 31 categories.
(\underline{5}) \textbf{Academic}: 1,000 samples sourced from the arXiv, with 25 categories.
We follow its original 6:1:3 split for training, validation, and testing.

\noindent\textbf{Evaluation Metrics.}
We conduct our experiments using four standard metrics, grouped into two categories. 
(\underline{1}) \textbf{Similarity Assessment}: \textit{Fr\'echet Inception Distance (FID)}~\citep{heusel2017gans} measures the similarity between generated and ground truth layouts by comparing their feature distributions in the embedding space of a deep neural network; \textit{Maximum Intersection over Union (mIoU)}~\citep{kikuchi2021constrained} evaluates spatial alignment by optimally matching generated elements with their ground truth counterparts to maximize the average IoU. 
(\underline{2}) \textbf{Aesthetic Consistency}: We also adopt the \textit{Alignment} (scaled by $100\times$ for better visualization) and \textit{Overlap} metrics from LayoutGAN++~\citep{kikuchi2021constrained} to evaluate generation quality from the perspective of aesthetic principles. 

\noindent\textbf{Implementation Details.}
We choose Qwen2.5-0.5B-Instruct
as our base model.
In coarse-grained learning, we constructed $\sim$9M samples from OmniDocLayout-1M across five tasks with a ratio of 1:1:1:3:3. Our model was trained for 1 epoch on 40 NVIDIA A100 GPUs with a batch size of 16 per device and learning rate of 1e-4, which took about 20 hours. For fine-grained adaptation, we adopted the same data construction strategy and trained for 5 epochs on different categories, respectively. This process was conducted using 8 NVIDIA A100 GPUs, taking about 2 hours, with the same batch size and a lower learning rate of 5e-5.


\begin{figure*}[!t]
    \centering
    \includegraphics[width=1\textwidth]{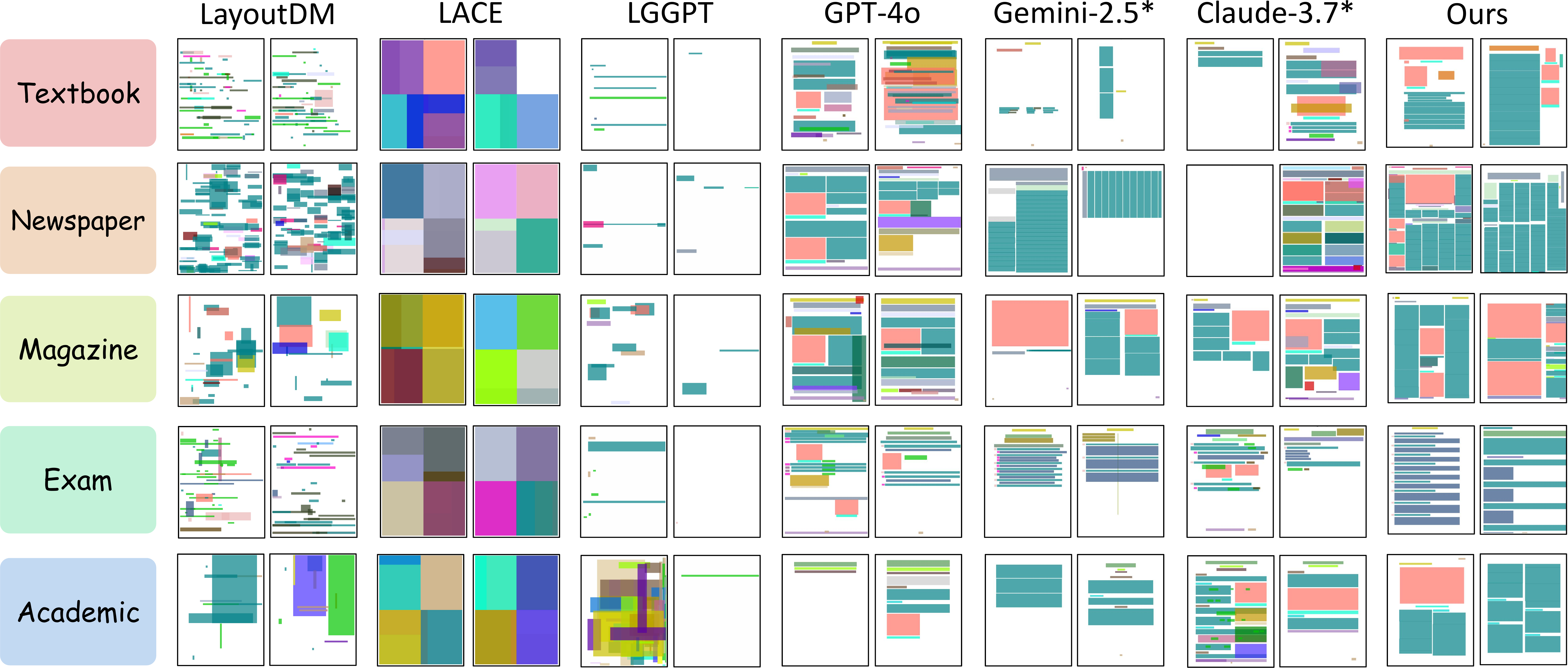}
    \vspace{-1.5em}
    \caption{\textbf{Visualization Examples of Various Methods with U-Cond Task.} For general-purpose LLMs, we adopt the strongest 5-shot setting.}
    \label{fig:visual_comapre}
    \vspace{-1.5em}
\end{figure*}

\subsection{Comparison with Layout Experts}

\noindent\textbf{Baselines.}
For layout generation experts, we compare our approach against four representative methods spanning two major categories:
(1) \textbf{Diffusion-based Models}: LayoutDM~\citep{inoue2023layoutdm} and LACE~\citep{chen2024towards}. 
(2) \textbf{LLM-based Methods}: LayoutPrompter~\citep{lin2023layoutprompter} and LGGPT~\citep{zhang2025smaller}. 
Several work are excluded from comparison for the following reasons:
(1) \textbf{Early Vintage.}
Earlier research such as LayoutGAN++~\citep{kikuchi2021constrained} and LayoutFormer++~\citep{jiang2023layoutformer++} are no longer suitable as fair baselines against modern models.
(2) \textbf{Unavailable or Buggy Implementation.} The latest work like LayoutCoT~\citep{shi2025layoutcot} or LayoutRAG~\citep{wu2025layoutrag} lack publicly available code repositories, and the released implementation of LayoutNUWA~\citep{tang2023layoutnuwa} is hard to reproduce.
(3) \textbf{Poor Convergence.} We trained on LayoutFlow~\citep{guerreiro2024layoutflow} for more than 100K epochs but failed to converge to a satisfactory result.

\vspace{-0.3mm}
\noindent \textbf{Analysis.}
We adopt a unified domain-specific training strategy, selecting the checkpoint with the lowest validation loss for fair comparison. The detailed results are shown in Table~\ref{tab:expert_comparision}.
We observe that: 
(1) For the two diffusion-based models, the overall performance on \textit{FID} is unsatisfactory. This can be attributed to the intrinsic nature of diffusion models, which require substantially more training data and longer convergence time to accurately learn probability distributions. As a result, they fail in low-resource and complex domains. Although LACE achieves significant improvement in element \textit{alignment} through post-processing, it still struggles to control \textit{overlap}. 
(2) For the three LLM-based models, thanks to the autoregressive formulation and strong long-context modeling capability, they can naturally follow aesthetic layout rules without the need for specially designed post-processing. An exception is LGGPT, where the underlying GPT2-XL~\citep{gpt22019radford} often produces incoherent or nonsensical outputs when handling long prompt sequences, a problem not observed on shorter-sequence datasets like PubLayNet. Compared to these baselines, our model achieves consistently superior results across all metrics, with particularly notable gains on \textit{mIoU}.




\subsection{Comparison with General-purpose LLMs}

\noindent\textbf{Baselines.}
For general-purpose LLMs, we select three powerful LLMs: GPT-4o~\citep{openai2024hello}, Gemini-2.5-Flash~\citep{gemini-2.5-flash}, and Claude-3.7-Sonnet~\citep{anthropic2025claude3.7}, 
chosen for their strong long-context capability, with the latter two also offering advanced reasoning abilities relevant to complex layout generation.

\vspace{0.55mm}
\noindent \textbf{Analysis.}
We conduct evaluation under the classic 0/1/5-shot settings to assess whether existing LLMs can achieve competitive performance on conditional document layout generation solely through in-context learning. 
Owing to the page limit, Table~\ref{tab:LLMs_comparision} reports only zero-shot results, while few-shot results are provided in the Appendix. 
We observe that: 
(1) Under zero-shot setting, all general-purpose LLMs achieve reasonably good \textit{alignment} and \textit{overlap}, but exhibit high stochasticity, sometimes leading to extreme outliers. For example, Gemini-2.5-Flash attains an unexpectedly high \textit{overlap} score on the textbook when performing U-Cond task. Moreover, complex layouts such as Newspaper remain the most challenging, as reflected by the highest average \textit{FID}, whereas performance on Academic is overall better, likely because such style are more prevalent in pre-training corpora. Among the three models, Claude-3.7-Sonnet delivers the best results. 
(2) Under few-shot settings, all LLMs improve as the number of shots increases, confirming that in-context learning indeed enables better layout generation. Nevertheless, although additional shots yield better performance, the improvement tends to converge to an intrinsic upper bound, and comes at the cost of longer input sequences, higher API expenses, and slower inference.
The visualization results of different methods on the U-Cond task are shown in Fig.~\ref{fig:visual_comapre}, and more visual cases of our OmniDocLayout-LLM can be found in the Appendix.


\subsection{Ablation Study}



\makeatletter
\newcommand{\verytinysize}{\@setfontsize\verytinysize{5.5pt}{7pt}}
\makeatother

\begin{table}[t]
\vspace{1mm}
\centering
\renewcommand{\arraystretch}{1.115}
\setlength{\tabcolsep}{2pt}
\caption{\textbf{Ablation on Model Sizes and Learning Stages.}
F. and C. denote Fine-grained Adaptation and Coarse-grained Learning only, respectively.}
\vspace{-2mm}

\begin{subtable}[t]{0.49\linewidth}
  \centering
  \verytinysize
  \begin{tabularx}{\linewidth}{
    >{\centering\arraybackslash}p{0.7cm}
    >{\centering\arraybackslash}p{0.4cm}
    *{4}{>{\centering\arraybackslash}X}
  }
    \toprule[1.2pt]
    \textbf{Task} & \textbf{Para} & FID$\downarrow$ & Ali.$\rightarrow$ & Ove.$\rightarrow$ & mIoU$\uparrow$ \\
    \midrule
    \multirow{3}{*}{\textbf{U-Cond}} & 3B & \textbf{35.81} & 0.009 & \textbf{0.052} & 0.000 \\
    & 1.5B & \underline{36.58} & \textbf{0.013} & 0.105 & 0.000 \\
    \rowcolor{LightGreen}
    & 0.5B & 39.73 & \underline{0.015} & \underline{0.084} & 0.000 \\
    \midrule
    \multirow{3}{*}{\textbf{C$\rightarrow$S+P}} & 3B & 26.88 & 0.007 & 0.110 & 0.000 \\
    & 1.5B & \textbf{10.63} & \textbf{0.012} & \underline{0.097} & \underline{0.179} \\
    \rowcolor{LightGreen}
    & 0.5B & \underline{10.71} & \underline{0.014} & \textbf{0.086} & \textbf{0.185} \\
    \midrule
    \multirow{3}{*}{\textbf{C+S$\rightarrow$P}} & 3B & 17.12 & \textbf{0.020} & \textbf{0.181} & \textbf{0.320} \\
    & 1.5B & \textbf{5.65} & 0.022 & 0.205 & 0.238 \\
    \rowcolor{LightGreen}
    & 0.5B & \underline{6.13} & \underline{0.021} & \underline{0.188} & \underline{0.240} \\
    \midrule
    \multirow{3}{*}{\textbf{Compl.}} & 3B & 27.08 & 0.008 & 0.125 & 0.000 \\
    & 1.5B & \underline{26.86} & \underline{0.011} & \textbf{0.093} & 0.000 \\
    \rowcolor{LightGreen}
    & 0.5B & \textbf{22.48} & \textbf{0.013} & \underline{0.098} & 0.000 \\
    \midrule
    \multirow{3}{*}{\textbf{Refin.}} & 3B & 67.24 & 0.017 & \underline{0.062} & 0.725 \\
    & 1.5B & \textbf{6.98} & 0.017 & \textbf{0.061} & \underline{0.730} \\
    \rowcolor{LightGreen}
    & 0.5B & \underline{10.60} & \textbf{0.017} & 0.064 & \textbf{0.732} \\
    \midrule
    \multicolumn{2}{c}{\textbf{Test Data}} & -- & 0.012 & 0.051 & -- \\
    \bottomrule[1.2pt]
  \end{tabularx}
\end{subtable}
\hfill
\begin{subtable}[t]{0.49\linewidth}
  \centering
  \verytinysize
  \begin{tabularx}{\linewidth}{
    >{\centering\arraybackslash}p{0.7cm}
    >{\centering\arraybackslash}p{0.4cm}
    *{4}{>{\centering\arraybackslash}X}
  }
    \toprule[1.2pt]
    \textbf{Task} & \textbf{Stage} & FID$\downarrow$ & Ali.$\rightarrow$ & Ove.$\rightarrow$ & mIoU$\uparrow$ \\
    \midrule
    \multirow{3}{*}{\textbf{U-Cond}} & F. & \underline{42.98} & 0.017 & 8.308 & 0.000 \\
    & C. & 249.1 & \underline{0.016} & \underline{0.388} & 0.000 \\
    \rowcolor{LightGreen}
    & Both & \textbf{39.73} & \textbf{0.015} & \textbf{0.084} & 0.000 \\
    \midrule
    \multirow{3}{*}{\textbf{C$\rightarrow$S+P}} & F. & \underline{14.88} & 0.024 & 0.493 & \underline{0.164} \\
    & C. & 246.4 & \underline{0.016} & \underline{0.357} & 0.000 \\
    \rowcolor{LightGreen}
    & Both & \textbf{10.71} & \textbf{0.014} & \textbf{0.086} & \textbf{0.185} \\
    \midrule
    \multirow{3}{*}{\textbf{C+S$\rightarrow$P}} & F. & \underline{11.24} & 0.023 & \underline{0.476} & \underline{0.220} \\
    & C. & 235.8 & \underline{0.021} & 1.162 & 0.000 \\
    \rowcolor{LightGreen}
    & Both & \textbf{6.13} & \textbf{0.021} & \textbf{0.188} & \textbf{0.240} \\
    \midrule
    \multirow{3}{*}{\textbf{Compl.}} & F. & \underline{36.99} & 0.015 & 6.627 & 0.000 \\
    & C. & 248.79 & \underline{0.015} & \underline{0.480} & 0.000 \\
    \rowcolor{LightGreen}
    & Both & \textbf{22.48} & \textbf{0.013} & \textbf{0.098} & 0.000 \\
    \midrule
    \multirow{3}{*}{\textbf{Refin.}} & F. & \underline{22.07} & 0.023 & 1.452 & \underline{0.618} \\
    & C. & 254.98 & \underline{0.018} & \underline{0.386} & 0.000 \\
    \rowcolor{LightGreen}
    & Both & \textbf{10.60} & \textbf{0.017} & \textbf{0.064} & \textbf{0.732} \\
    \midrule
    \multicolumn{2}{c}{\textbf{Test Data}} & -- & 0.012 & 0.051 & -- \\
    \bottomrule[1.2pt]
  \end{tabularx}
\end{subtable}

\label{tab:combined_ablation}
\vspace{-5mm}
\end{table}
In this section, we perform ablation studies on the number of parameters and the two stages of our Coarse-to-Fine learning paradigm, conducted in the most challenging newspaper domain. The results are shown in Table~\ref{tab:combined_ablation}.
We observe that: 
(1) For model size, the overall differences are marginal. The 3B model achieves slightly lower \textit{Alignment} scores, but its \textit{FID} increases in most tasks. 
This phenomenon is likely due to two main factors. First, as reported in LGGPT, larger LLMs may suffer from insufficient optimization across all parameters or even overfitting, leading to degraded performance. As a result, layout generation \textbf{does not strictly follow the conventional scaling law} observed in other domains. Second, the inherent volatility of \textit{FID} when evaluated on limited test samples makes the metric unstable and sensitive to minor distributional variations, which may also explain the occasional large FID outliers observed in the Refinement task.
(2) 
For learning paradigm, two stages are clearly effective. Coarse-grained learning alone strongly improves global layout organization, as shown by the sharp drop in \textit{Overlap}. Although its \textit{FID} is worse because it does not match the target domain, it achieves much better alignment and especially lower overlap than the fine-only setting, indicating that coarse-grained pre-training enforces fundamental aesthetic rules. With this foundation, fine-grained adaptation further enriches element-level detail. As a result, the full \textit{Coarse~+~Fine} strategy consistently outperforms both coarse-only and fine-only variants.
Notably, multiple zero scores in \textit{mIoU} are attributed to its definition: it requires exact label-level matches, which are often absent in complex multi-element layout generation, particularly in U-Cond and Completion tasks.

\section{Conclusion}
\label{sec:conclusion}


In this work, we move beyond the domain limitations of previous studies and explore complex document layout generation with LLMs. To address the scarcity of diverse training data, we introduce OmniDocLayout-1M, the first million-scale dataset for document layouts, covering six common types such as newspapers and textbooks. Moreover, leveraging the strong capability of LLMs in long-sequence generation, we propose a \textit{Coarse-to-Fine learning paradigm}: first acquiring fundamental aesthetic layout rules from comprehensive document types, and then performing fine-grained adaptation on specific complex domain. Our approach significantly surpasses both existing layout generation experts and powerful general-purpose LLMs. However, our experiments also reveal challenges, such as the inadequacy of current metrics when evaluating complex layouts under limited samples. We will continue to investigate these issues to further advance the field of Document AI.





{
    \small
    \bibliographystyle{ieeenat_fullname}
    \bibliography{main}

\begin{thebibliography}{55}
\providecommand{\natexlab}[1]{#1}
\providecommand{\url}[1]{\texttt{#1}}
\expandafter\ifx\csname urlstyle\endcsname\relax
  \providecommand{\doi}[1]{doi: #1}\else
  \providecommand{\doi}{doi: \begingroup \urlstyle{rm}\Url}\fi

\bibitem[Anthropic(2025)]{anthropic2025claude3.7}
Anthropic.
\newblock Claude-3.7-sonnet.
\newblock \url{https://www.anthropic.com/news/claude-3-7-sonnet}, 2025.

\bibitem[{Anthropic}(2025)]{anthropic2025claude4}
{Anthropic}.
\newblock Claude sonnet 4, 2025.
\newblock Accessed: 2025-05-23.

\bibitem[Antonacopoulos et~al.(2009)Antonacopoulos, Bridson, Papadopoulos, and Pletschacher]{antonacopoulos2009realistic}
Apostolos Antonacopoulos, David Bridson, Christos Papadopoulos, and Stefan Pletschacher.
\newblock A realistic dataset for performance evaluation of document layout analysis.
\newblock In \emph{2009 10th International Conference on Document Analysis and Recognition}, pages 296--300. IEEE, 2009.

\bibitem[Chen et~al.(2024)Chen, Zhang, Zhou, Jain, Xu, Rossi, and Chen]{chen2024towards}
Jian Chen, Ruiyi Zhang, Yufan Zhou, Rajiv Jain, Zhiqiang Xu, Ryan Rossi, and Changyou Chen.
\newblock Towards aligned layout generation via diffusion model with aesthetic constraints.
\newblock \emph{arXiv preprint arXiv:2402.04754}, 2024.

\bibitem[Cheng et~al.(2023)Cheng, Zhang, Wu, Zhang, Zhu, Xie, Li, Ding, and Jin]{cheng2023m6doc}
Hiuyi Cheng, Peirong Zhang, Sihang Wu, Jiaxin Zhang, Qiyuan Zhu, Zecheng Xie, Jing Li, Kai Ding, and Lianwen Jin.
\newblock M6doc: A large-scale multi-format, multi-type, multi-layout, multi-language, multi-annotation category dataset for modern document layout analysis.
\newblock In \emph{Proceedings of the IEEE/CVF Conference on Computer Vision and Pattern Recognition}, pages 15138--15147, 2023.

\bibitem[Cui et~al.(2025)Cui, Sun, Lin, Gao, Zhang, Liu, Wang, Zhang, Zhou, Liu, Zhang, Lv, Huang, Zhang, Zhang, Zhang, Liu, Yu, and Ma]{cui2025paddleocr30technicalreport}
Cheng Cui, Ting Sun, Manhui Lin, Tingquan Gao, Yubo Zhang, Jiaxuan Liu, Xueqing Wang, Zelun Zhang, Changda Zhou, Hongen Liu, Yue Zhang, Wenyu Lv, Kui Huang, Yichao Zhang, Jing Zhang, Jun Zhang, Yi Liu, Dianhai Yu, and Yanjun Ma.
\newblock Paddleocr 3.0 technical report, 2025.

\bibitem[Da et~al.(2023)Da, Luo, Zheng, and Yao]{da2023vision}
Cheng Da, Chuwei Luo, Qi Zheng, and Cong Yao.
\newblock Vision grid transformer for document layout analysis.
\newblock In \emph{Proceedings of the IEEE/CVF international conference on computer vision}, pages 19462--19472, 2023.

\bibitem[Deka et~al.(2017)Deka, Huang, Franzen, Hibschman, Afergan, Li, Nichols, and Kumar]{deka2017rico}
Biplab Deka, Zifeng Huang, Chad Franzen, Joshua Hibschman, Daniel Afergan, Yang Li, Jeffrey Nichols, and Ranjitha Kumar.
\newblock Rico: A mobile app dataset for building data-driven design applications.
\newblock In \emph{Proceedings of the 30th annual ACM symposium on user interface software and technology}, pages 845--854, 2017.

\bibitem[Google(2025{\natexlab{a}})]{gemini-2.5-flash}
Google.
\newblock gemini-2.5-flash.
\newblock \url{https://deepmind.google/models/gemini/flash/}, 2025{\natexlab{a}}.

\bibitem[Google(2025{\natexlab{b}})]{gemini-2.5-pro}
Google.
\newblock gemini-2.5-pro.
\newblock \url{https://deepmind.google/models/gemini/pro/}, 2025{\natexlab{b}}.

\bibitem[Guerreiro et~al.(2024)Guerreiro, Inoue, Masui, Otani, and Nakayama]{guerreiro2024layoutflow}
Julian Jorge~Andrade Guerreiro, Naoto Inoue, Kento Masui, Mayu Otani, and Hideki Nakayama.
\newblock Layoutflow: flow matching for layout generation.
\newblock In \emph{European Conference on Computer Vision}, pages 56--72. Springer, 2024.

\bibitem[Gupta et~al.(2021)Gupta, Lazarow, Achille, Davis, Mahadevan, and Shrivastava]{gupta2021layouttransformer}
Kamal Gupta, Justin Lazarow, Alessandro Achille, Larry~S Davis, Vijay Mahadevan, and Abhinav Shrivastava.
\newblock Layouttransformer: Layout generation and completion with self-attention.
\newblock In \emph{Proceedings of the IEEE/CVF International Conference on Computer Vision}, pages 1004--1014, 2021.

\bibitem[Harley et~al.(2015)Harley, Ufkes, and Derpanis]{harley2015evaluation}
Adam~W Harley, Alex Ufkes, and Konstantinos~G Derpanis.
\newblock Evaluation of deep convolutional nets for document image classification and retrieval.
\newblock In \emph{2015 13th international conference on document analysis and recognition (ICDAR)}, pages 991--995. IEEE, 2015.

\bibitem[Heusel et~al.(2017)Heusel, Ramsauer, Unterthiner, Nessler, and Hochreiter]{heusel2017gans}
Martin Heusel, Hubert Ramsauer, Thomas Unterthiner, Bernhard Nessler, and Sepp Hochreiter.
\newblock Gans trained by a two time-scale update rule converge to a local nash equilibrium.
\newblock \emph{Advances in neural information processing systems}, 30, 2017.

\bibitem[Hsu et~al.(2023)Hsu, He, Peng, Kong, and Zhang]{Hsu_2023_CVPR}
Hsiao~Yuan Hsu, Xiangteng He, Yuxin Peng, Hao Kong, and Qing Zhang.
\newblock Posterlayout: A new benchmark and approach for content-aware visual-textual presentation layout.
\newblock In \emph{Proceedings of the IEEE/CVF Conference on Computer Vision and Pattern Recognition (CVPR)}, pages 6018--6026, 2023.

\bibitem[Hu et~al.(2020)Hu, Huang, Tang, Van~Kaick, Zhang, and Huang]{hu2020graph2plan}
Ruizhen Hu, Zeyu Huang, Yuhan Tang, Oliver Van~Kaick, Hao Zhang, and Hui Huang.
\newblock Graph2plan: Learning floorplan generation from layout graphs.
\newblock \emph{ACM Transactions on Graphics (TOG)}, 39\penalty0 (4):\penalty0 118--1, 2020.

\bibitem[Inoue et~al.(2023)Inoue, Kikuchi, Simo-Serra, Otani, and Yamaguchi]{inoue2023layoutdm}
Naoto Inoue, Kotaro Kikuchi, Edgar Simo-Serra, Mayu Otani, and Kota Yamaguchi.
\newblock Layoutdm: Discrete diffusion model for controllable layout generation.
\newblock In \emph{Proceedings of the IEEE/CVF Conference on Computer Vision and Pattern Recognition}, pages 10167--10176, 2023.

\bibitem[Jiang et~al.(2023)Jiang, Guo, Sun, Deng, Wu, Mijovic, Yang, Lou, and Zhang]{jiang2023layoutformer++}
Zhaoyun Jiang, Jiaqi Guo, Shizhao Sun, Huayu Deng, Zhongkai Wu, Vuksan Mijovic, Zijiang~James Yang, Jian-Guang Lou, and Dongmei Zhang.
\newblock Layoutformer++: Conditional graphic layout generation via constraint serialization and decoding space restriction.
\newblock In \emph{Proceedings of the IEEE/CVF Conference on Computer Vision and Pattern Recognition}, pages 18403--18412, 2023.

\bibitem[Kang et~al.(2025)Kang, Wen, Wen, Ye, Li, Feng, Zhou, Wang, Lin, Zhang, et~al.]{kang2025legion}
Hengrui Kang, Siwei Wen, Zichen Wen, Junyan Ye, Weijia Li, Peilin Feng, Baichuan Zhou, Bin Wang, Dahua Lin, Linfeng Zhang, et~al.
\newblock Legion: Learning to ground and explain for synthetic image detection.
\newblock \emph{arXiv preprint arXiv:2503.15264}, 2025.

\bibitem[Kikuchi et~al.(2021)Kikuchi, Simo-Serra, Otani, and Yamaguchi]{kikuchi2021constrained}
Kotaro Kikuchi, Edgar Simo-Serra, Mayu Otani, and Kota Yamaguchi.
\newblock Constrained graphic layout generation via latent optimization.
\newblock In \emph{Proceedings of the 29th ACM International Conference on Multimedia}, pages 88--96, 2021.

\bibitem[Kong et~al.(2022)Kong, Jiang, Chang, Zhang, Hao, Gong, and Essa]{kong2022blt}
Xiang Kong, Lu Jiang, Huiwen Chang, Han Zhang, Yuan Hao, Haifeng Gong, and Irfan Essa.
\newblock Blt: Bidirectional layout transformer for controllable layout generation.
\newblock In \emph{European Conference on Computer Vision}, pages 474--490. Springer, 2022.

\bibitem[Li et~al.(2020)Li, Xu, Cui, Huang, Wei, Li, and Zhou]{li2020docbank}
Minghao Li, Yiheng Xu, Lei Cui, Shaohan Huang, Furu Wei, Zhoujun Li, and Ming Zhou.
\newblock Docbank: A benchmark dataset for document layout analysis.
\newblock \emph{arXiv preprint arXiv:2006.01038}, 2020.

\bibitem[Li et~al.(2025)Li, Liu, Liu, Ma, Zhang, Zhang, Guo, Zhang, Wang, and Bai]{li2025monkeyocrdocumentparsingstructurerecognitionrelation}
Zhang Li, Yuliang Liu, Qiang Liu, Zhiyin Ma, Ziyang Zhang, Shuo Zhang, Zidun Guo, Jiarui Zhang, Xinyu Wang, and Xiang Bai.
\newblock Monkeyocr: Document parsing with a structure-recognition-relation triplet paradigm, 2025.

\bibitem[Lin et~al.(2023)Lin, Guo, Sun, Yang, Lou, and Zhang]{lin2023layoutprompter}
Jiawei Lin, Jiaqi Guo, Shizhao Sun, Zijiang Yang, Jian-Guang Lou, and Dongmei Zhang.
\newblock Layoutprompter: Awaken the design ability of large language models.
\newblock \emph{Advances in Neural Information Processing Systems}, 36:\penalty0 43852--43879, 2023.

\bibitem[Liu et~al.(2025{\natexlab{a}})Liu, Wang, Chen, Han, Wang, Yuan, Song, Zhang, Huang, and Chen]{liu2025global}
Xuyang Liu, Ziming Wang, Junjie Chen, Yuhang Han, Yingyao Wang, Jiale Yuan, Jun Song, Linfeng Zhang, Siteng Huang, and Honggang Chen.
\newblock Global compression commander: Plug-and-play inference acceleration for high-resolution large vision-language models.
\newblock \emph{arXiv preprint arXiv:2501.05179}, 2025{\natexlab{a}}.

\bibitem[Liu et~al.(2025{\natexlab{b}})Liu, Wen, Wang, Chen, Tao, Wang, Jin, Zou, Wang, Liao, et~al.]{liu2025shifting}
Xuyang Liu, Zichen Wen, Shaobo Wang, Junjie Chen, Zhishan Tao, Yubo Wang, Xiangqi Jin, Chang Zou, Yiyu Wang, Chenfei Liao, et~al.
\newblock Shifting ai efficiency from model-centric to data-centric compression.
\newblock \emph{arXiv preprint arXiv:2505.19147}, 2025{\natexlab{b}}.

\bibitem[Ma et~al.(2025)Ma, Zhang, Lu, Wang, Zhou, Song, and Zhang]{ma2025mmg}
Junpeng Ma, Qizhe Zhang, Ming Lu, Zhibin Wang, Qiang Zhou, Jun Song, and Shanghang Zhang.
\newblock Mmg-vid: Maximizing marginal gains at segment-level and token-level for efficient video llms.
\newblock \emph{arXiv preprint arXiv:2508.21044}, 2025.

\bibitem[McInnes et~al.(2018)McInnes, Healy, and Melville]{mcinnes2018umap}
Leland McInnes, John Healy, and James Melville.
\newblock Umap: Uniform manifold approximation and projection for dimension reduction.
\newblock \emph{arXiv preprint arXiv:1802.03426}, 2018.

\bibitem[Nauata et~al.(2020)Nauata, Chang, Cheng, Mori, and Furukawa]{nauata2020house}
Nelson Nauata, Kai-Hung Chang, Chin-Yi Cheng, Greg Mori, and Yasutaka Furukawa.
\newblock House-gan: Relational generative adversarial networks for graph-constrained house layout generation.
\newblock In \emph{European Conference on Computer Vision}, pages 162--177. Springer, 2020.

\bibitem[OpenAI(2024)]{openai2024hello}
OpenAI.
\newblock Hello gpt-4o.
\newblock \url{https://openai.com/index/hello-gpt-4o/}, 2024.

\bibitem[Ouyang et~al.(2025)Ouyang, Qu, Zhou, Zhu, Zhang, Lin, Wang, Zhao, Jiang, Zhao, et~al.]{ouyang2025omnidocbench}
Linke Ouyang, Yuan Qu, Hongbin Zhou, Jiawei Zhu, Rui Zhang, Qunshu Lin, Bin Wang, Zhiyuan Zhao, Man Jiang, Xiaomeng Zhao, et~al.
\newblock Omnidocbench: Benchmarking diverse pdf document parsing with comprehensive annotations.
\newblock In \emph{Proceedings of the Computer Vision and Pattern Recognition Conference}, pages 24838--24848, 2025.

\bibitem[Pfitzmann et~al.(2022)Pfitzmann, Auer, Dolfi, Nassar, and Staar]{pfitzmann2022doclaynet}
Birgit Pfitzmann, Christoph Auer, Michele Dolfi, Ahmed~S Nassar, and Peter Staar.
\newblock Doclaynet: A large human-annotated dataset for document-layout segmentation.
\newblock In \emph{Proceedings of the 28th ACM SIGKDD conference on knowledge discovery and data mining}, pages 3743--3751, 2022.

\bibitem[Radford et~al.(2019)Radford, Wu, and et~al.]{gpt22019radford}
Alec Radford, Jeffrey Wu, and et al.
\newblock {Language Models are Unsupervised Multitask Learners}.
\newblock \emph{OpenAI blog}, 1\penalty0 (8):\penalty0 9, 2019.

\bibitem[rednote(2025)]{dotsocr}
rednote.
\newblock dotsocr.
\newblock \url{https://github.com/rednote-hilab/dots.ocr}, 2025.

\bibitem[Shi et~al.(2025)Shi, Su, Ning, Wei, and Gao]{shi2025layoutcot}
Hengyu Shi, Junhao Su, Huansheng Ning, Xiaoming Wei, and Jialin Gao.
\newblock Layoutcot: Unleashing the deep reasoning potential of large language models for layout generation.
\newblock \emph{arXiv preprint arXiv:2504.10829}, 2025.

\bibitem[Tang et~al.(2023)Tang, Wu, Li, and Duan]{tang2023layoutnuwa}
Zecheng Tang, Chenfei Wu, Juntao Li, and Nan Duan.
\newblock Layoutnuwa: Revealing the hidden layout expertise of large language models.
\newblock \emph{arXiv preprint arXiv:2309.09506}, 2023.

\bibitem[Wang et~al.(2024)Wang, Xu, Zhao, Ouyang, Wu, Zhao, Xu, Liu, Qu, Shang, Zhang, Wei, Sui, Li, Shi, Qiao, Lin, and He]{wang2024mineruopensourcesolutionprecise}
Bin Wang, Chao Xu, Xiaomeng Zhao, Linke Ouyang, Fan Wu, Zhiyuan Zhao, Rui Xu, Kaiwen Liu, Yuan Qu, Fukai Shang, Bo Zhang, Liqun Wei, Zhihao Sui, Wei Li, Botian Shi, Yu Qiao, Dahua Lin, and Conghui He.
\newblock Mineru: An open-source solution for precise document content extraction, 2024.

\bibitem[Wang et~al.(2025)Wang, Jin, Wang, Wang, Zhang, Li, Wen, Li, He, Hu, et~al.]{wang2025data}
Shaobo Wang, Xiangqi Jin, Ziming Wang, Jize Wang, Jiajun Zhang, Kaixin Li, Zichen Wen, Zhong Li, Conghui He, Xuming Hu, et~al.
\newblock Data whisperer: Efficient data selection for task-specific llm fine-tuning via few-shot in-context learning.
\newblock \emph{arXiv preprint arXiv:2505.12212}, 2025.

\bibitem[Wen et~al.(2025{\natexlab{a}})Wen, Ye, Feng, Kang, Wen, Chen, Wu, Wu, He, and Li]{wen2025spot}
Siwei Wen, Junyan Ye, Peilin Feng, Hengrui Kang, Zichen Wen, Yize Chen, Jiang Wu, Wenjun Wu, Conghui He, and Weijia Li.
\newblock Spot the fake: Large multimodal model-based synthetic image detection with artifact explanation.
\newblock \emph{arXiv preprint arXiv:2503.14905}, 2025{\natexlab{a}}.

\bibitem[Wen et~al.(2024)Wen, Guo, and Zhang]{wen2024aidbench}
Zichen Wen, Dadi Guo, and Huishuai Zhang.
\newblock Aidbench: A benchmark for evaluating the authorship identification capability of large language models.
\newblock \emph{arXiv preprint arXiv:2411.13226}, 2024.

\bibitem[Wen et~al.(2025{\natexlab{b}})Wen, Gao, Li, He, and Zhang]{wen2025token}
Zichen Wen, Yifeng Gao, Weijia Li, Conghui He, and Linfeng Zhang.
\newblock Token pruning in multimodal large language models: Are we solving the right problem?
\newblock \emph{arXiv preprint arXiv:2502.11501}, 2025{\natexlab{b}}.

\bibitem[Wen et~al.(2025{\natexlab{c}})Wen, Gao, Wang, Zhang, Zhang, Li, He, and Zhang]{wen2025stop}
Zichen Wen, Yifeng Gao, Shaobo Wang, Junyuan Zhang, Qintong Zhang, Weijia Li, Conghui He, and Linfeng Zhang.
\newblock Stop looking for important tokens in multimodal language models: Duplication matters more.
\newblock \emph{arXiv preprint arXiv:2502.11494}, 2025{\natexlab{c}}.

\bibitem[Wen et~al.(2025{\natexlab{d}})Wen, Wang, Zhou, Zhang, Zhang, Gao, Chen, Wang, Li, He, et~al.]{wen2025efficient}
Zichen Wen, Shaobo Wang, Yufa Zhou, Junyuan Zhang, Qintong Zhang, Yifeng Gao, Zhaorun Chen, Bin Wang, Weijia Li, Conghui He, et~al.
\newblock Efficient multi-modal large language models via progressive consistency distillation.
\newblock \emph{arXiv preprint arXiv:2510.00515}, 2025{\natexlab{d}}.

\bibitem[Wen et~al.(2025{\natexlab{e}})Wen, Wang, Liao, Yang, Li, Liu, He, Feng, Liu, Lyu, et~al.]{wen2025ai}
Zichen Wen, Yiyu Wang, Chenfei Liao, Boxue Yang, Junxian Li, Weifeng Liu, Haocong He, Bolong Feng, Xuyang Liu, Yuanhuiyi Lyu, et~al.
\newblock Ai for service: Proactive assistance with ai glasses.
\newblock \emph{arXiv preprint arXiv:2510.14359}, 2025{\natexlab{e}}.

\bibitem[Wu et~al.(2025)Wu, Wang, Zhou, Liu, Hua, and Li]{wu2025layoutrag}
Yuxuan Wu, Le Wang, Sanping Zhou, Mengnan Liu, Gang Hua, and Haoxiang Li.
\newblock Layoutrag: Retrieval-augmented model for content-agnostic conditional layout generation.
\newblock \emph{arXiv preprint arXiv:2506.02697}, 2025.

\bibitem[Xiong et~al.(2025)Xiong, Wen, Gu, Liu, Zhang, Kang, Yang, Zhang, Li, He, et~al.]{xiong2025prune2drive}
Minhao Xiong, Zichen Wen, Zhuangcheng Gu, Xuyang Liu, Rui Zhang, Hengrui Kang, Jiabing Yang, Junyuan Zhang, Weijia Li, Conghui He, et~al.
\newblock Prune2drive: A plug-and-play framework for accelerating vision-language models in autonomous driving.
\newblock \emph{arXiv preprint arXiv:2508.13305}, 2025.

\bibitem[Yang et~al.(2025)Yang, Chen, Wen, Cui, Li, Xu, Fang, Huang, and Wang]{yang2025dtpa}
Jiabing Yang, Yixiang Chen, Zichen Wen, Chenhang Cui, Peiyan Li, Yuan Xu, Bowen Fang, Yan Huang, and Liang Wang.
\newblock Dtpa: Dynamic token-level prefix augmentation for controllable text generation.
\newblock \emph{arXiv preprint arXiv:2508.04047}, 2025.

\bibitem[Yang et~al.(2017)Yang, Yumer, Asente, Kraley, Kifer, and Lee~Giles]{yang2017learning}
Xiao Yang, Ersin Yumer, Paul Asente, Mike Kraley, Daniel Kifer, and C Lee~Giles.
\newblock Learning to extract semantic structure from documents using multimodal fully convolutional neural networks.
\newblock In \emph{Proceedings of the IEEE conference on computer vision and pattern recognition}, pages 5315--5324, 2017.

\bibitem[Yang et~al.(2024)Yang, Wen, Qu, Chen, Xiang, Chen, and Yao]{yang2024memorization}
Xinyu Yang, Zichen Wen, Wenjie Qu, Zhaorun Chen, Zhiying Xiang, Beidi Chen, and Huaxiu Yao.
\newblock Memorization and privacy risks in domain-specific large language models.
\newblock In \emph{ICLR 2024 Workshop on Reliable and Responsible Foundation Models}, 2024.

\bibitem[Zhang et~al.(2025{\natexlab{a}})Zhang, Zhang, Wang, Ouyang, Wen, Li, Chow, He, and Zhang]{zhang2025ocr}
Junyuan Zhang, Qintong Zhang, Bin Wang, Linke Ouyang, Zichen Wen, Ying Li, Ka-Ho Chow, Conghui He, and Wentao Zhang.
\newblock Ocr hinders rag: Evaluating the cascading impact of ocr on retrieval-augmented generation.
\newblock In \emph{Proceedings of the IEEE/CVF International Conference on Computer Vision}, pages 17443--17453, 2025{\natexlab{a}}.

\bibitem[Zhang et~al.(2025{\natexlab{b}})Zhang, Zhang, Cao, Li, and Jin]{zhang2025smaller}
Peirong Zhang, Jiaxin Zhang, Jiahuan Cao, Hongliang Li, and Lianwen Jin.
\newblock Smaller but better: Unifying layout generation with smaller large language models.
\newblock \emph{International Journal of Computer Vision}, 133\penalty0 (7):\penalty0 3891--3917, 2025{\natexlab{b}}.

\bibitem[Zhao et~al.(2024)Zhao, Kang, Wang, and He]{zhao2024doclayout}
Zhiyuan Zhao, Hengrui Kang, Bin Wang, and Conghui He.
\newblock Doclayout-yolo: Enhancing document layout analysis through diverse synthetic data and global-to-local adaptive perception.
\newblock \emph{arXiv preprint arXiv:2410.12628}, 2024.

\bibitem[Zheng et~al.(2019)Zheng, Qiao, Cao, and Lau]{zheng2019content}
Xinru Zheng, Xiaotian Qiao, Ying Cao, and Rynson~WH Lau.
\newblock Content-aware generative modeling of graphic design layouts.
\newblock \emph{ACM Transactions on Graphics (TOG)}, 38\penalty0 (4):\penalty0 1--15, 2019.

\bibitem[Zhong et~al.(2019)Zhong, Tang, and Yepes]{zhong2019publaynet}
Xu Zhong, Jianbin Tang, and Antonio~Jimeno Yepes.
\newblock Publaynet: largest dataset ever for document layout analysis.
\newblock In \emph{2019 International conference on document analysis and recognition (ICDAR)}, pages 1015--1022. IEEE, 2019.

\bibitem[Zhou et~al.(2022)Zhou, Xu, Ma, Ge, Jiang, and Xu]{zhou2022composition}
Min Zhou, Chenchen Xu, Ye Ma, Tiezheng Ge, Yuning Jiang, and Weiwei Xu.
\newblock Composition-aware graphic layout gan for visual-textual presentation designs.
\newblock \emph{arXiv preprint arXiv:2205.00303}, 2022.

\end{thebibliography}
}

\clearpage
\setcounter{page}{1}
\maketitlesupplementary


\appendix

\large
\noindent\textbf{Contents of the Appendices:}
\vspace{1mm}

\normalsize
Section~\ref{app:clarification}. Comparison of Scenario Scope and Layout Complexity with Similar Works.

Section~\ref{app:dataset}. More Details on Our OmniDocLayout-1M.

Section~\ref{app:model}. More Qualitative and Quantitative Analysis on our OmniDocLayout-LLM.





\section{Comparison of Scenario Scope and Layout Complexity with Similar Works}\label{app:clarification}

\subsection{Comparison Across Multiple Scenario Scopes}
\begin{figure}[!h] 
    \centering
    \vspace{-2mm}
    \includegraphics[width=1.0\linewidth]{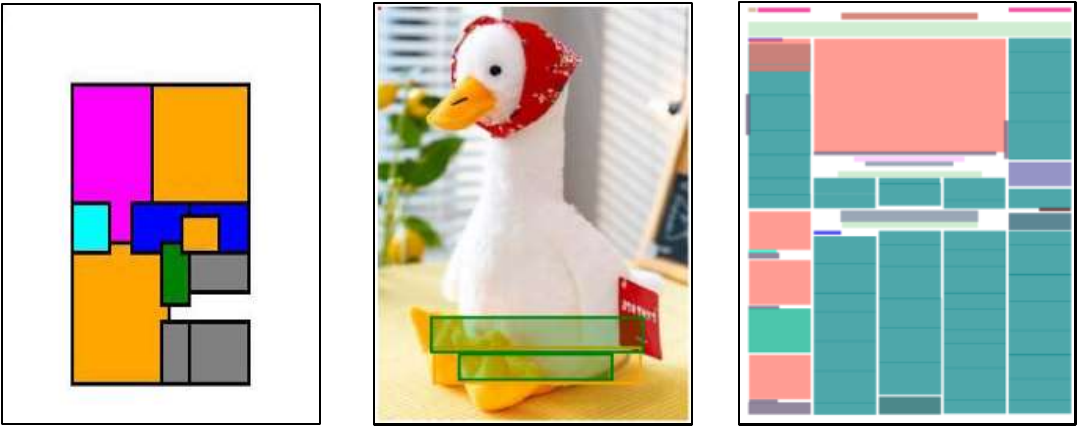}
    \vspace{-6mm}
    \caption{Scenario Scope Comparison. (Left) shows a generated room layout by~\citep{nauata2020house}. (Middle) shows a generated graphic layout by~\citep{Hsu_2023_CVPR}. (Right) shows a generated document layout by our OmniDocLayout-LLM.}
    \label{fig:appli_compare}
    \vspace{-5mm}
\end{figure}

Recently, layout generation has been extensively studied across diverse application scenarios. Existing approaches can be broadly categorized into three groups: room layout planning, graphic layout design, and document layout generation. To be specific, room layout planning~\citep{nauata2020house,hu2020graph2plan} focuses on partitioning indoor spaces into functional regions while satisfying spatial and accessibility constraints (\emph{e.g.}, ensuring navigable and appropriately adjacent areas), typically involving a small and fixed set of element types. Graphic layout design (\emph{e.g.}, poster~\citep{zhou2022composition,Hsu_2023_CVPR}), in contrast, emphasizes saliency-aware placement of a few elements such that the main subject of a background image remains unobstructed. Our work instead targets document layout generation, where numerous heterogeneous elements (\emph{e.g.}, headers, paragraphs, tables, figures) must be arranged within a rectangular page according to the stylistic norms of different document types. Among these three application scenarios, as illustrated in Fig.~\ref{fig:appli_compare}, document layout generation is the most challenging, featuring:

\begin{itemize}[leftmargin=10pt, topsep=0pt, itemsep=1pt, partopsep=1pt, parsep=1pt]
\item richer element categories, deeper hierarchical structures, and substantial style diversity across document types;

\item stricter layout rules across multiple dimensions, including alignment, reading order, spacing, and style;

\item significant challenges in modeling and spatial organization when handling a large number of elements per page.

\end{itemize}

\begin{figure}[!t]
    \centering

    \includegraphics[width=1.0\linewidth]{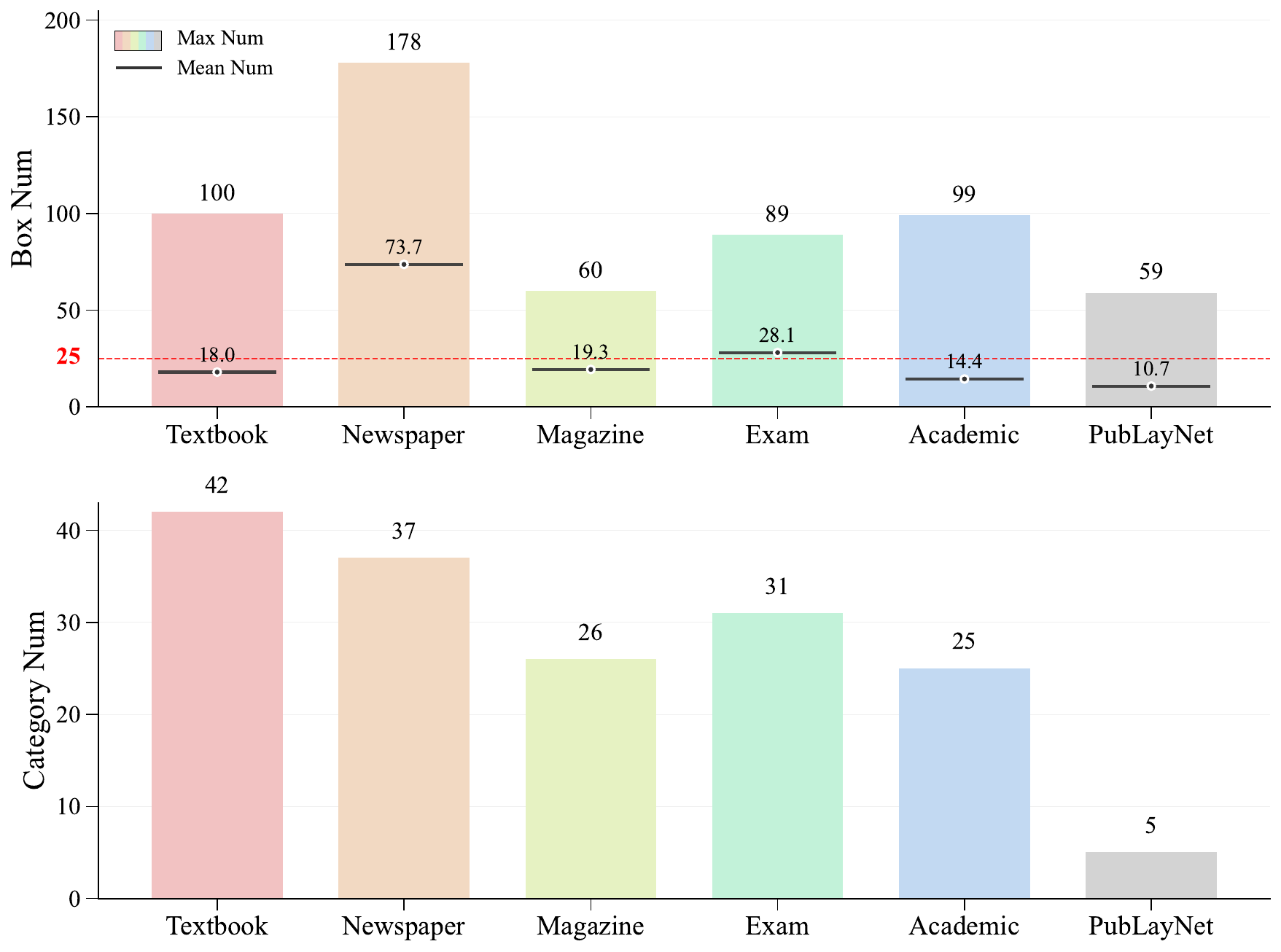}

    \vspace{-2mm}
    \caption{Layout Complexity Comparison Between Five Complex Types in M\textsuperscript{6}Doc and Widely-used PubLayNet. 
    (Top) compares the maximum and average number of elements per page. (Bottom) compares the granularity of element categorization.
    The \textcolor{red}{red dashed line} indicates the default maximum number of elements (\textcolor{red}{25}) allowed in prior methods.}
    \label{fig:m6doc_compare}
    \vspace{-5mm}
\end{figure}

\subsection{Comparison of Layout Complexity Across Various Documents}
In this section, we examine prior work on document layout generation. Existing studies~\citep{inoue2023layoutdm,chen2024towards,guerreiro2024layoutflow,shi2025layoutcot,wu2025layoutrag,lin2023layoutprompter,jiang2023layoutformer++} primarily evaluate models on PubLayNet~\citep{zhong2019publaynet}, which largely consists of simple layouts from single- or double-column academic papers. Our goal is to expand layout generation to a broader range of highly complex documents, and we therefore use the M\textsuperscript{6}Doc~\citep{cheng2023m6doc} dataset for evaluation.
We compare our test set with PubLayNet from two perspectives, as illustrated in Figure~\ref{fig:m6doc_compare}.
(1) Maximum and average number of elements per page. The five document types in M\textsuperscript{6}Doc consistently exhibit higher complexity in both metrics. In particular, the newspaper type reaches a maximum of 178 elements and an average of 73.7 elements per page, while PubLayNet contains only 59 elements at maximum and 10.7 on average.
(2) Granularity of element categories. PubLayNet defines only 5 element types: text, title, list, table, and figure. In contrast, M\textsuperscript{6}Doc provides much finer-grained semantic categories. The textbook category contains 42 element classes, and even the academic category includes 25 classes. These findings indicate that our evaluation setting involves substantially higher layout complexity and generation difficulty than previous benchmarks, while also better reflecting real-world document diversity.

Notably, several recent studies, such as LayoutNUWA~\citep{tang2023layoutnuwa} and LGGPT~\citep{zhang2025smaller}, have also attempted to extend layout generation to more diverse and complex document styles. However, the range of document types they cover remains substantially narrower than ours, indicating that their evaluation settings are still less comprehensive.






\section{OmniDocLayout-1M Dataset}\label{app:dataset}

\subsection{Curation Details}
\noindent \textbf{Dual-deduplication.}
To remove duplicate samples from the final dataset, we adopted a \textit{dual-deduplication strategy}: image-level deduplication was first applied during preprocessing, followed by layout-level deduplication after annotation.
For image-level deduplication, we employed perceptual hashing (pHash) for document image deduplication. Each image was converted to grayscale and resized to 32x32 pixels, generating a 1024-bit hash. We compared hash values using Hamming distance, with a normalized threshold of 0.05, allowing for a certain tolerance of noise while identifying duplicates or near-duplicates. This resolution was chosen because document images contain structured layouts and fine textual details, and a higher number of bits provides a better representation to achieve more reliable accuracy.
For layout-level deduplication, we calculated the mIoU (max Intersection over Union) score between each pair of samples. For pairs with an mIoU greater than 0.95, we retained only one sample selected at random.

\noindent \textbf{Data Filtering.} 
To ensure the quality of annotations from the automated labeling pipeline and discard low-quality samples that do not conform to prior knowledge, we performed a filtering step after annotation. We filtered the samples based on three dimensions: 
(1) Minimum number of elements. Based on empirical intuition, for documents such as textbooks, magazines, exams, and academic papers, we discarded pages with fewer than 5 elements, while for more complex documents like newspapers, this threshold was increased to 10.
(2) Fill ratio. For certain densely arranged document types, such as newspapers, large amounts of white space are generally not allowed in the layout. Therefore, we filtered out pages with a fill ratio of less than 60\%.
(3) Overlap Score. Although overlaps are common in UI or poster design tasks, they are generally avoided in document layouts. Therefore, samples with abnormally high overlap scores were marked as incorrectly annotated and discarded. As for alignment score, it can be naturally high in certain non-Manhattan layout document types and is thus not suitable as a filtering criterion, as using it would lead to a significant loss of layout diversity.

\begin{figure}[!t]
    \centering

    \includegraphics[width=1.0\linewidth]{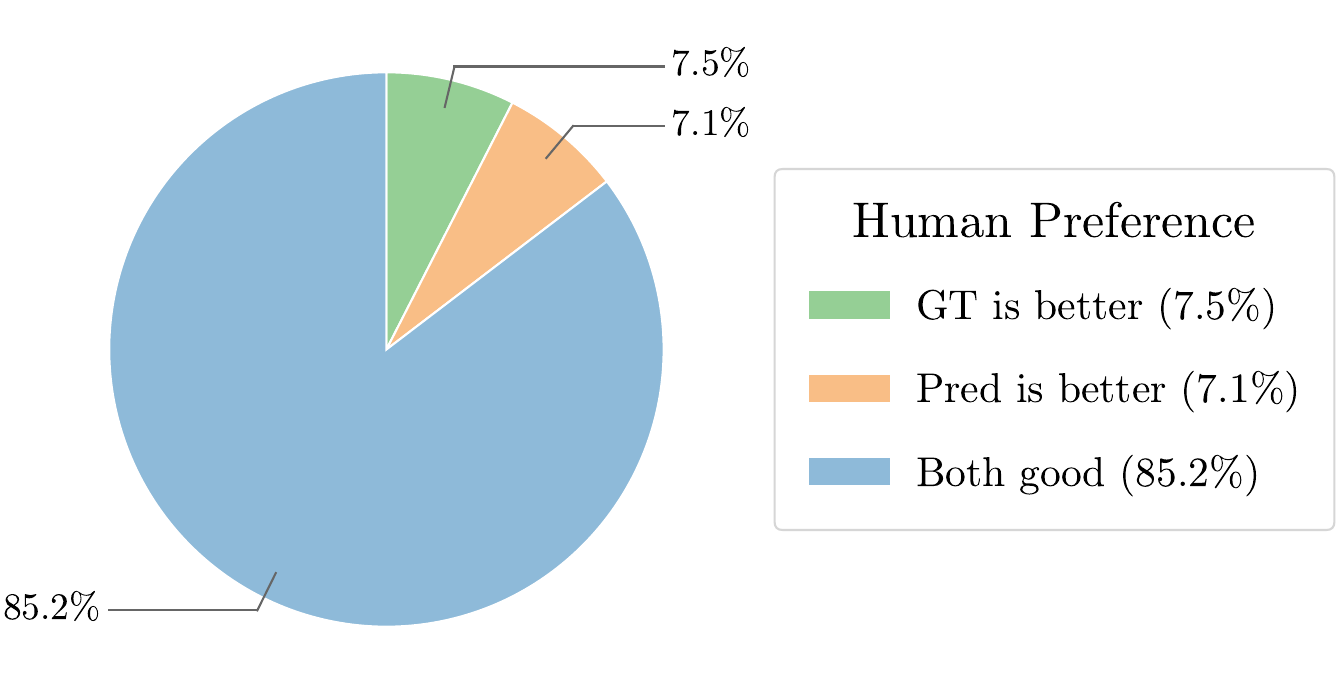}

    \vspace{-4mm}
    \caption{Human evaluation of 1,200 pages comparing model-generated and human-annotated layouts. ``Pred'' refer to the bounding box predicted by MinerU~\citep{wang2024mineruopensourcesolutionprecise}.}
    \label{fig:human_eval}
    \vspace{-5.5mm}
\end{figure}

\subsection{Final Quality Control}
To address the concern that using model-generated bounding boxes as training data may introduce annotation noise or bias, we design a blind human evaluation to directly assess the perceptual quality of model-annotated layouts relative to human-annotated ground truth.

\begin{figure*}[!t]
    \centering

    \begin{subfigure}[b]{0.32\textwidth}
        \centering
        \includegraphics[width=\textwidth]{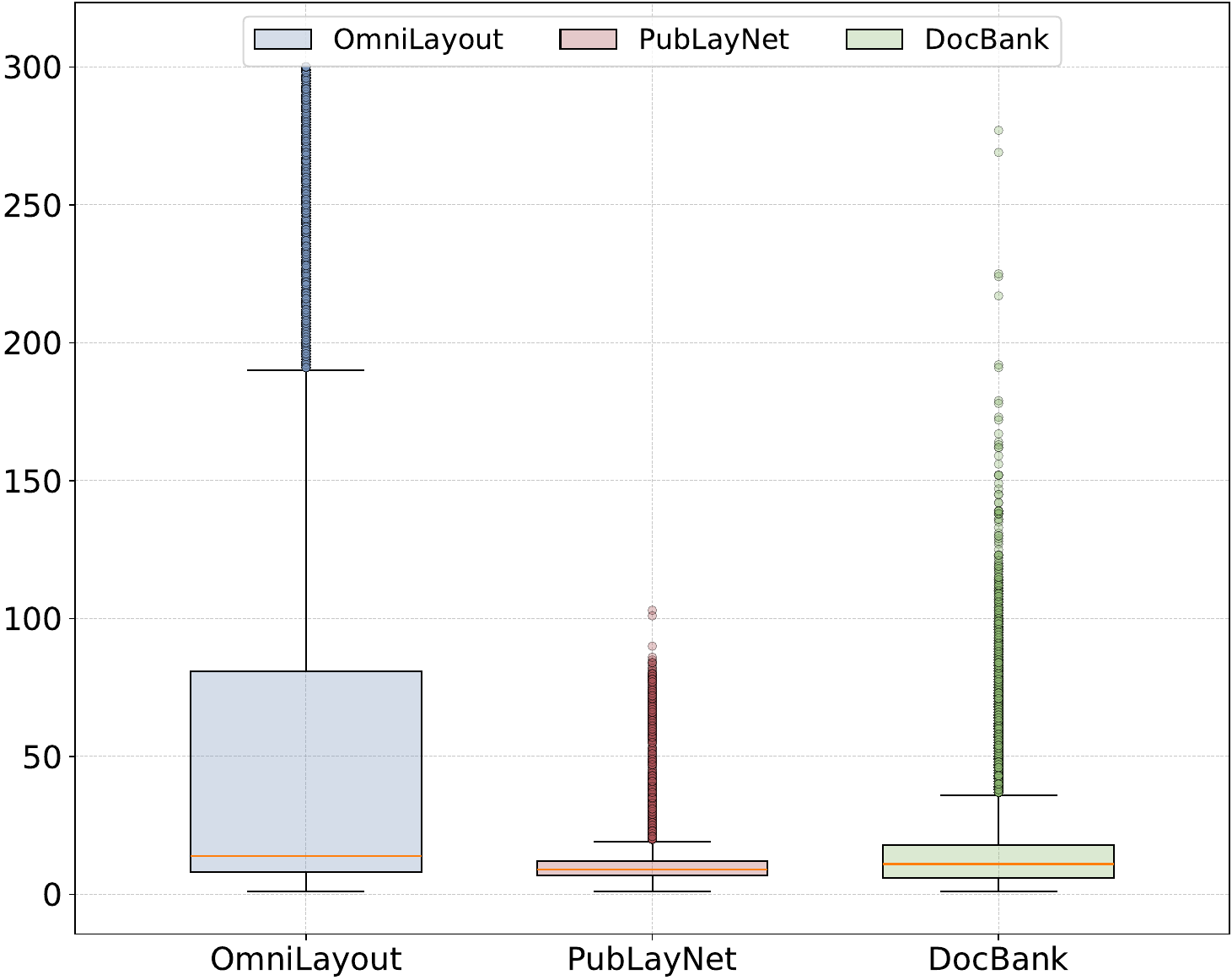}
        \caption{Element numbers per page.}
        \label{fig:sub1}
    \end{subfigure}
    \hfill 
    \begin{subfigure}[b]{0.32\textwidth}
        \centering
        \includegraphics[width=\textwidth]{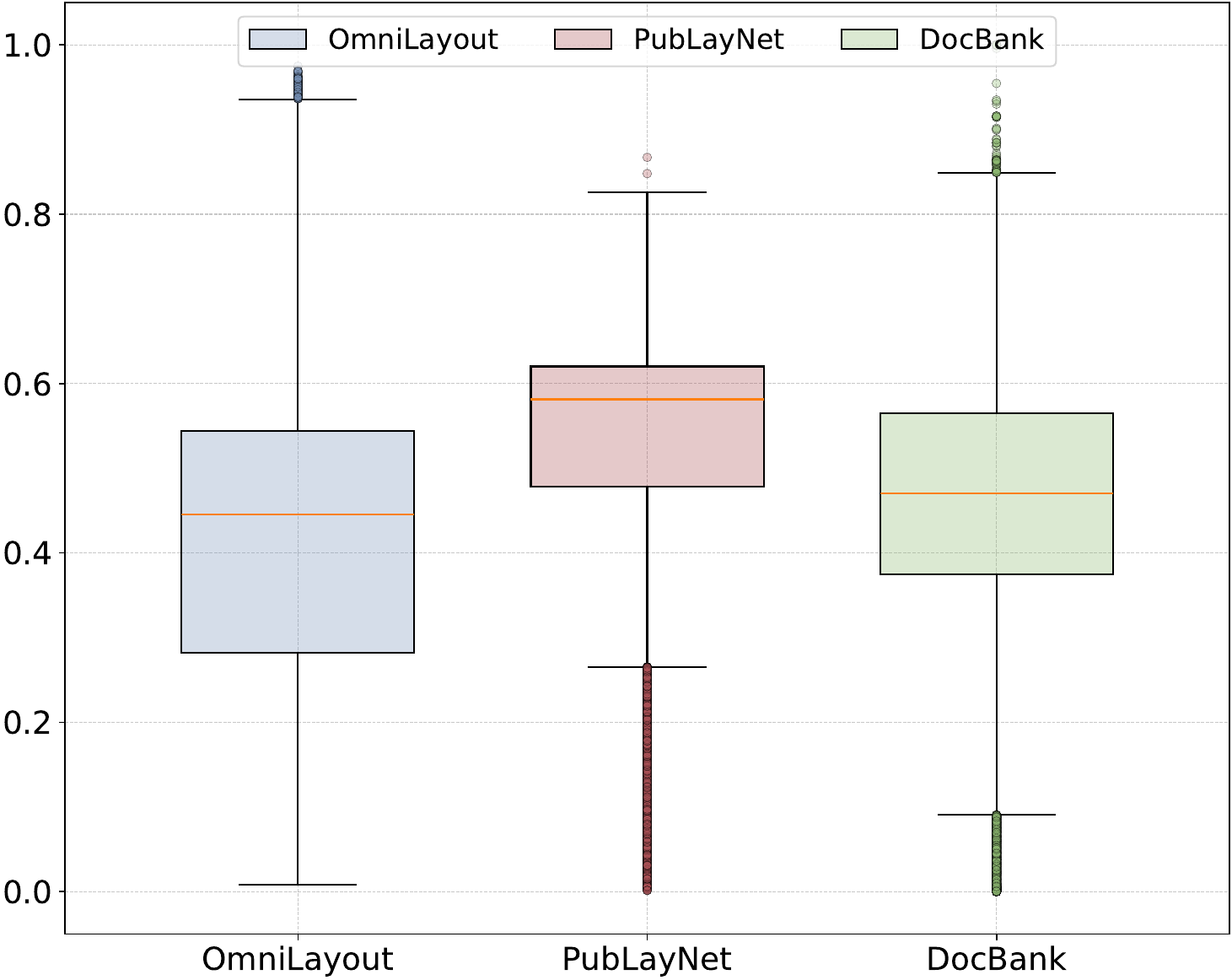}
        \caption{Area ratio per page.}
        \label{fig:sub2}
    \end{subfigure}
    \hfill 
    \begin{subfigure}[b]{0.32\textwidth}
        \centering
        \includegraphics[width=\textwidth]{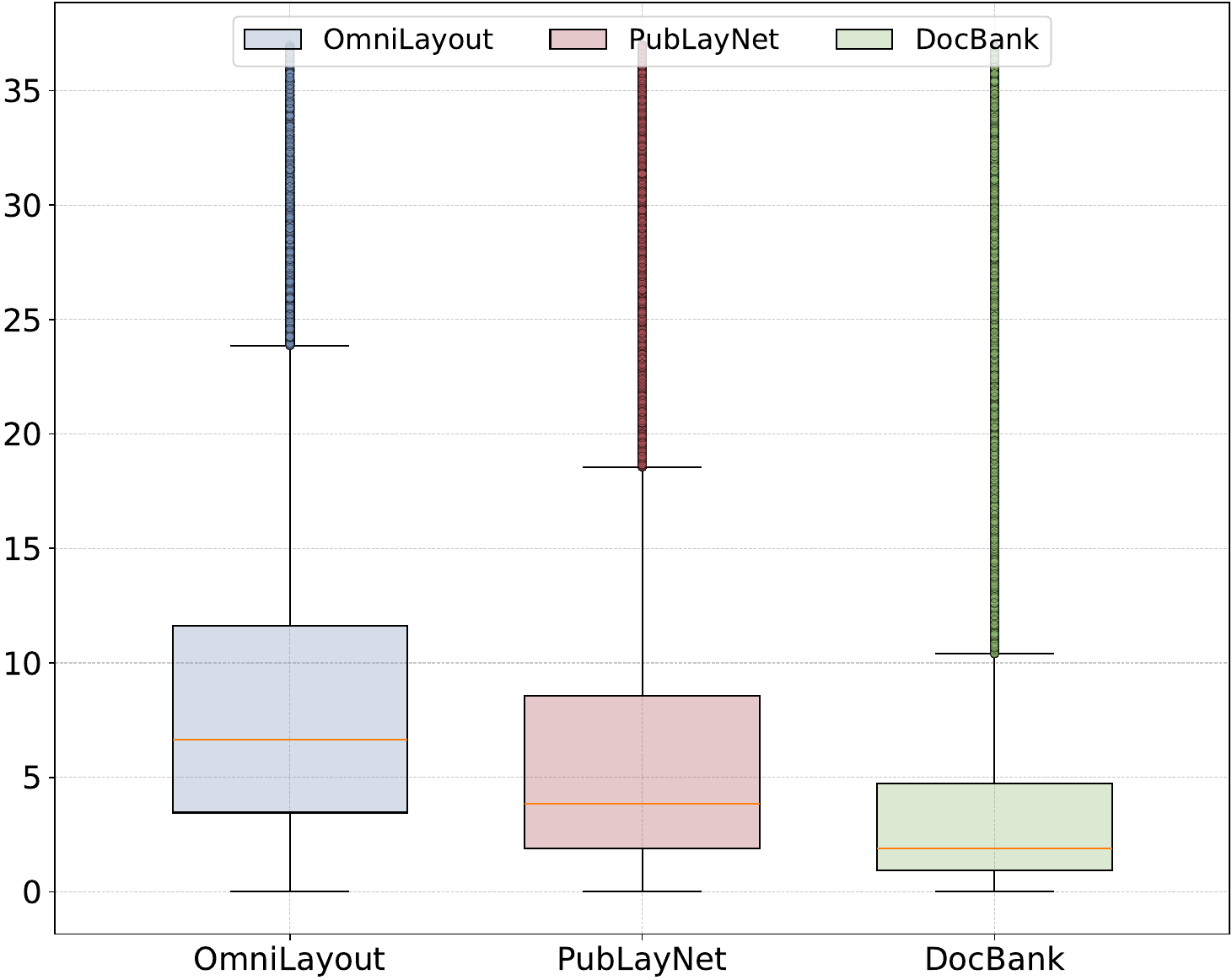}
        \caption{Overall element aspect Ratio.}
        \label{fig:sub3}
    \end{subfigure}
    \vspace{-2mm}
    \caption{Element-Level Statistical Comparison Between Our OmniDocLayout-1M and Two Widely-Used Datasets: PubLayNet and DocBank.}
    \label{fig:stat_element}
    \vspace{-1em}
\end{figure*}

\noindent\textbf{Setup.}
We randomly sample 1,200 document pages from the OmniDocLayout-1M dataset, selecting 200 pages for each document type (textbook, newspaper, magazine, exam, academic and slide), and obtain two sets of annotations for each page: (\emph{i}) human-annotated bounding boxes (GT), manually labeled by trained annotators following our specification, and (\emph{ii}) model-generated bounding boxes produced by our layout extraction model without post-processing. For every page, we construct a pair of annotations (A, B), where one corresponds to the human annotation and the other to the model output, ensuring fully unbiased comparisons, with their order randomized and hidden from evaluators.

\noindent\textbf{Evaluation Procedure.}
The evaluation interface displays the document page image alongside two independently rendered layout annotations (A and B). Annotators are asked to decide whether A is better, B is better, or both are good. Selecting “Both good” indicates that the two annotations are both accurate and visually aligned with the page structure.
All evaluators are familiar with document layout quality criteria but are not informed which annotation is the ground truth or the model prediction produced by MinerU~\citep{wang2024mineruopensourcesolutionprecise}. In total, four annotators participated in the study, and each sample is reviewed by two annotators, with disagreements resolved through majority voting.



\noindent\textbf{Result Analysis.}
Across all 1,200 evaluated pages, the distribution of votes in \cref{fig:human_eval} shows that 1,022 samples (85.2\%) fall into the “Both good” category, indicating that human and model-generated layouts are largely indistinguishable in perceptual quality. An additional 85 samples (7.1\%) are judged as “Model is better,” and together these outcomes account for over 92\% of cases where model-generated bounding boxes are considered at least as good as human annotations. These results demonstrate a high level of human–model consistency in layout quality.



\subsection{Element-level Statistical Analysis}
First, we analyze the diversity of OmniDocLayout-1M from the perspective of element distribution. Specifically, we examine element diversity in three aspects: the number of elements per page, the proportion of the layout area occupied by all elements on a single page, and the aspect ratios of the elements. The data distribution is illustrated in Fig.~\ref{fig:stat_element}. As can be observed, OmniDocLayout-1M exhibits significantly greater diversity in elements compared to PubLayNet~\citep{zhong2019publaynet} and DocBank~\citep{li2020docbank}. This ensures the robustness of the pre-trained model, enabling our proposed method to adapt to various element types (with different aspect ratios and categories) and diverse layout attributes (with varying densities and numbers of elements) in downstream tasks.

\subsection{Layout Diversity}
\begin{figure*}[b]
    \centering
    \begin{subfigure}{0.95\textwidth}
        \centering
        \includegraphics[width=0.95\textwidth]{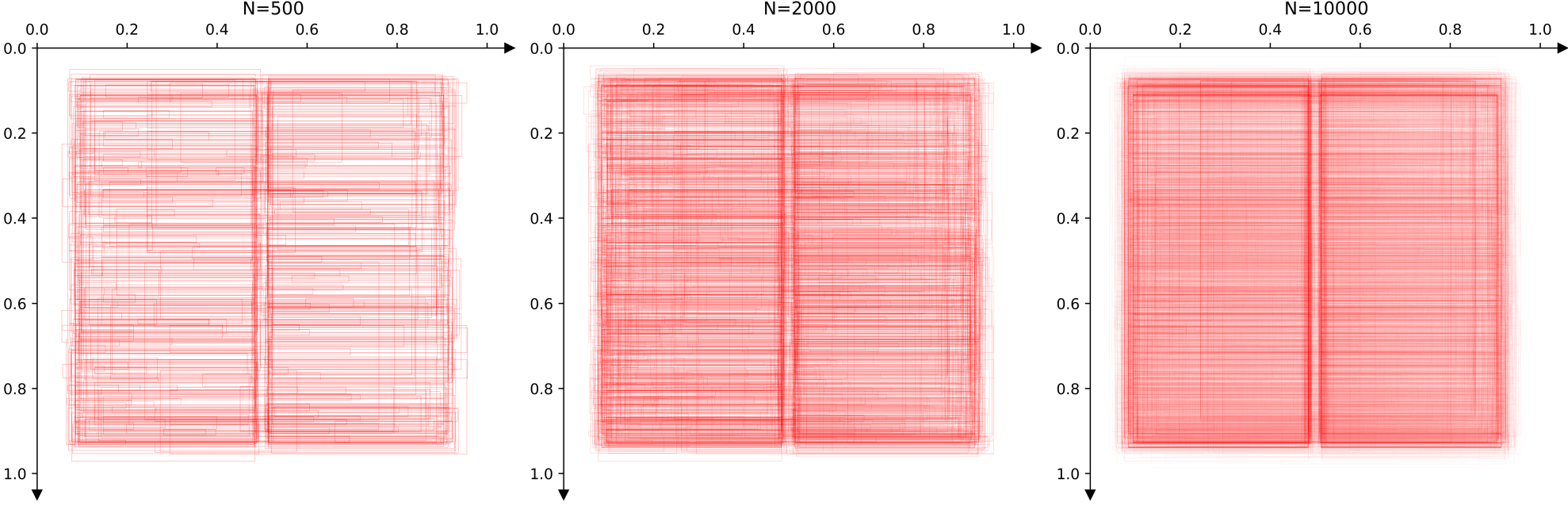}
        \caption{Document layout distribution of PubLayNet.}
        \label{fig:layout_pub}
    \end{subfigure}
    \begin{subfigure}{0.95\textwidth}
        \centering
        \includegraphics[width=0.95\textwidth]{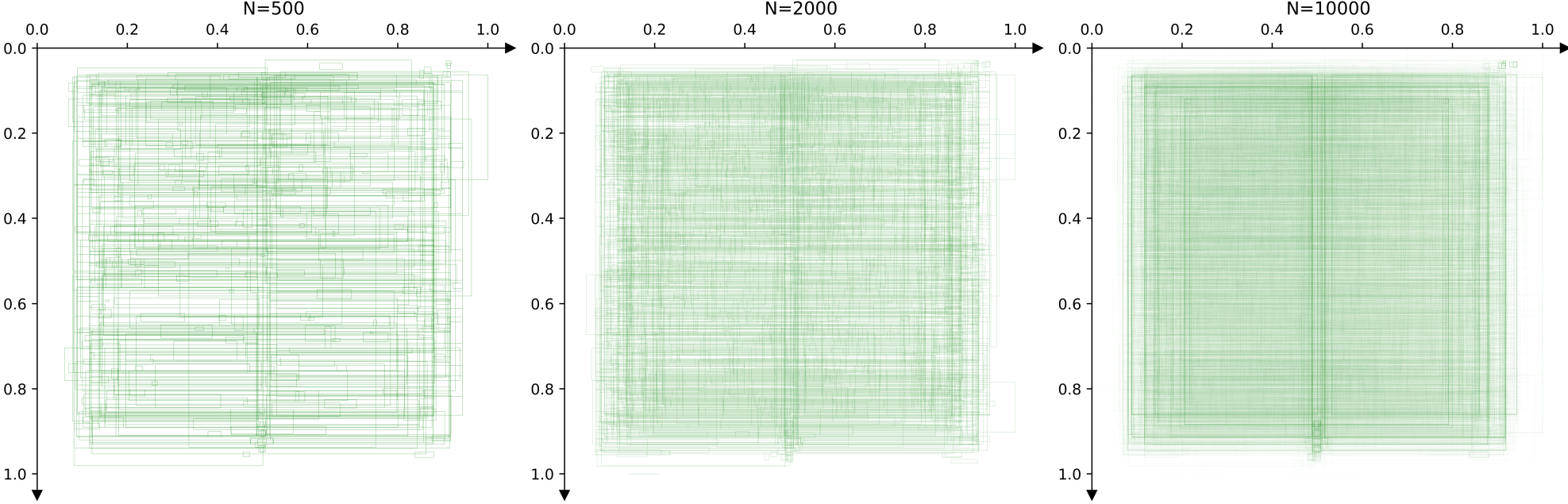}
        \caption{Document layout distribution of DocBank.}
        \label{fig:layout_docbank}
    \end{subfigure}
    \begin{subfigure}{0.95\textwidth}
        \centering
        \includegraphics[width=0.95\textwidth]{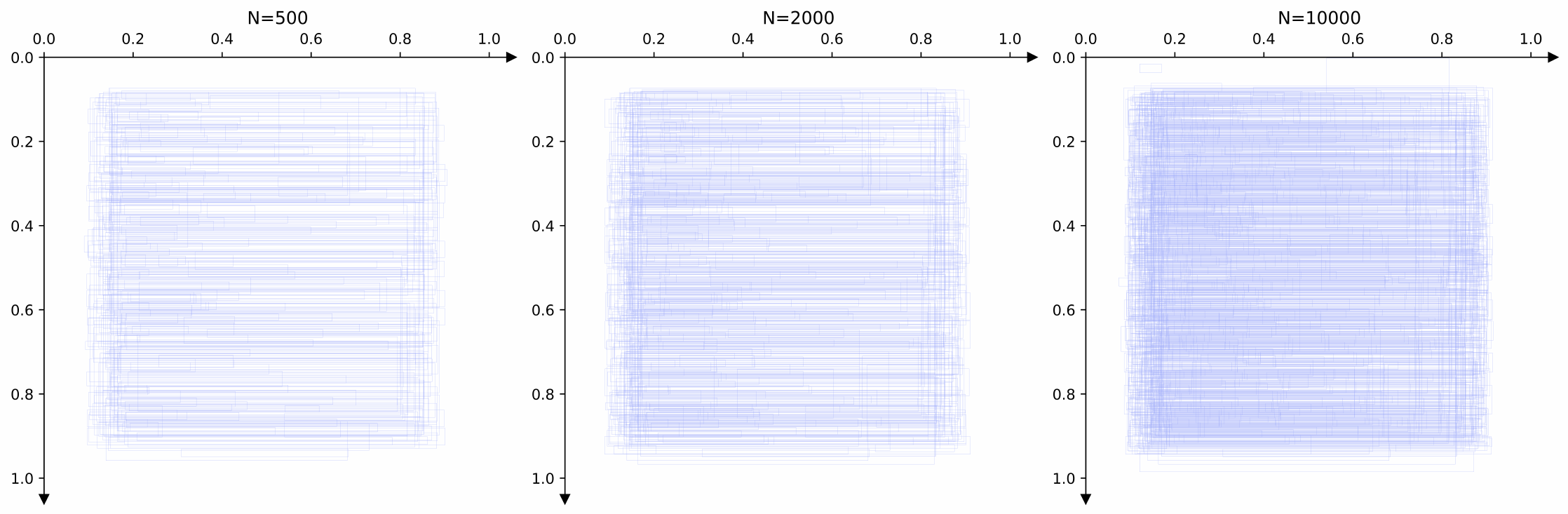}
        \caption{Document layout distribution of textbook in OmniDocLayout-1M.}
        \label{fig:layout_textbook}
    \end{subfigure}
    \begin{subfigure}{0.95\textwidth}
        \centering
        \includegraphics[width=0.95\textwidth]{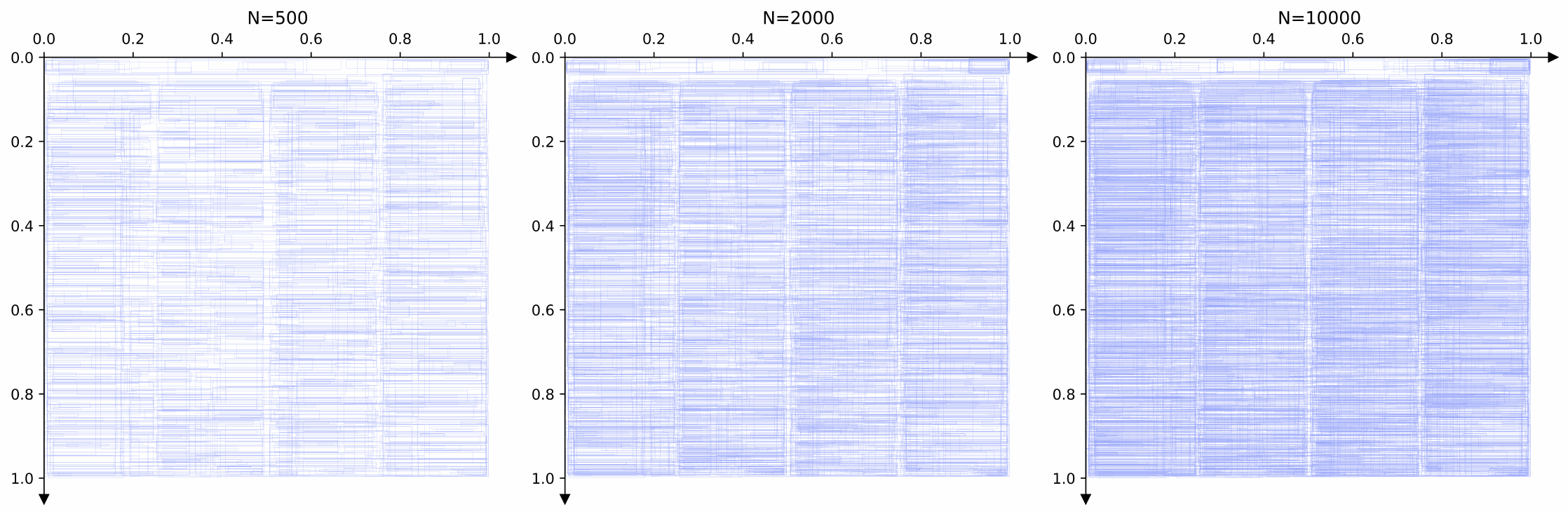}
        \caption{Document layout distribution of newspaper in OmniDocLayout-1M.}
        \label{fig:layout_pub}
    \end{subfigure}
    \vspace{-1mm}
    \caption{Document Layout Distributions of Two Widely-used Benchmarks, (a) PubLayNet and (b) DocBank, and Two Document Types from Our OmniDocLayout-1M Dataset, (c) Textbook and (d) Newspaper.}
\label{fig:layout_diversity1}
\end{figure*}

\begin{figure*}[t]
    \centering
    \begin{subfigure}{0.95\textwidth}
        \centering
        \includegraphics[width=0.95\textwidth]{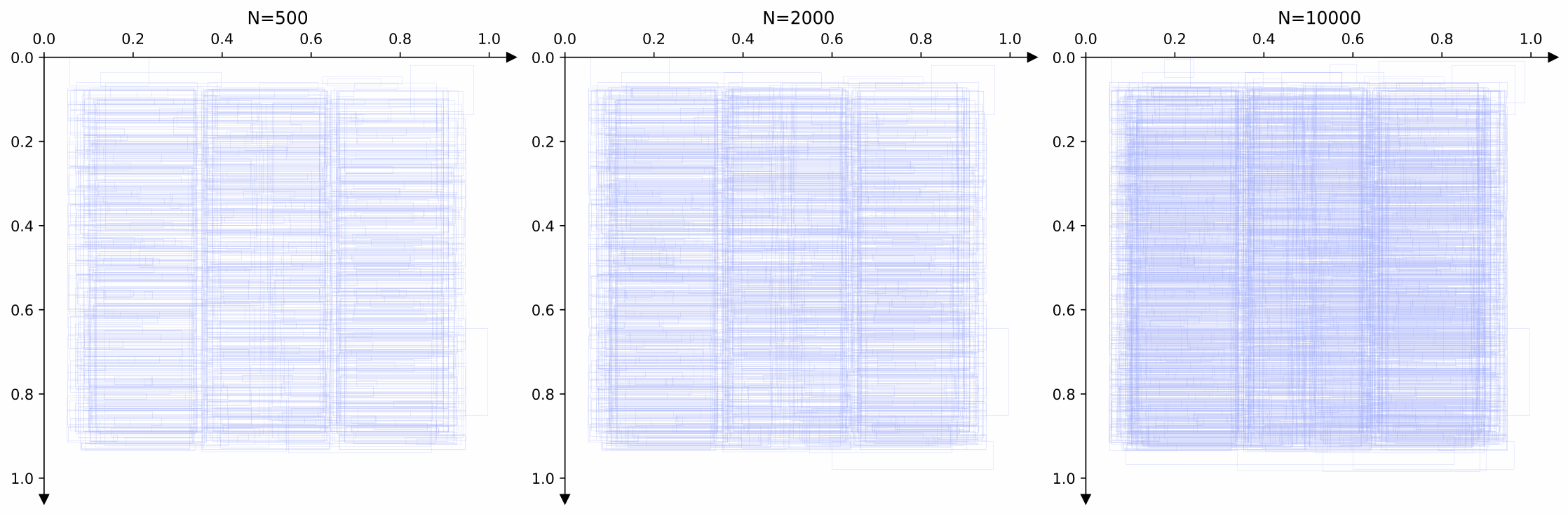}
        \caption{Document layout distribution of magazine in OmniDocLayout-1M.}
        \label{fig:layout_maga}
    \end{subfigure}
    \begin{subfigure}{0.95\textwidth}
        \centering
        \includegraphics[width=0.95\textwidth]{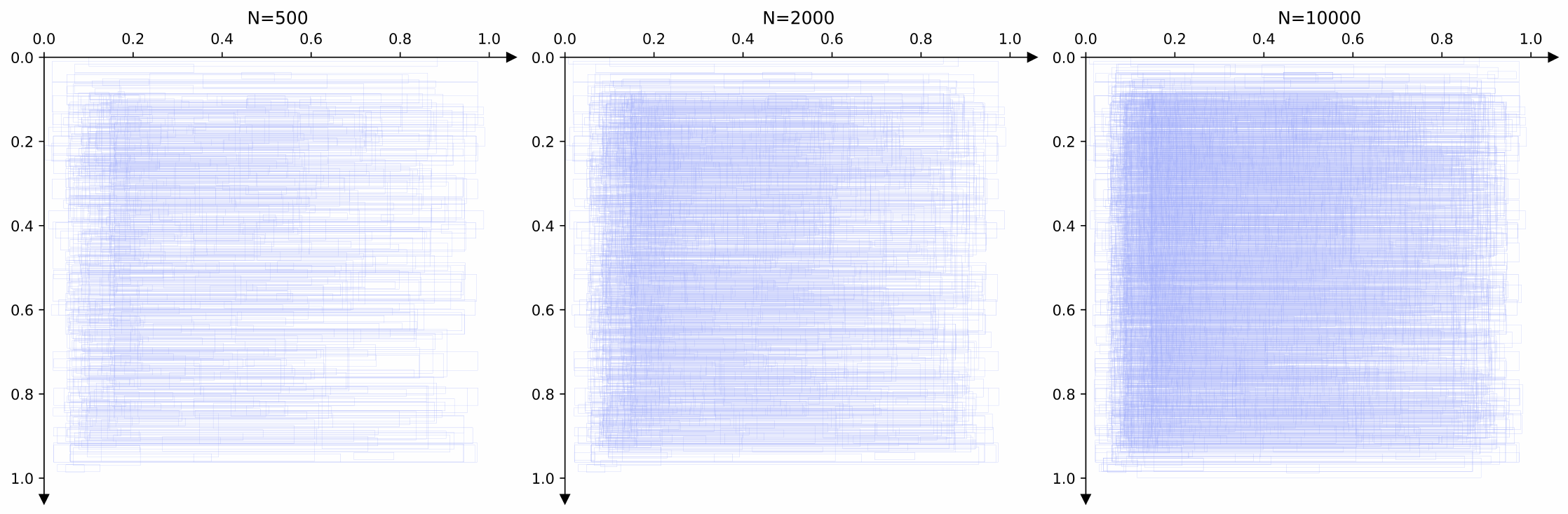}
        \caption{Document layout distribution of exam in OmniDocLayout-1M.}
        \label{fig:layout_exam}
    \end{subfigure}
    \begin{subfigure}{0.95\textwidth}
        \centering
        \includegraphics[width=0.95\textwidth]{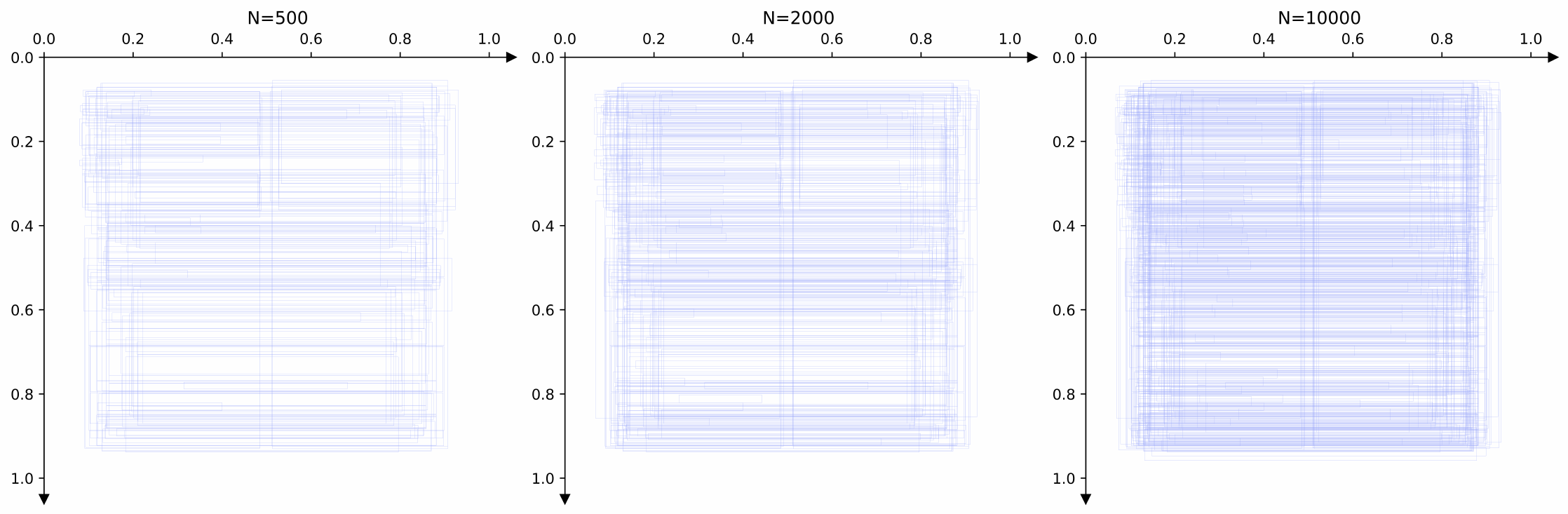}
        \caption{Document layout distribution of academic in OmniDocLayout-1M.}
        \label{fig:layout_aca}
    \end{subfigure}
    \begin{subfigure}{0.95\textwidth}
        \centering
        \includegraphics[width=0.95\textwidth]{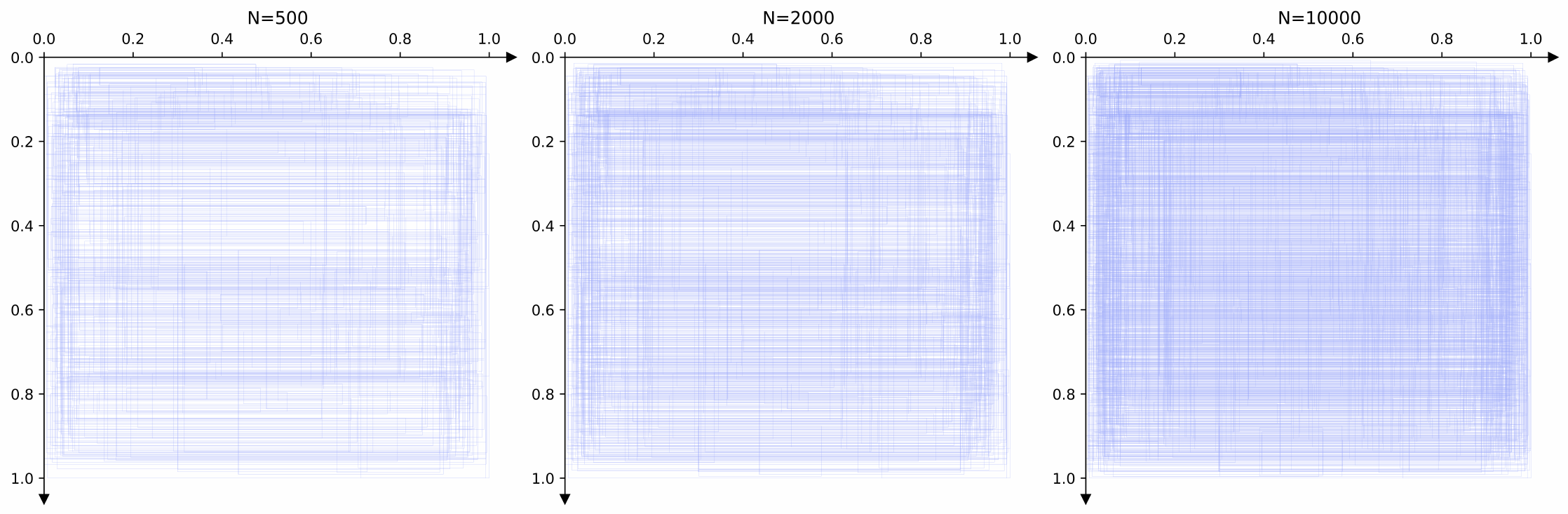}
        \caption{Document layout distribution of slide in OmniDocLayout-1M.}
        \label{fig:layout_slide}
    \end{subfigure}
    \vspace{-1mm}
    \caption{{Document Layout Distributions of (a) Magazine, (b) Exam, (c) Academic, and (d) Slide in Our OmniDocLayout-1M Dataset.}}
\label{fig:layout_diversity2}
\end{figure*}

Next, we visualize and compare the document layout diversity of PubLayNet, DocBank, and OmniDocLayout-1M as shown in Fig.~\ref{fig:layout_diversity1} and Fig.~\ref{fig:layout_diversity2}. $N$ indicates the number of documents used for visualization. Compared with two-column format and Manhattan layout typical of academic papers in PubLayNet or DocBank, document layout in OmniDocLayout-1M significant variation and diversity.

\subsection{More Visual Examples}
\begin{figure*}
    \centering
    \includegraphics[width=0.70\textwidth]{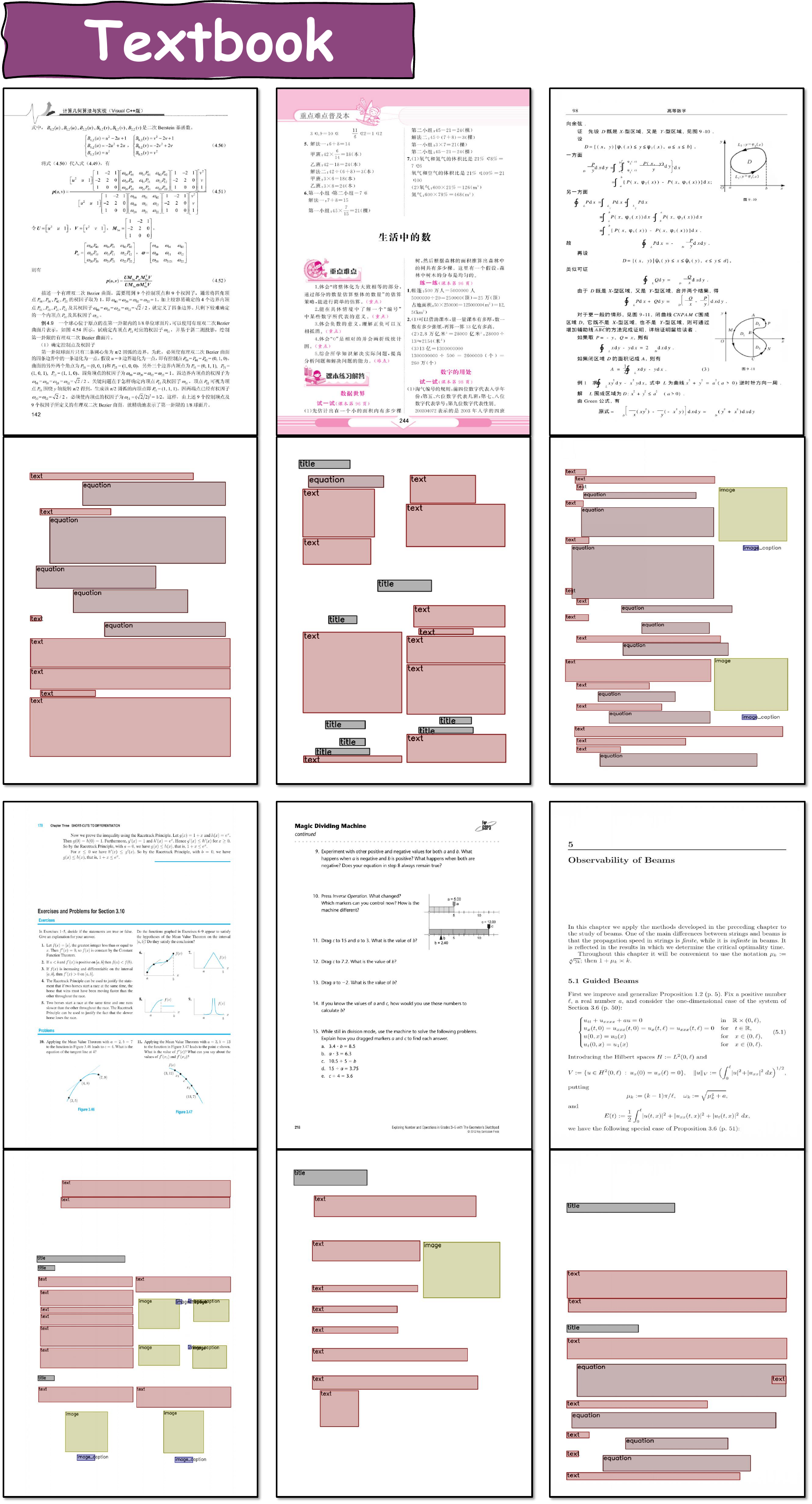}
    \caption{Example Textbook Layouts from Our OmniDocLayout-1M Dataset.}
    \label{fig:dataset_vis_textbook}
\end{figure*}

\begin{figure*}
    \centering
    \includegraphics[width=0.72\textwidth]{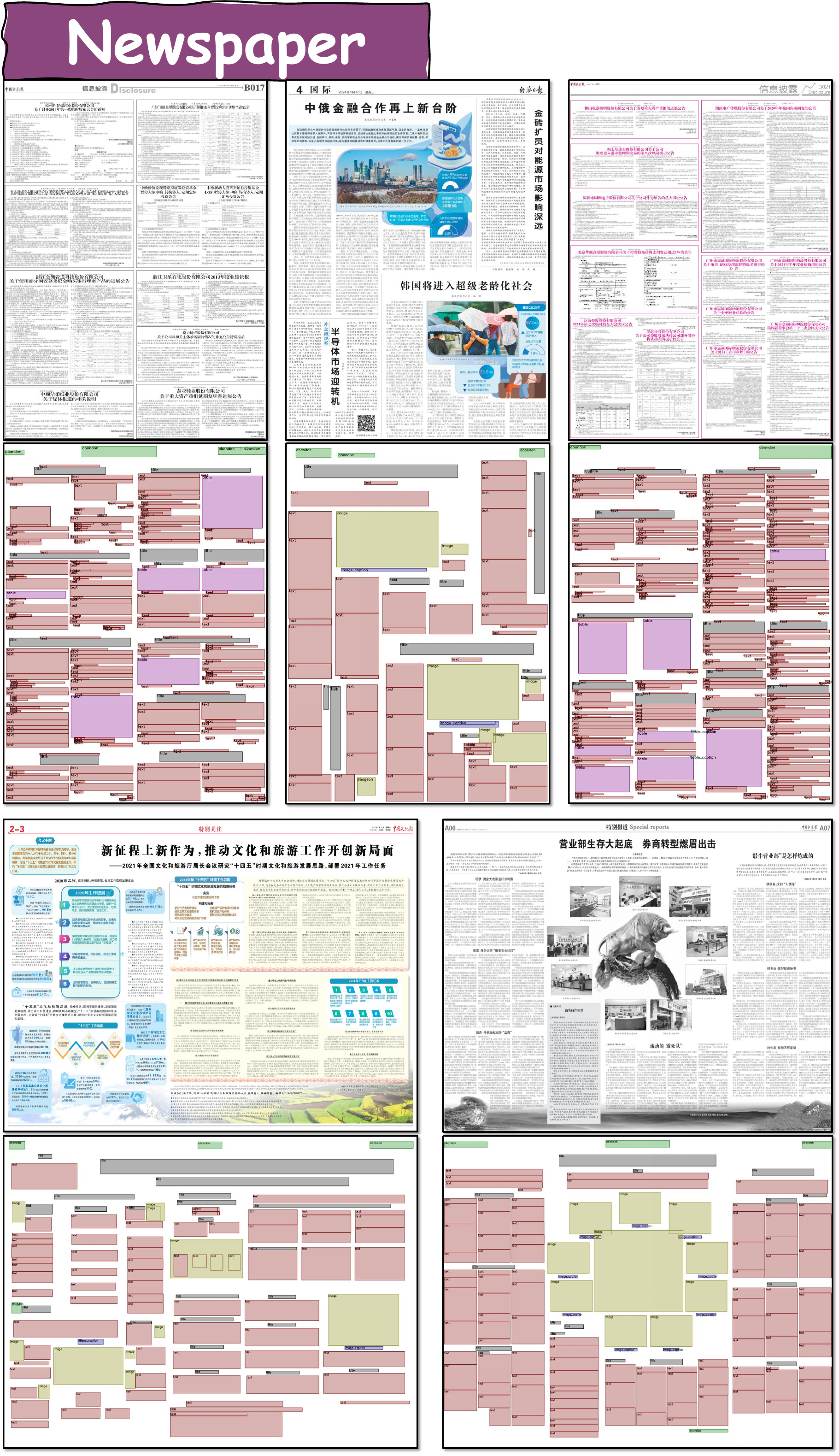}
    \caption{Example Newspaper Layouts from Our OmniDocLayout-1M Dataset.}
    \label{fig:dataset_vis_news}
\end{figure*}

\begin{figure*}
    \centering
    \includegraphics[width=0.70\textwidth]{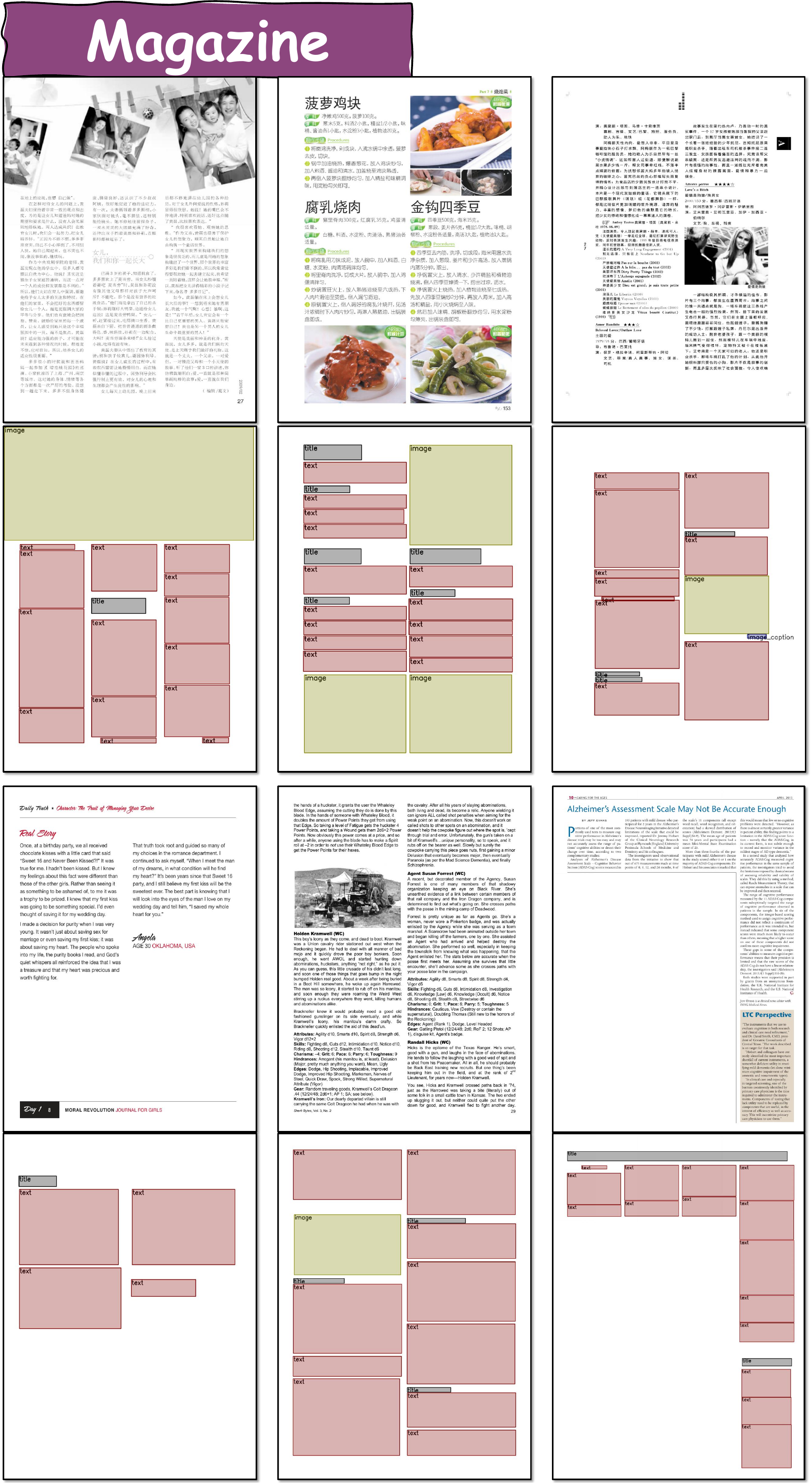}
    \caption{Example Magazine Layouts from Our OmniDocLayout-1M Dataset.}
    \label{fig:dataset_vis_maga}
\end{figure*}

\begin{figure*}
    \centering
    \includegraphics[width=0.71\textwidth]{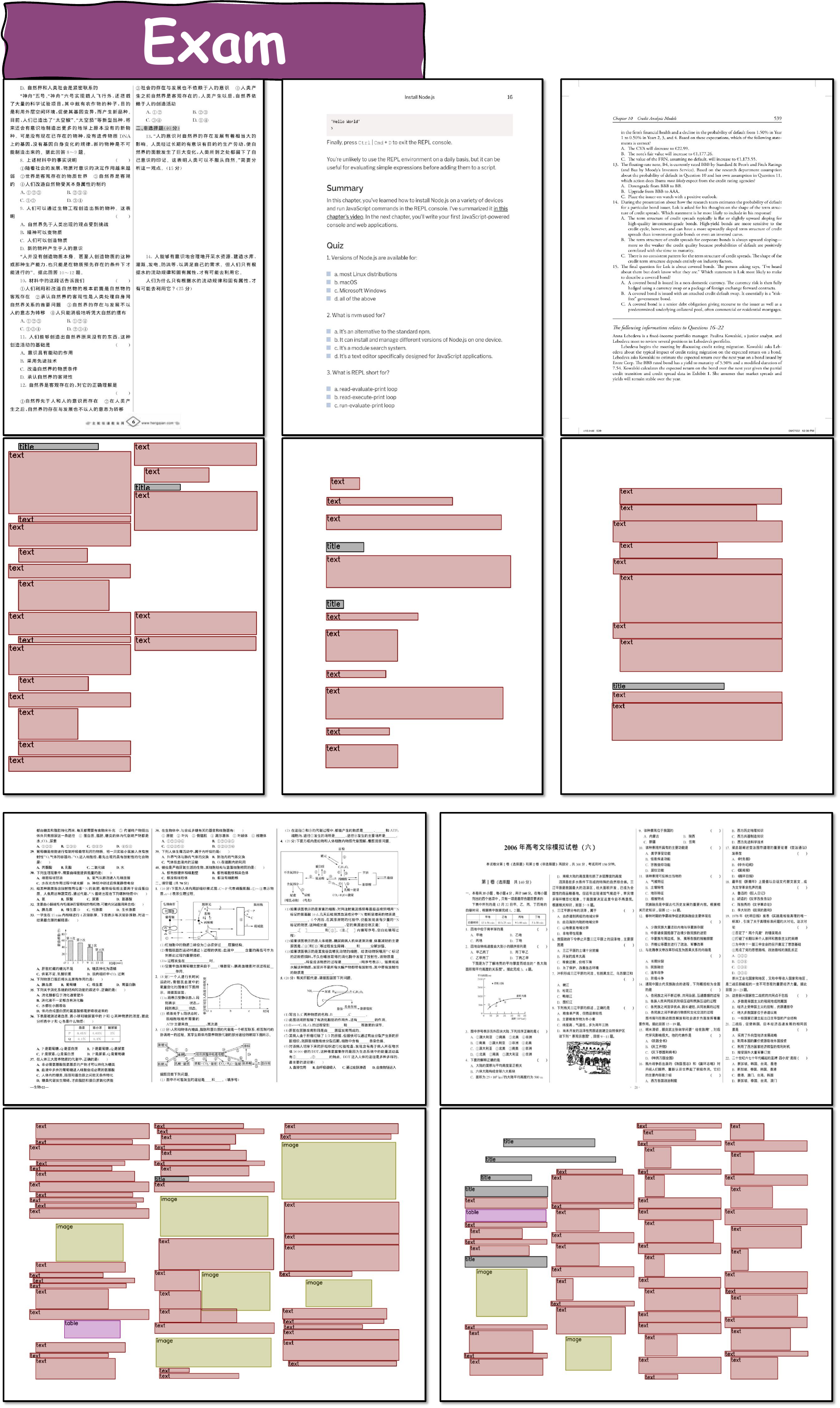}
    \caption{Example Exam Layouts from Our OmniDocLayout-1M Dataset.}
    \label{fig:dataset_vis_exam}
\end{figure*}

\begin{figure*}
    \centering
    \includegraphics[width=0.70\textwidth]{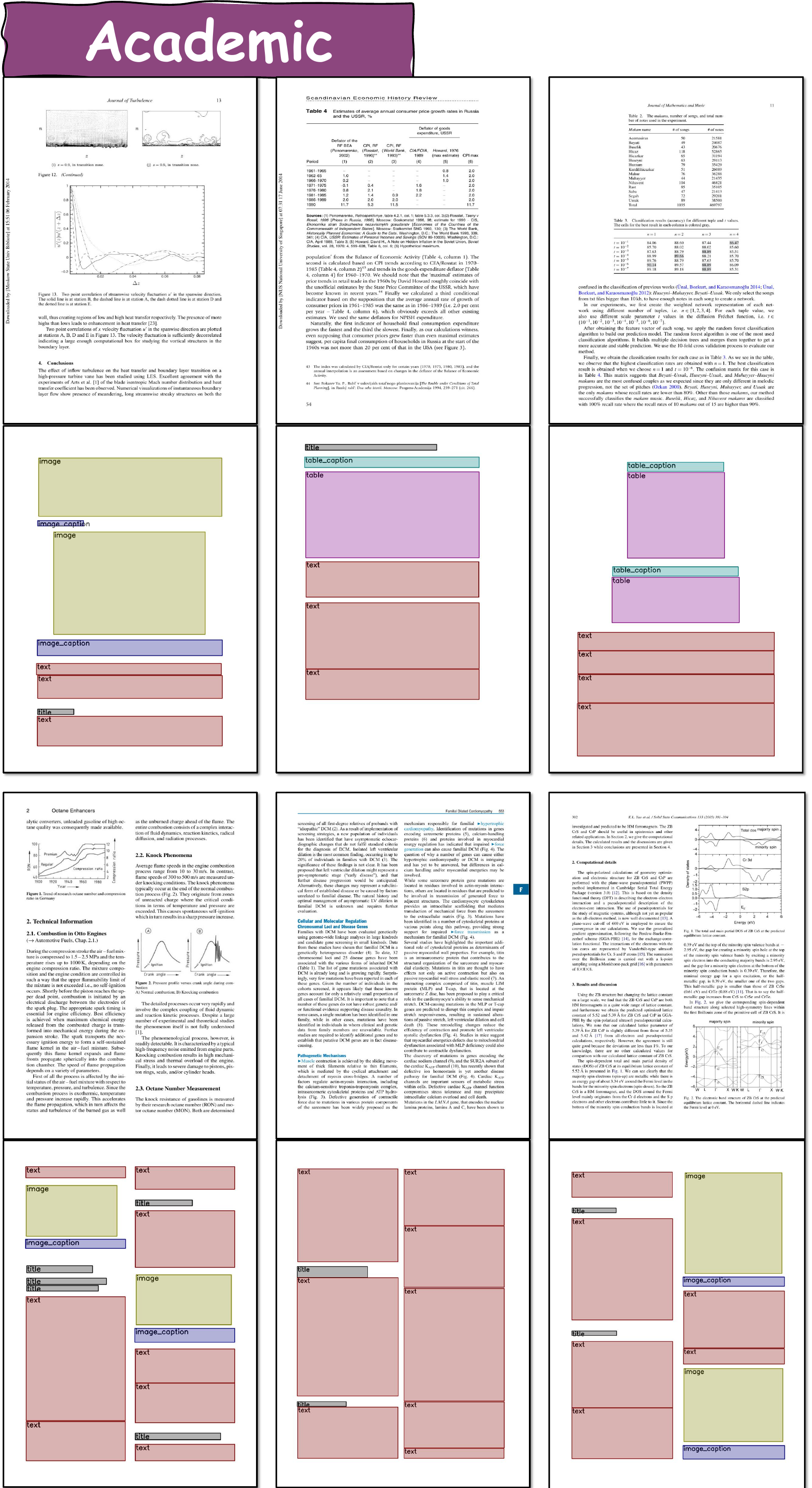}
    \caption{Example Academic Layouts from Our OmniDocLayout-1M Dataset.}
    \label{fig:dataset_vis_academic}
\end{figure*}

\begin{figure*}
    \centering
    \includegraphics[width=0.74\textwidth]{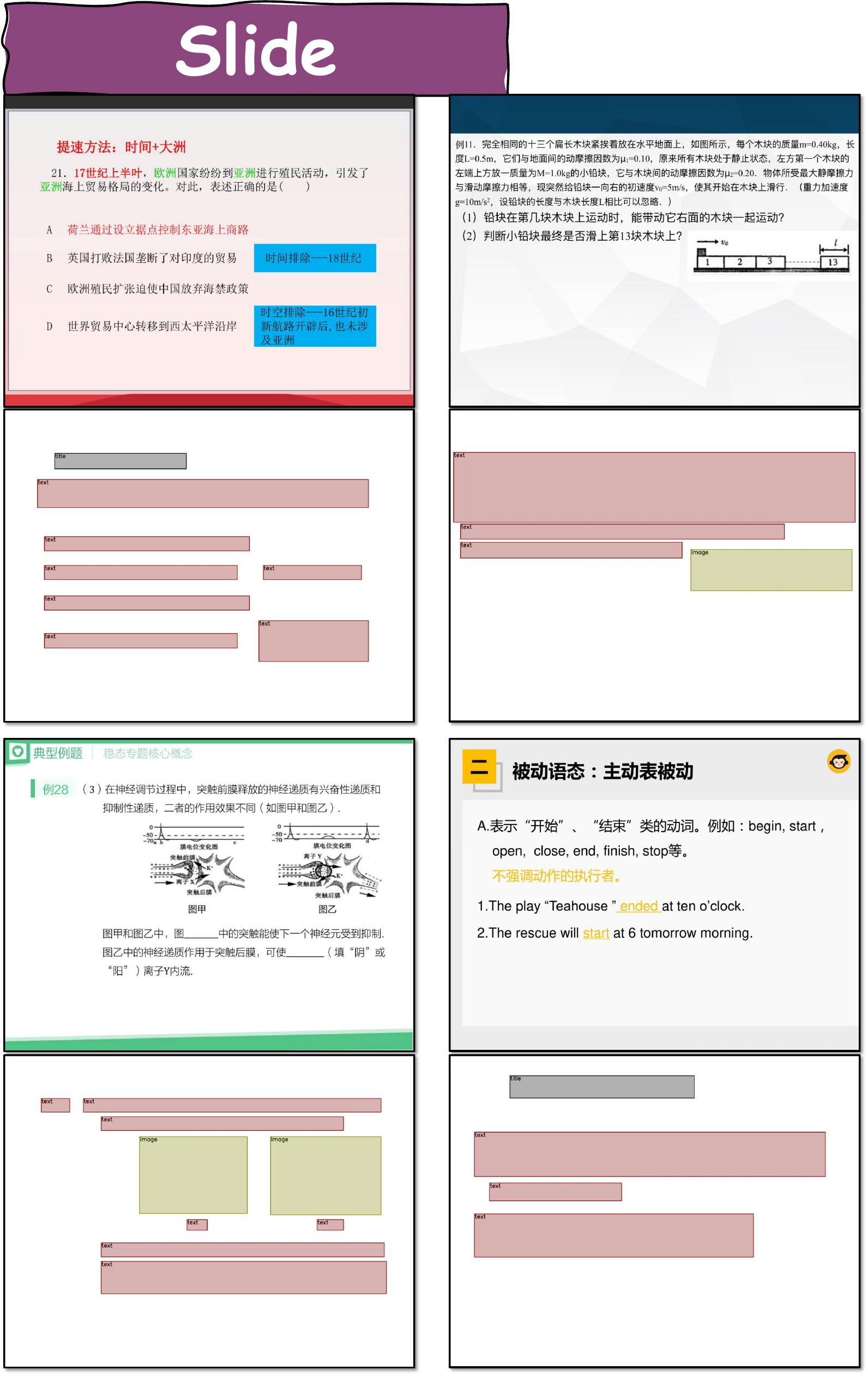}
    \caption{Example Slide Layouts from Our OmniDocLayout-1M Dataset.}
    \label{fig:dataset_vis_slide}
\end{figure*}

In this section, we present more visual examples from our OmniDocLayout-1M dataset, accompanied by high-quality annotations extracted with MinerU~\citep{wang2024mineruopensourcesolutionprecise}. We visualize annotated examples of six document types: textbook~(Fig.~\ref{fig:dataset_vis_textbook}), newspaper~(Fig.~\ref{fig:dataset_vis_news}), magazine~(Fig.~\ref{fig:dataset_vis_maga}), exam~(Fig.~\ref{fig:dataset_vis_exam}), academic~(Fig.~\ref{fig:dataset_vis_academic}), slide~(Fig.~\ref{fig:dataset_vis_slide}) are shown.

\section{OmniDocLayout-LLM Analysis}\label{app:model}

\subsection{Few-shot Performance of General-purpose LLMs}
Due to the page limit, the complete 0/1/5-shot comparison results are reported in Table~\ref{tab:complete_LLMs_comparision} of the appendix. In particular, evaluation of complex document layouts requires significantly longer inference time and more than 10,000 USD in API costs owing to excessive sequence length.

\noindent \textbf{General-purpose LLMs will overtake?} 
Experimental results have shown improved performance with increasing numbers of in-context examples, raising concerns about whether future versions of such models, or simply scaling the number of shots, could replace specialized models in this scenario. It is a challenge faced by nearly all task-specific research areas. However, fully relying on general-purpose LLMs or in-context learning to take over document layout generation remains difficult for two key reasons: 
(1) Lack of task-specific pretraining and limited spatial reasoning.
Our zero-shot experiments show that general models often produce layouts with extensive overlaps, even when explicitly instructed to avoid them. This indicates that document layout generation is still a niche task, and current or future general models are unlikely to undergo targeted pretraining for this domain. Furthermore, these models exhibit limited spatial reasoning capabilities, which are essential for generating well-organized layouts.
(2) Unacceptable inference-time and financial overhead when scaling shots.
Increasing the number of in-context examples does not scale effectively in practice. This is because each complex layout will be serialized into a long sequence, and a 5-shot prompt already results in substantial token length, inference latency, and API cost. Scaling to 10 shots or more becomes impractical due to both inference time and financial cost.
Together, these observations indicate that, while general-purpose LLMs show promising flexibility, they are unlikely to replace domain-specific approaches for document layout generation in the foreseeable future.

\subsection{More Generated Layouts}

\begin{figure*}[h]
    \centering
    \includegraphics[width=0.9\textwidth]{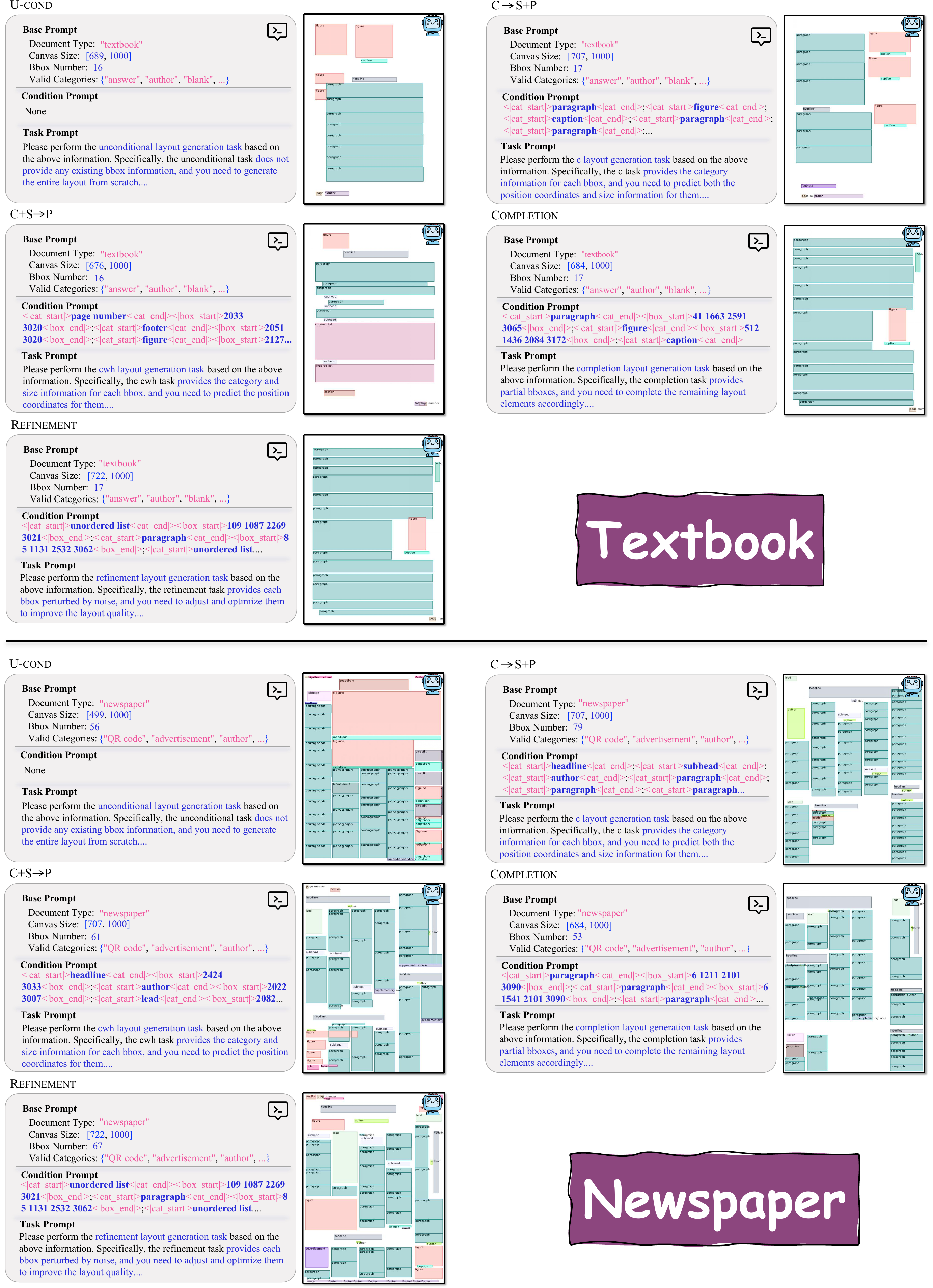}
    \caption{Examples of Layouts Generated by Our OmniDocLayout-LLM on Five Generation Tasks~(Textbook and Newspaper).}
    \label{fig:lggpt1}
\end{figure*}

\begin{figure*}[h]
    \centering
    \includegraphics[width=0.9\textwidth]{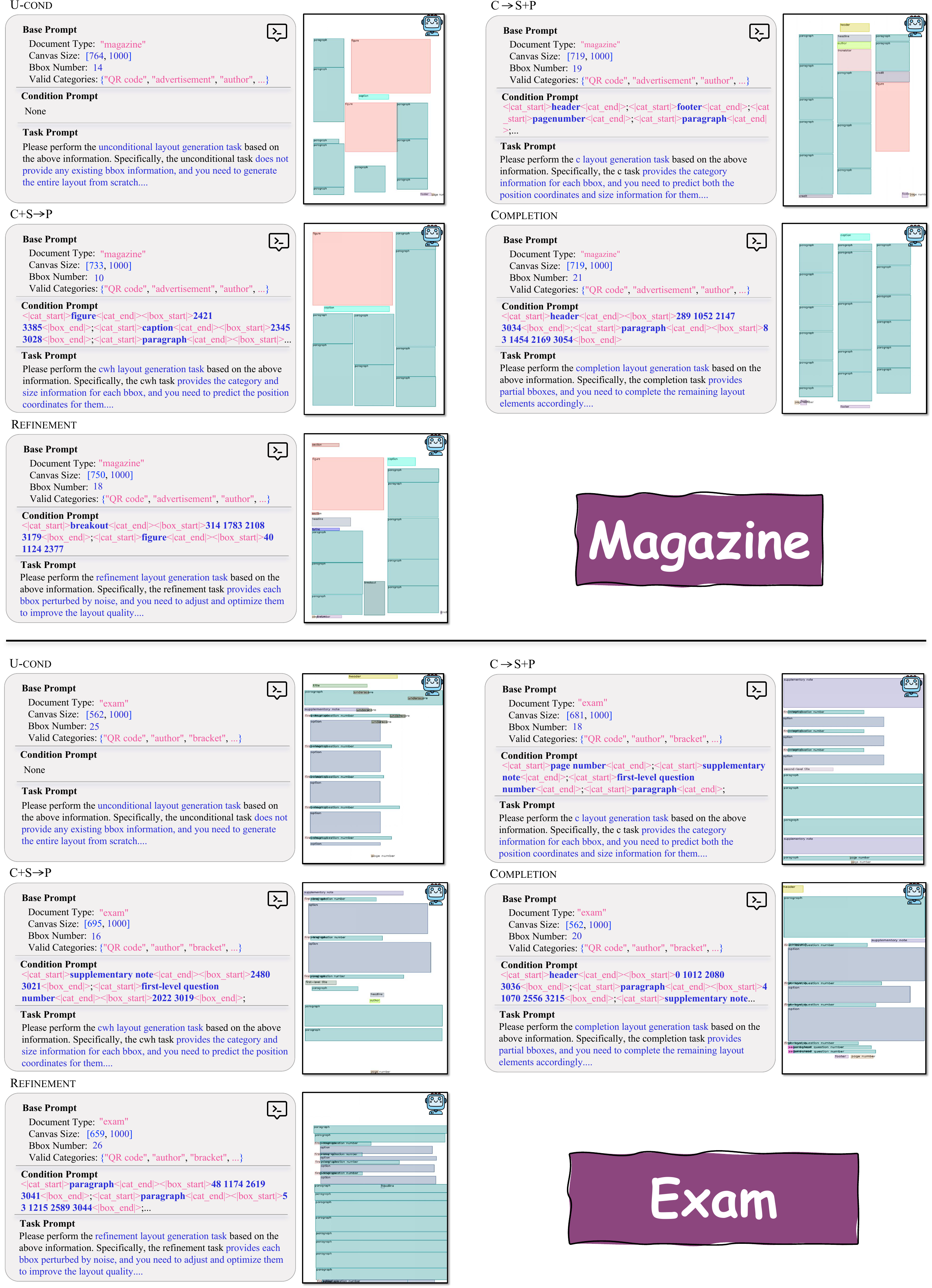}
    \caption{Examples of Layouts Generated by Our OmniDocLayout-LLM on Five Generation Tasks~(Magazine and Exam).}
    \label{fig:lggpt2}
\end{figure*}

\begin{figure*}[t]
    \centering
    \includegraphics[width=0.9\textwidth]{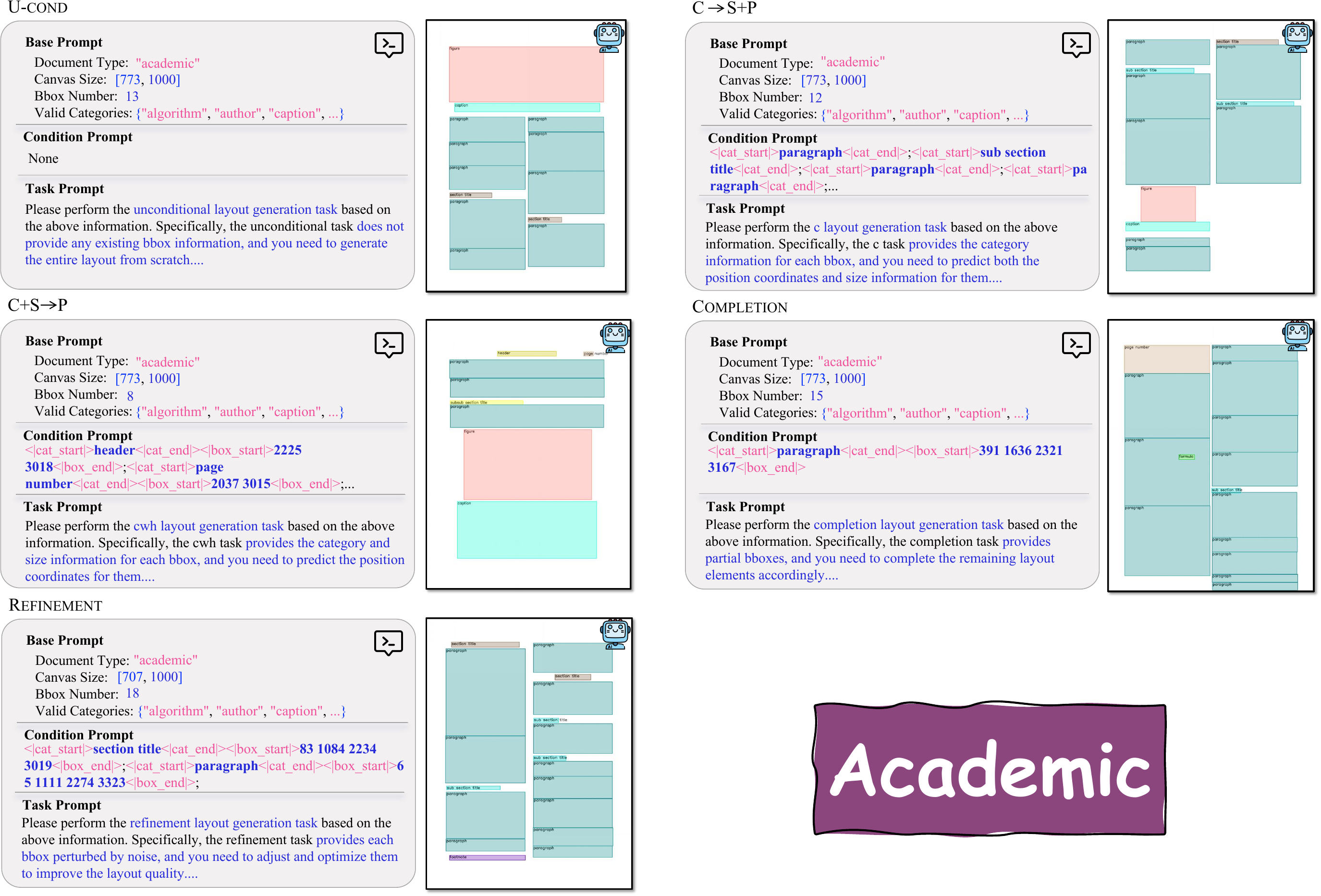}
    \caption{Examples of Layouts Generated by Our OmniDocLayout-LLM on Five Generation Tasks~(Academic).}
    \label{fig:lggpt3}
\end{figure*}

In this section, we demonstrate more qualitative examples generated by OmniDocLayout-LLM in five generation tasks: U-Cond, C$\rightarrow$S+P, C+S$\rightarrow$P, Complement and Refinement. Using M\textsuperscript{6}Doc as the test set, we show the generated layouts in Fig.~\ref{fig:lggpt1} for textbooks and newspapers, Fig.~\ref{fig:lggpt2} for magazines and exams, and Fig.~\ref{fig:lggpt3} for academic papers. The visualization results indicate that our model can generate both reasonable and aesthetically pleasing layouts for diverse and complicated document types. Furthermore, it effectively adheres to the requirements of different generation tasks and adapts well to various constraints, showcasing its ability to perform under diverse conditions and tasks.

\renewcommand{\multirowsetup}{\centering}
\definecolor{mygray}{gray}{.92}
\definecolor{ForestGreen}{RGB}{34,139,34}
\definecolor{Forestred}{RGB}{220,50,50}
\begin{table*}[!t]
    \centering
    \caption{Comparison with Powerful General-purpose LLMs in 0/1/5-shot Setting Across Five Document Types in M\textsuperscript{6}Doc.}
    \vspace{-1mm}
    \belowrulesep=-0.25pt
    \aboverulesep=-0.25pt
    \setlength{\tabcolsep}{3pt}
    \renewcommand{\arraystretch}{1.6}
    \footnotesize
	\centering
    \resizebox{1.0\textwidth}{!}{
        \begin{tabular}{
          >{\centering\arraybackslash}p{1cm} 
          >{\centering\arraybackslash}p{2.5cm}
          >{\centering\arraybackslash}p{1.2cm}
          *{20}{>{\centering\arraybackslash}p{0.9cm}}
        }
            \toprule[1.2pt]
            \multirow{2}{*}{\textbf{Task}} & \multirow{2}{*}{\textbf{Method}} & \multirow{2}{*}{\textbf{Setting}} &\multicolumn{4}{c}{\textbf{Textbook}} & \multicolumn{4}{c}{\textbf{Newspaper}} & \multicolumn{4}{c}{\textbf{Magazine}} & \multicolumn{4}{c}{\textbf{Exam}} & \multicolumn{4}{c}{\textbf{Academic}} \\
            \cmidrule(lr){4-7} \cmidrule(lr){8-11} \cmidrule(lr){12-15} \cmidrule(lr){16-19} \cmidrule(lr){20-23}
            & & &
            FID$\downarrow$ & Ali.$\rightarrow$ & Ove.$\rightarrow$ & mIoU$\uparrow$ &
            FID$\downarrow$ & Ali.$\rightarrow$ & Ove.$\rightarrow$ & mIoU$\uparrow$ &
            FID$\downarrow$ & Ali.$\rightarrow$ & Ove.$\rightarrow$ & mIoU$\uparrow$ &
            FID$\downarrow$ & Ali.$\rightarrow$ & Ove.$\rightarrow$ & mIoU$\uparrow$ &
            FID$\downarrow$ & Ali.$\rightarrow$ & Ove.$\rightarrow$ & mIoU$\uparrow$ \\

            \hline
            \multirow{10}{*}{\textbf{U-Cond}} & \multirow{3}{*}{GPT-4o} & 0-shot & 135.32 & 0.017 & \textbf{0.007} & 0.060 & 193.13 & 0.020 & 0.007 & 0.000 & 236.11 & 0.015 & \textbf{0.040} & 0.089 & 163.94 & 0.020 & 0.010 & 0.000 & 135.60 & 0.006 & 0.006 & 0.000 \\

            &  & 1-shot & 97.34 & 0.105 & 0.099 & 0.061 & 100.18 & 0.031 & \textbf{0.052} & 0.000 & 177.61 & 0.038 & 0.215 & 0.158 & 111.87 & 0.043 & 0.049 & 0.000 & 100.70 & 0.023 & 0.011 & 0.000 \\

            &  & 5-shot & 71.56 & 0.149 & \underline{0.083} & 0.177 & 54.34 & 0.032 & \underline{0.074} & 0.000 & 143.70 & 0.051 & 0.198 & 0.174 & 76.48 & 0.049 & 0.085 & 0.000 & 90.93 & 0.060 & 0.024 & \textbf{0.460} \\

            \cmidrule(lr){2-23}
            
            & \multirow{3}{*}{Gemini-2.5*} & 0-shot & 147.88 & \textbf{0.264} & 6.490 & 0.154 & 194.77 & 0.078 & 0.098 & 0.000 & 118.78 & 0.041 & 0.213 & 0.226 & 140.48 & \textbf{0.071} & \underline{0.313} & 0.000 & 57.36 & 0.347 & 0.089 & 0.000 \\

            & & 1-shot & 84.16 & 0.159 & 0.552 & 0.221 & 74.14 & 0.015 & 0.098 & 0.000 & 88.91 & \underline{0.063} & 0.152 & 0.214 & 90.10 & 0.037 & \textbf{0.299} & 0.000 & 82.93 & \underline{0.107} & 0.054 & 0.000 \\

            & & 5-shot & \underline{57.30} & 0.173 & 0.219 & 0.202 & \underline{30.55} & \textbf{0.014} & 0.094 & 0.000 & \underline{66.30} & 0.190 & 0.136 & \textbf{0.289} & 70.05 & 0.038 & 0.086 & 0.000 & \underline{51.41} & 0.074 & 0.061 & 0.322 \\

            \cmidrule(lr){2-23}

            & \multirow{3}{*}{Claude-3.7*} & 0-shot & 96.23 & 0.102 & 0.145 & \underline{0.236} & 171.01 & 0.079 & 0.031 & 0.000 & 165.76 & 0.030 & 0.182 & 0.000 & 114.90 & 0.014 & 0.038 & 0.000 & 106.98 & 0.030 & \underline{0.101} & 0.000 \\

            & & 1-shot & 87.87 & 0.096 & 0.164 & 0.171 & 73.78 & 0.003 & 0.512 & 0.000 & 141.53 & 0.023 & 0.251 & 0.000 & 72.29 & 0.025 & 0.120 & 0.000 & 100.70 & 0.049 & \textbf{0.112} & 0.000 \\

            & & 5-shot & 59.73 & \underline{0.093} & 0.114 & 0.091 & \textbf{25.57} & 0.007 & 0.245 & 0.000 & 84.23 & 0.044 & 0.093 & 0.165 & \underline{50.83} & 0.043 & 0.131 & 0.000 & 75.50 & 0.042 & 0.098 & 0.064 \\

            \cmidrule(lr){2-23}
            
            \rowcolor{LightGreen}
            & Ours & - & \textbf{40.28} & \underline{0.219} & 0.102 & \textbf{0.288} & 39.73 & \underline{0.015} & 0.084 & 0.000 & \textbf{41.82} & \textbf{0.089} & 0.151 & \underline{0.266} & \textbf{40.32} & \underline{0.072} & 0.182 & \textbf{0.236} & \textbf{36.48} & \textbf{0.089} & 0.062 & \underline{0.415} \\
            \hline

            \multirow{10}{*}{\textbf{C$\rightarrow$S+P}} & \multirow{3}{*}{GPT-4o} & 0-shot & 103.15 & 0.084 & 0.072 & 0.119 & 202.84 & 0.002 & 0.112 & 0.028 & 165.36 & 0.055 & 0.164 & 0.097 & 107.17 & 0.040 & 0.049 & 0.082 & 90.67 & 0.224 & 0.027 & 0.123 \\

            & & 1-shot & 72.34 & 0.111 & \textbf{0.059} & 0.115 & 152.61 & 0.006 & 0.230 & 0.043 & 134.33 & 0.108 & 0.178 & 0.102 & 58.18 & 0.030 & 0.054 & 0.087 & 64.51 & 0.166 & 0.040 & 0.125 \\

            & & 5-shot & 51.50 & 0.146 & 0.074 & 0.118 & 91.87 & \textbf{0.012} & 0.312 & 0.059 & 104.85 & \underline{0.112} & 0.193 & 0.110 & 31.96 & 0.040 & 0.087 & 0.091 & 41.54 & 0.169 & 0.030 & 0.142 \\

            \cmidrule(lr){2-23}

            & \multirow{3}{*}{Gemini-2.5*} & 0-shot & 54.74 & 0.175 & \underline{0.060} & 0.078 & 69.40 & 0.569 & 0.666 & 0.070 & 53.32 & 0.065 & \textbf{0.104} & 0.101 & 42.25 & 0.053 & 0.063 & 0.036 & 31.79 & 0.351 & 0.051 & 0.084 \\

            & & 1-shot & 42.26 & \underline{0.217} & 0.114 & 0.089 & 37.62 & 0.005 & 0.750 & 0.073 & 53.72 & 0.057 & 0.240 & 0.110 & 33.10 & \textbf{0.059} & 0.169 & 0.047 & 28.47 & 0.201 & 0.036 & 0.118 \\

            & & 5-shot & 35.33 & 0.152 & 0.106 & 0.098 & 17.09 & 0.010 & 0.353 & 0.095 & 43.72 & \underline{0.071} & 0.326 & 0.128 & 21.22 & \underline{0.069} & 0.136 & 0.060 & 18.49 & \textbf{0.103} & 0.036 & 0.148 \\

            \cmidrule(lr){2-23}
            
            & \multirow{3}{*}{Claude-3.7*} & 0-shot 
            & 42.99 & 0.117 & 0.068 & 0.127 & 53.62 & 0.001 & 0.520 & 0.079 & 87.00 & 0.071 & \underline{0.109} & 0.126 & 27.96 & 0.041 & 0.080 & 0.087 & 66.22 & 0.138 & 0.075 & 0.139 \\

            & & 1-shot & 33.61 & 0.111 & 0.083 & \underline{0.127} & 36.33 & 0.003 & 0.416 & \underline{0.096} & 55.11 & 0.067 & 0.204 & 0.128 & \underline{16.32} & 0.021 & \underline{0.227} & 0.095 & 44.09 & \underline{0.109} & \textbf{0.109} & 0.144 \\

            & & 5-shot & \underline{18.86} & 0.103 & 0.102 & 0.122 & \underline{13.14} & 0.004 & 0.332 & 0.000 & \underline{27.34} & 0.061 & 0.112 & \underline{0.139} & 17.34 & 0.025 & 0.130 & \underline{0.100} & \underline{17.49} & 0.062 & \underline{0.105} & \underline{0.175} \\

            \cmidrule(lr){2-23}

            \rowcolor{LightGreen}
            & Ours & - & \textbf{18.38} & \textbf{0.228} & 0.121 & \textbf{0.154} & \textbf{10.71} & \underline{0.014} & \textbf{0.086} & \textbf{0.185} & \textbf{21.08} & \textbf{0.092} & 0.138 & \textbf{0.221} & \textbf{8.68} & 0.074 & \textbf{0.241} & \textbf{0.121} & \textbf{16.84} & 0.084 & 0.070 & \textbf{0.246} \\
            \hline

            \multirow{10}{*}{\textbf{C+S$\rightarrow$P}} & \multirow{3}{*}{GPT-4o} & 0-shot & 64.67 & 0.448 & 0.363 & 0.091 & 106.97 & 0.043 & 4.759 & 0.052 & 112.38 & 0.332 & 0.765 & 0.076 & 61.67 & 0.187 & 0.905 & 0.049 & 58.49 & 0.743 & 0.852 & 0.075 \\

            & & 1-shot & 52.66 & 0.447 & \underline{0.094} & 0.132 & 52.16 & 0.027 & 0.396 & 0.100 & 69.15 & 0.242 & 0.292 & 0.143 & 19.53 & 0.163 & 0.085 & 0.102 & 32.53 & 0.669 & \underline{0.115} & 0.150 \\

            & & 5-shot & 46.00 & 0.514 & 0.113 & 0.136 & 30.81 & 0.034 & 0.516 & 0.108 & 63.56 & 0.259 & 0.298 & 0.146 & 13.33 & 0.169 & 0.121 & 0.116 & 26.73 & 0.632 & 0.101 & 0.176 \\

            \cmidrule(lr){2-23}

            & \multirow{3}{*}{Gemini-2.5*} & 0-shot & 139.01 & 1.103 & 0.751 & 0.057 & 117.93 & 0.034 & 6.159 & 0.039 & 110.78 & 0.259 & 0.969 & 0.085 & 43.38 & 0.138 & 0.937 & 0.050 & 62.75 & 0.994 & 0.788 & 0.063 \\

            & & 1-shot & 94.61 & 0.480 & 0.398 & 0.103 & 65.27 & \textbf{0.015} & 1.470 & 0.089 & 88.37 & 0.264 & 0.600 & 0.124 & 15.94 & 0.088 & 0.546 & 0.082 & 28.41 & 0.333 & 0.379 & 0.154 \\

            & & 5-shot & 73.67 & 0.534 & 0.316 & 0.117 & 35.17 & 0.035 & 0.700 & 0.127 & 66.91 & 0.305 & 0.426 & 0.134 & 13.06 & 0.116 & 0.398 & 0.094 & 33.06 & 0.391 & 0.319 & 0.177 \\

            \cmidrule(lr){2-23}
            
            & \multirow{3}{*}{Claude-3.7*} & 0-shot & 26.86 & 0.147 & 0.103 & 0.136 & 30.80 & 0.002 & \underline{0.300} & 0.127 & 39.05 & \textbf{0.086} & 0.247 & 0.160 & 12.69 & \textbf{0.054} & 0.170 & 0.096 & 26.47 & 0.236 & \textbf{0.116} & 0.161 \\

            & & 1-shot & \underline{20.04} & 0.136 & 0.130 & 0.146 & \underline{17.91} & 0.005 & 0.381 & 0.147 & \underline{33.47} & \underline{0.078} & \underline{0.226} & 0.171 & 9.75 & 0.028 & \textbf{0.274} & 0.113 & 22.69 & \textbf{0.142} & 0.143 & 0.186 \\

            & & 5-shot & 21.47 & \underline{0.159} & \textbf{0.082} & \underline{0.156} & 18.75 & \underline{0.006} & 0.409 & \underline{0.159} & 34.77 & 0.061 & 0.331 & \underline{0.178} & \underline{7.64} & 0.031 & 0.390 & \underline{0.122} & \underline{14.14} & 0.257 & 0.084 & \underline{0.222} \\

            \cmidrule(lr){2-23}
            \rowcolor{LightGreen}
            & Ours & - & \textbf{16.92} & \textbf{0.366} & 0.122 & \textbf{0.219} & \textbf{6.13} & 0.021 & \textbf{0.188} & \textbf{0.240} & \textbf{20.74} & 0.130 & \textbf{0.174} & \textbf{0.256} & \textbf{5.42} & \underline{0.083} & \underline{0.235} & \textbf{0.200} & \textbf{9.02} & \underline{0.162} & 0.085 & \textbf{0.360} \\
            \hline

            \multirow{10}{*}{\textbf{Compl.}} & \multirow{3}{*}{GPT-4o} & 0-shot & 61.20 & 0.240 & \underline{0.051} & \textbf{0.522} & 97.60 & 0.227 & \textbf{0.057} & 0.000 & 155.36 & 0.115 & \textbf{0.072} & 0.075 & 116.18 & 0.124 & 0.068 & 0.000 & 93.49 & 0.068 & 0.063 & 0.000 \\

            & & 1-shot & 44.68 & \textbf{0.309} & \textbf{0.045} & 0.131 & 84.59 & 0.144 & \underline{0.058} & 0.000 & 130.47 & 0.125 & \underline{0.083} & 0.139 & 78.13 & 0.168 & 0.060 & \underline{0.320} & 70.53 & 0.112 & 0.068 & 0.000 \\

            & & 5-shot & 33.44 & \underline{0.340} & 0.052 & 0.268 & 50.11 & 0.090 & 0.118 & 0.000 & 86.84 & 0.143 & 0.099 & 0.169 & 39.22 & 0.184 & 0.087 & 0.290 & 46.80 & 0.135 & 0.060 & 0.438 \\

            \cmidrule(lr){2-23}
            
            & \multirow{3}{*}{Gemini-2.5*} & 0-shot & 108.60 & 0.511 & 7.337 & 0.219 & 95.02 & 0.165 & 0.252 & 0.000 & 111.59 & 0.209 & 0.463 & \textbf{0.355} & 91.24 & 0.225 & 0.929 & 0.210 & 52.29 & 0.254 & 0.778 & 0.284 \\

            & & 1-shot & 86.44 & 0.553 & 0.699 & 0.302 & 81.28 & 0.184 & 5.176 & 0.000 & 89.80 & 0.135 & 0.227 & 0.168 & 50.65 & 0.243 & 0.578 & \textbf{0.402} & 35.75 & 0.247 & 0.300 & 0.272 \\

            & & 5-shot & 65.85 & 0.394 & 0.360 & 0.310 & 45.02 & 0.049 & 0.166 & 0.000 & 68.69 & 0.127 & 0.223 & 0.180 & 36.62 & 0.289 & \textbf{0.220} & 0.025 & \textbf{28.30} & 0.164 & 0.073 & 0.440 \\

            \cmidrule(lr){2-23}
            
            & \multirow{3}{*}{Claude-3.7*} & 0-shot & 61.14 & 0.135 & 0.054 & 0.275 & 90.96 & 0.025 & 0.072 & 0.000 & 118.13 & 0.062 & 0.103 & 0.195 & 63.31 & \underline{0.063} & 0.042 & 0.000 & 77.85 & 0.067 & 0.053 & 0.000 \\

            & & 1-shot & 54.28 & 0.103 & 0.190 & 0.259 & 53.40 & 0.008 & 0.173 & 0.000 & 100.98 & 0.042 & 0.209 & 0.232 & 45.10 & \textbf{0.061} & 0.110 & 0.000 & 68.76 & 0.070 & \underline{0.082} & \textbf{1.000} \\

            & & 5-shot & \textbf{29.74} & 0.225 & 0.111 & 0.331 & \textbf{18.54} & \textbf{0.012} & 0.167 & 0.000 & \underline{47.29} & \textbf{0.072} & 0.148 & 0.172 & \textbf{25.88} & 0.071 & 0.139 & 0.089 & 36.78 & \underline{0.112} & \textbf{0.128} & 0.329 \\

            \cmidrule(lr){2-23}
            \rowcolor{LightGreen}
            & Ours & - & \underline{31.58} & 0.235 & 0.123 & \underline{0.478} & \underline{22.48} & \underline{0.013} & 0.098 & 0.000 & \textbf{38.56} & \underline{0.098} & 0.153 & \underline{0.288} & \underline{25.92} & 0.068 & \underline{0.203} & 0.310 & \underline{30.56} & \textbf{0.106} & 0.070 & \underline{0.620} \\
            \hline

            \multirow{10}{*}{\textbf{Refin.}} & \multirow{3}{*}{GPT-4o} & 0-shot & 12.71 & 0.371 & 0.162 & 0.616 & 67.25 & 0.040 & 0.172 & 0.628 & 7.76 & 0.198 & 0.108 & 0.654 & 5.88 & 0.121 & \textbf{0.278} & 0.577 & 3.27 & 0.178 & \underline{0.127} & 0.618 \\

            & & 1-shot & \underline{6.75} & 0.392 & \underline{0.157} & 0.646 & 23.67 & 0.042 & 0.175 & 0.639 & 7.63 & 0.190 & 0.104 & 0.670 & \underline{4.32} & 0.125 & \underline{0.286} & 0.599 & 2.10 & 0.162 & 0.134 & 0.640 \\

            & & 5-shot & 10.25 & 0.397 & 0.180 & 0.650 & 31.24 & 0.040 & 0.170 & 0.639 & 8.92 & 0.194 & 0.116 & 0.672 & 4.89 & 0.124 & 0.291 & 0.601 & \underline{1.38} & 0.198 & 0.134 & 0.658 \\

            \cmidrule(lr){2-23}
            
            & \multirow{3}{*}{Gemini-2.5*} & 0-shot & 23.88 & 0.394 & 0.171 & 0.631 & 8.92 & 0.034 & 0.186 & 0.627 & 20.76 & 0.206 & 0.111 & 0.661 & 10.59 & 0.125 & 0.350 & 0.585 & 5.78 & 0.203 & 0.167 & 0.624 \\

            & & 1-shot & 9.72 & 0.386 & 0.165 & 0.656 & \textbf{1.05} & 0.036 & 0.182 & 0.638 & 8.62 & 0.196 & 0.105 & 0.665 & 5.21 & 0.124 & 0.328 & 0.597 & 3.79 & 0.188 & 0.159 & 0.646 \\

            & & 5-shot & 8.18 & 0.392 & 0.158 & \underline{0.657} & \underline{1.27} & 0.034 & 0.183 & 0.637 & 6.87 & 0.200 & \underline{0.104} & 0.669 & \textbf{1.02} & 0.124 & 0.321 & 0.606 & \textbf{1.32} & 0.189 & 0.163 & \underline{0.661} \\

            \cmidrule(lr){2-23}
            
            & \multirow{3}{*}{Claude-3.7*} & 0-shot & 15.02 & \underline{0.272} & 0.176 & 0.603 & 3.86 & 0.028 & 0.118 & 0.635 & 17.93 & 0.116 & 0.304 & 0.635 & 6.08 & 0.095 & 0.375 & 0.584 & 1.67 & 0.136 & \textbf{0.127} & 0.651 \\

            & & 1-shot & 18.97 & 0.231 & 0.543 & 0.638 & 3.12 & 0.028 & \underline{0.102} & \underline{0.643} & 11.01 & \textbf{0.099} & 0.624 & 0.663 & 16.28 & \textbf{0.073} & 0.729 & \underline{0.608} & 4.22 & \textbf{0.111} & 0.410 & 0.650 \\

            & & 5-shot & 11.23 & \textbf{0.284} & 0.378 & 0.642 & 3.85 & \underline{0.027} & 0.165 & 0.638 & \textbf{3.33} & 0.175 & 0.173 & \underline{0.676} & 5.84 & 0.093 & 0.545 & 0.596 & 3.72 & 0.158 & 0.325 & 0.631 \\

            \cmidrule(lr){2-23}

            \rowcolor{LightGreen}
            & Ours & - & \textbf{4.51} & 0.317 & \textbf{0.145} & \textbf{0.681} & 10.60 & \textbf{0.017} & \textbf{0.064} & \textbf{0.732} & \underline{4.73} & \underline{0.113} & \textbf{0.072} & \textbf{0.752} & 6.66 & \underline{0.079} & 0.295 & \textbf{0.641} & 8.25 & \underline{0.132} & 0.138 & \textbf{0.708} \\
            \hline

            \multicolumn{3}{c}{\textbf{Test Data}} & - & 0.289 & 0.010 & - & - & 0.012 & 0.051 & - & - & 0.083 & 0.054 & - & - & 0.062 & 0.266 & - & - & 0.097 & 0.126 \\
            
            \bottomrule[1.2pt]
    	\end{tabular}
    }
    \label{tab:complete_LLMs_comparision}
     \vspace{-5mm}
\end{table*}

\clearpage

\end{document}